%% file: main.tex
\documentclass{article} 
\usepackage{iclr2027_conference,times}

\input{math_commands.tex}

\usepackage{hyperref}
\usepackage{url}

\usepackage{algorithm}     
\usepackage{algpseudocode} 
\usepackage{hyperref}
\usepackage{url}
\usepackage{microtype}
\usepackage{booktabs}
\usepackage{amsmath}
\usepackage{amssymb}
\usepackage{graphicx}
\usepackage{subcaption}
\usepackage{xcolor}
\usepackage{tikz}
\usepackage{comment}
\usepackage{placeins}

\newcommand{\appbarrier}{\FloatBarrier}
\usetikzlibrary{arrows.meta,positioning,shapes.geometric}

\title{LLM Benchmark Compression via Maximum Independent Set Prompt Selection on Proximity Graphs}

\author{Denica Kjorvezir$^{1,2}$, Marko Djukanovi\'c$^{3}$, Ana Gjorgjevikj$^{1}$, Gjorgjina Cenikj$^{1}$, \\
\textbf{Kiril Gashteovski$^{4,5}$, Tome Eftimov$^{1,2}$} \\[4pt]
$^{1}$Computer Systems Department, Jo\v{z}ef Stefan Institute, Ljubljana, Slovenia \\
$^{2}$Jo\v{z}ef Stefan International Postgraduate School, Ljubljana, Slovenia \\
$^{3}$Center for Astrophysics and Cosmology, University of Nova Gorica, Nova Gorica, Slovenia \\
$^{4}$NEC Laboratories Europe, Germany \\
$^{5}$CAIR, Ss.\ Cyril and Methodius University of Skopje, North Macedonia
}

\iclrfinalcopy

\begin{document}
\maketitle
\lhead{}
\renewcommand{\headrulewidth}{0pt} 
\begin{abstract}
Evaluating LLMs across comprehensive benchmarks is expensive, motivating benchmark compression: replacing a full benchmark with a compact subset that is cheaper to run. Whether such a method is usable comes down to fidelity---when does the subset reproduce the full-benchmark LLM ranking, and when does it diverge? We model each benchmark as a prompt proximity graph---nodes are prompts, and an edge joins two prompts whose embeddings are close enough to be treated as semantically redundant---and apply Maximum Independent Set (MIS) solvers to select a maximally diverse, non-redundant subset in which no semantic region is doubly counted. We evaluated four MIS solvers across six embeddings, three distances, six percentile thresholds, and four benchmarks (GPQA, IFEval, MMLU-Pro, Omni-MATH) covering 66 LLMs, spanning 2{,}558 configurations. Two axes emerge and move almost independently. First, selection is reproducible: different solver seeds yield near-identical rankings (Kendall's $W \geq 0.90$ in 84.9\% of stochastic configurations; mean $W = 0.950 \pm 0.089$), so a single published subset is a stable artifact. Second, agreement with the full-benchmark ranking is governed by how aggressively the graph is pruned, giving a controllable efficiency--fidelity trade-off: mild thresholds ($p_{10}$--$p_{20}$) preserve the full ranking ($\rho = 0.97$--$0.98$) while removing 64--82\% of prompts, whereas aggressive pruning ($p_{80}$--$p_{95}$) trades fidelity for compression ($\rho = 0.67$--$0.81$). Compared with random subsets drawn at the subset size determined by MIS, MIS is close in fidelity at mild compression but not better on average; its distinctive property is label-free coverage redistribution (more category-balanced than 92--99\% of same-size random subsets). Control analyses show that embedding proximity captures category and difficulty structure, not prompt interchangeability. We recommend ReduMIS with \texttt{qwen8b} embeddings, standardized-Euclidean distance, and a $p_{20}$ threshold as a default when a reproducible, coverage-balanced subset is the goal.
\end{abstract}

\section{Introduction}
Rigorous evaluation of large language models (LLMs) requires benchmarks large enough to yield statistically reliable scores, yet small enough to be tractable at scale. Widely used benchmarks such as MMLU-Pro~\citep{wang2024mmlu} and GPQA~\citep{rein2023gpqa} contain hundreds to thousands of questions, incurring substantial inference costs for frontier LLMs and large-scale leaderboards~\citep{polo2024tinybenchmarks,liang2022holistic}. This motivates \emph{benchmark compression}: replacing a full benchmark with a compact subset that is far cheaper to run. The question that decides whether a compression method is usable is one of \emph{fidelity}: when does the subset reproduce the LLM ranking of the full benchmark, and when does it diverge? This paper is organized around characterizing that question so that the choice can be made deliberately.

\begin{figure}[t]
  \centering
  \includegraphics[width=\textwidth]{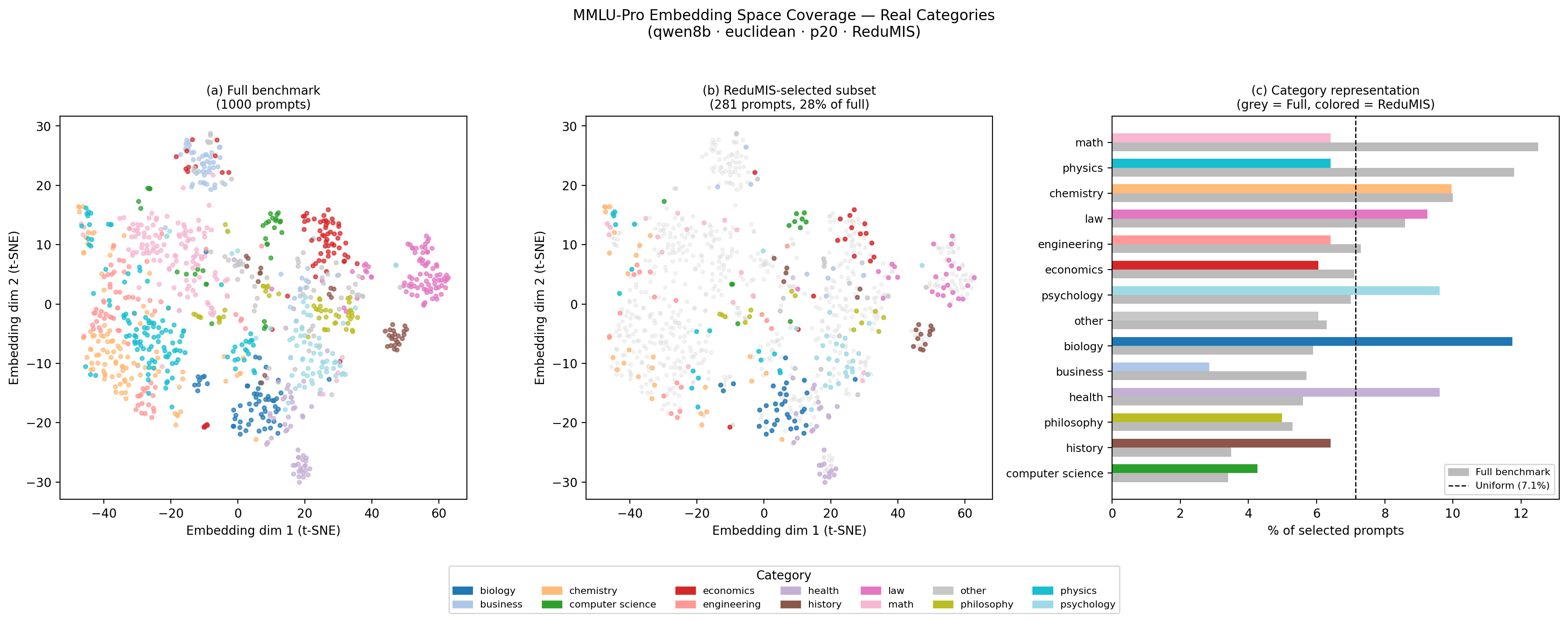}
  \caption{Embedding-space coverage on \textbf{real MMLU-Pro categories} (\texttt{qwen8b},
  standardized Euclidean, $p_{20}$, ReduMIS, no chunking). (a)~Full benchmark ($1000$ prompts), colored by
  ground-truth category; some categories (e.g.\ math, physics) are overrepresented. (b)~The
  ReduMIS-selected subset ($281$ prompts) spreads across the space rather than concentrating in the
  dense regions. (c)~Category shares within each set: the full benchmark (grey) departs from the
  uniform share ($7.1\%$, dashed), and MIS selection (colored) moves the most overrepresented categories
  toward it (math and physics: $12.5\%$ and $11.8\%\to6.4\%$) while increasing several smaller ones
  (biology $5.9\%\to11.7\%$, health $5.6\%\to9.6\%$).}
  \label{fig:coverage}
\end{figure}

Our objective is therefore \emph{efficiency}: select a compact, representative subset and quantify how faithfully it preserves the relative ordering of LLMs as a function of how aggressively it compresses. A subset can compress a benchmark with little change in the rankings or with substantial change, and neither is universally preferable. The right choice depends on whether the practitioner values maximal compression or maximal fidelity. A second, \emph{interpretive} question arises whenever a compressed ranking does diverge from the full one: is the divergence merely information lost to pruning, or does it reflect \emph{coverage structure} in the original benchmark? If a benchmark's prompts are unevenly distributed in embedding space---overrepresenting certain semantic regions---then aggregate scores implicitly upweights the capabilities those regions probe; for instance, a reasoning benchmark with disproportionately many similar chemistry questions rewards chemistry ability more than its label suggests~\citep{bohacek2025competency}. We treat this coverage lens as a way to \emph{describe} what changes under compression, not as a premise that the full benchmark is wrong or that the compressed ranking is more correct. Both the efficiency objective and the coverage lens are the consequence of an identical structural feature of the prompts: \emph{semantic redundancy}. When multiple prompts are close in the embedding space, they \textit{(i)}~spend evaluation budget on similar content and \textit{(ii)}~tend to share category and difficulty, so dense regions weigh more in the aggregate score (Section~\ref{sec:results} quantifies how far this proximity extends to model behavior). Our compression is based on the removal of redundancy; therefore, it also reshapes benchmark prompt distribution. We model each benchmark as a prompt proximity graph $G=(V,E)$, where each prompt is a node $v\in V$ and an edge $(u,v)\in E$ connects two prompts whenever they are close in the embedding space based on different distance thresholds to be considered semantically redundant, i.e.\ $d(e_u,e_v)\le\theta$ for a distance threshold $\theta$. A Maximum Independent Set (MIS) of this graph---the largest set of mutually \emph{non-adjacent} nodes---is exactly a maximally non-redundant subset: any two selected prompts satisfy $d(e_i,e_j)>\theta$, so no two selected prompts are semantically close and no semantic region is doubly counted. Figure~\ref{fig:pipeline} shows the full pipeline; Figure~\ref{fig:coverage} illustrates the coverage redistribution that accompanies selection on real MMLU-Pro benchmark categories.

This pipeline gives rise to two of our central empirical questions: \emph{(i) how does a compressed subset's agreement with the full-benchmark ranking behave as compression increases, and (ii) is the selection procedure itself reliable?} Agreement with the full-benchmark ranking is not fixed: it is high under mild compression and falls smoothly as pruning becomes aggressive, tracing a \emph{controllable} efficiency--fidelity trade-off. We map this trade-off across the configuration grid so that a practitioner can pick an operating point rather than accept a fixed one, and we separately characterize what changes when the rankings diverge. We find that repeated stochastic MIS selection under different random seeds yields \emph{consistent} rankings---seeds agree closely with each other---so a published subset is reproducible. 
\textbf{Contributions:}
\textbf{(i)} \emph{A graph-theoretic prompt-selection framework} for benchmark compression that selects a maximally diverse, non-redundant subset via MIS, supporting four MIS solvers, six prompt embedding methods, three distance metrics, and six percentile thresholds, with selected subset size inferred from the graph rather than fixed a priori.
\textbf{(ii)} \emph{An empirical characterization of the compression--fidelity trade-off} across 2{,}558 configurations and 66 LLMs on four benchmarks. The two governing axes, random seed concordance and full-benchmark agreement, move almost independently: concordance is high ($W \geq 0.90$ in 84.9\% of stochastic configurations), while agreement tracks pruning aggressiveness---mild thresholds ($p_{10}$--$p_{20}$) preserve rankings ($\rho = 0.97$--$0.98$), with divergence concentrated at high thresholds and on GPQA/IFEval benchmarks. This turns the grid into an operating-point recipe: the threshold sets whether a configuration stays reliable-and-faithful, and the embedding sets how it fails.
\textbf{(iii)} \emph{Baselines and validation.} We compare MIS with random, stratified and uniform-stratified sampling and with other embedding-only selectors (k-center, k-medoids, k-means, SemDeDup) at exactly the same subset size. Random sampling is slightly closer to the full ranking on average, and diversity-oriented selectors (MIS, k-center, SemDeDup) behave alike; MIS's distinctive contribution is a reproducible subset with label-free coverage redistribution. Behavioral and annotation-based tests with control pairs show that embedding proximity captures category and difficulty structure, but not prompt interchangeability.

\section{Related Work}
\noindent\textbf{LLM benchmark design and efficiency.} LLM benchmarks span reasoning~\citep{wang2024mmlu,rein2023gpqa}, instruction following~\citep{zhou2023instruction}, and advanced mathematics~\citep{cobbe2021training,hendrycks2021measuring,gao2025omni}, with large suites such as HELM~\citep{liang2022holistic}, Chatbot Arena~\citep{zheng2023judging}, the Open LLM Leaderboard~\citep{open-llm-leaderboard}, and LiveBench~\citep{white2024livebench} providing standardized evaluation. Because full-benchmark evaluation is costly, prior work has explored subset selection while preserving ranking fidelity~\citep{perlitz2023efficient,vivek2024anchor}. Item Response Theory (IRT) methods use item difficulty, discrimination, or model ability to select informative instances~\citep{polo2024tinybenchmarks,zhou2026lost,truong2025reliable}, while other work studies frontier degradation, benchmark inconsistencies, and multidimensional latent structure~\citep{zhang2026benchmark,schilling2025lifting,chen2025learning}. These approaches typically estimate model ability on a fixed benchmark and require model interaction. Our approach is complementary and \emph{model-agnostic}: it selects representative prompts from semantic proximity graphs without model evaluations during subset construction. Appendix~\ref{app:related_table} (Table~\ref{tab:related_appendix}) further situates our method by selection signal and granularity. \textbf{Coverage bias in benchmarks.} Overrepresentation of particular question types can amplify the capabilities they probe. \citet{bohacek2025competency} show that a small number of concepts dominate coverage in popular benchmarks and disproportionately affect aggregate scores. Other work documents sensitivity to prompt phrasing and annotation artifacts~\citep{mizrahi2024state,gururangan2018annotation}. We formalize overrepresentation geometrically as prompt density in embedding space: MIS selection redistributes mass away from dense regions because selected prompts cannot be semantically close. We use this descriptively to explain changes under compression rather than assuming that redistribution improves rankings. \textbf{Graph-based selection.} Graph-based methods such as SELECTOR~\citep{selector} construct compact, coverage-preserving subsets, while related approaches have been applied to time-series classification, optimization, and algorithm portfolios~\citep{less_is_more,transoptas,kostovska2023ps}. Unlike clustering, MIS determines the subset directly from the graph. We systematically evaluate graph-based semantic prompt selection across six embeddings, three distance measures, six data-driven thresholds, and four MIS solvers on four NLP benchmarks and 66 LLMs.

\section{Methodology}
\label{sec:method}
Figure~\ref{fig:pipeline} shows the full pipeline. Throughout, we describe the construction from the perspective of \emph{semantic redundancy}: two prompts are linked when they are close enough to be semantically redundant based on a particular distance metric, and the selected subset consists of mutually \emph{non-redundant} prompts. 
\begin{figure}[t]
  \centering
  \begin{tikzpicture}[
      font=\small,
      box/.style={rectangle, rounded corners=3pt, draw=black!55, line width=0.6pt,
                  align=center, minimum height=1.55cm, text width=2.05cm, inner sep=3pt},
      arr/.style={-{Latex[length=2.2mm]}, line width=0.8pt, black!65}]
    \node[box, fill=orange!12]  (p)  {\textbf{Benchmark}\\\textbf{prompts}\\[1pt]\scriptsize $N\!=\!446$--$1000$};
    \node[box, fill=blue!10, right=0.7cm of p]   (e)  {\textbf{Embedding}\\\textbf{model}\\[1pt]\scriptsize BGE, E5, GIST,\\ Nemotron, Qwen};
    \node[box, fill=green!10, right=0.7cm of e]  (g)  {\textbf{Proximity}\\\textbf{graph}\\[1pt]\scriptsize $G\!=\!(V,E)$\\ edge if $d\!\le\!\theta$};
    \node[box, fill=purple!10, right=0.7cm of g] (m)  {\textbf{MIS}\\\textbf{solver}\\[1pt]\scriptsize CPLEX, GREEDY,\\ Online-MIS, ReduMIS};
    \node[box, fill=red!10, right=0.7cm of m]    (s)  {\textbf{Subset}\\ $S\!\subseteq\!\Pi$\\\textbf{+ ranking}\\[1pt]\scriptsize $\rho$ over 66 LLMs};
    \draw[arr] (p) -- (e); \draw[arr] (e) -- (g); \draw[arr] (g) -- (m); \draw[arr] (m) -- (s);
  \end{tikzpicture}
  \caption{The MIS-based prompt-selection pipeline. Prompts are encoded; a proximity graph is
  constructed by connecting prompts whose embeddings are within a distance threshold $\theta$
  (i.e.\ semantically redundant); an MIS solver selects a maximally diverse, non-redundant subset
  $S\subseteq\Pi$ of mutually distant prompts for LLM evaluation.}
  \label{fig:pipeline}
\end{figure}

\noindent\textbf{(1) Embedding construction.} For each prompt $p_i\in\Pi$ a fixed-dimensional vector is created by an embedding model $f:\mathcal{P}\rightarrow\mathbb{R}^d$, $e_i=f(p_i)$. Embeddings place semantically similar inputs near one another, enabling distance-based reasoning about redundancy. Proximity is measured by a distance function $d$  on normalized vectors; larger $d$ means \emph{less} close in the embedding space with respect to the distance metric. \textbf{(2) Construction of the proximity graph.} Given embeddings $\{e_1,\dots,e_N\}$, we build a graph $G=(V,E)$ that makes redundancy explicit. Each prompt $p_i$ is a node $v_i$. An undirected edge $(v_i,v_j)$ is added whenever the distance between the embeddings of two prompts is at most a threshold, $d(e_i,e_j)\le\theta$. Therefore, edges connect \emph{redundant} pairs. The threshold $\theta$ is set to the $p$-th percentile of the empirical pairwise-distance distribution, giving a normalized, model-independent knob on graph density. A \emph{small} percentile ($p_{10}$) yields a \emph{small} $\theta$: only the very closest, most redundant pairs are connected, so the graph is sparse and little is pruned. A \emph{large} percentile ($p_{95}$) yields a \emph{large} $\theta$: even moderately distant pairs are treated as redundant, the graph is dense, and pruning is aggressive. \textbf{(3) Maximum Independent Set.} We cast compression as finding the largest subset $S\subseteq V$ with no edge inside it: for any $p_i,p_j\in S$, $(v_i,v_j)\notin E$, i.e.\ $d(e_i,e_j)>\theta$. Formally, the solver searches $G$ for an independent set of maximum cardinality, $S^\star=\arg\max_{S\subseteq V}\lvert S\rvert$ subject to $(v_i,v_j)\notin E$ for all $v_i,v_j\in S$ (more details in Appendix~\ref
{app:MIS_formulation}). The selected prompts are therefore pairwise \emph{non-redundant}: no two are within distance $\theta$, so at most one prompt is drawn from any embedding-space region of radius $\theta/2$. Random or stratified sampling does not provide such guarantee. Because the retained prompts jointly cover the space without redundancy, changes in downstream rankings reflect the semantic structure removed by compression, not merely sampling noise. \textbf{(4) Evaluation metrics for validation.} To address our first empirical question for each subset $S$ we compute mean LLM accuracy restricted to $S$, yielding a ranking $\sigma(S)$, and compare it to the full-benchmark ranking $\sigma(\Pi)$ via Spearman's $\rho$. For descriptive interpretation, we categorized the magnitude of Spearman's rank correlation as negligible ($|\rho| < 0.10$), weak ($0.10 \leq |\rho| < 0.40$), moderate ($0.40 \leq |\rho| < 0.70$), strong ($0.70 \leq |\rho| < 0.90$), and very strong ($|\rho| \geq 0.90$), following~\cite{schober2018correlation}. Positive values indicate agreement between rankings, while negative values indicate inverse agreement. In the case of the second empirical question, for stochastic solvers, agreement \emph{across seeds} is measured by Kendall's coefficient of concordance $W=\frac{12\mathcal{S}}{m^2(n^3-n)}$,
where $m$ is the number of seeds, $n$ the number of LLMs, and $\mathcal{S}=\sum_i(R_i-\bar R)^2$ the sum of squared deviations of per-item rank sums. We computed \(W\) both with and without a correction for tied ranks. $W\in[0,1]$, with $W=1$ indicating perfect agreement. Following~\citet{landis1977measurement}, $W\ge0.90$ is near-perfect, $W\ge0.80$ substantial, $W<0.60$ poor; we use $W\ge0.90$ as the high-concordance threshold.

\section{Experimental Setup}
\label{sec:setup}

\noindent\textbf{Benchmarks.} We evaluate on four benchmarks from the HELM Capabilities leaderboard~\citep{xu2025helmcapabilities} spanning knowledge-intensive QA (MMLU-Pro~\citep{wang2024mmlu} (1000 prompts), GPQA~\citep{rein2023gpqa} (446 prompts)), instruction following (IFEval~\citep{zhou2023instruction} (541 prompts)), and mathematical reasoning (Omni-MATH~\citep{gao2025omni} (1000 prompts)). We restrict to single-turn benchmarks so every instance is an independent text embedding. Table~\ref{tab:benchmarks} from Appendix~\ref{app:models} summarizes them across a fixed set of $66$ frontier and open LLMs, including GPT~\citep{singh2025openai,agarwal2025gpt}, Claude~\citep{anthropic2025claude4}, Gemini~\citep{team2023gemini}, Llama~\citep{grattafiori2024llama}, Qwen2.5/Qwen3~\citep{qwen2025,yang2025qwen3}, DeepSeek~\citep{liu2024deepseek}, Mistral~\citep{jiang2023mistral} and OLMo~\citep{olmo2024furious}. Scores were collected and reused from the HELM Capabilities leaderboard; as HELM may reflect differing prompting strategies across families, comparisons are relative to HELM-reported baselines rather than a controlled protocol. MMLU-Pro and GPQA aggregate binary per-item correctness (smooth accuracy, strong rank separability), while IFEval and Omni-MATH produce more discretized distributions; we therefore use rank-based metrics ($\rho$, $W$), both robust to discretization. \textbf{Sentence embeddings.} We use six models in two regimes: three small encoder-based models (\emph{bge-large-en-v1.5}~\citep{xiao2024c}, \emph{e5-large}~\citep{wang2022text}, \emph{GIST-small-Embedding-v0}~\citep{solatorio2024gistembed}) and three large decoder-based models (\emph{Llama-Embed-Nemotron-8B}~\citep{babakhin2025llama}, \emph{Qwen3-4B}, \emph{Qwen3-8B}~\citep{zhang2025qwen3}). Encoders use a $\sim$512-token window; we apply length-weighted averaging over non-overlapping 510-token chunks. Decoders support 4{,}096 tokens (sufficient for all prompts; max length $2{,}278.5$ tokens); we additionally evaluate a chunked decoder variant (overlapping windows, stride 256). All models use attention-mask-weighted mean pooling followed by L2 normalization. \textbf{Distance measures and thresholds.} We evaluate cosine, Pearson-correlation, and standardized-Euclidean (\emph{seuclidean}) distances. The threshold $\theta$ is a percentile of the empirical pairwise-distance distribution (estimated from $10{,}000$ random pairs); we sweep $p_{10}$ to $p_{95}$. Percentile thresholds give a model-independent way to control sparsity across embedding spaces with different distance scales (Appendix~\ref{app:data_driven}). \textbf{MIS algorithms.} We use four solvers. \textbf{CPLEX}~\citep{anand2017comparative} solves MIS exactly via an ILP with branch-and-bound/-cut. \textbf{GREEDY} iteratively selects the lowest-degree vertex and removes it and its neighbors. \textbf{Online-MIS}~\citep{DahlumLS0SW16} is a randomized streaming heuristic (five seeds). \textbf{ReduMIS}~\citep{LammSSSW17} combines kernelization and stochastic local search (five seeds). On a $1000$-node graph, CPLEX needs $\sim$30--120\,s, ReduMIS/Online-MIS up to 600\,s, and GREEDY under one second (more details in Appendix~\ref {app:compute}). \textbf{Configuration space.} The grid comprises embedding $\times$ distance $\times$ threshold $\times$ solver across four benchmarks; after removing a handful of runs that did not complete, we analyze $2{,}558$ unique configurations; stochastic solvers use five seeds each.

\begin{figure}[t]
  \centering
  \includegraphics[width=0.78\textwidth]{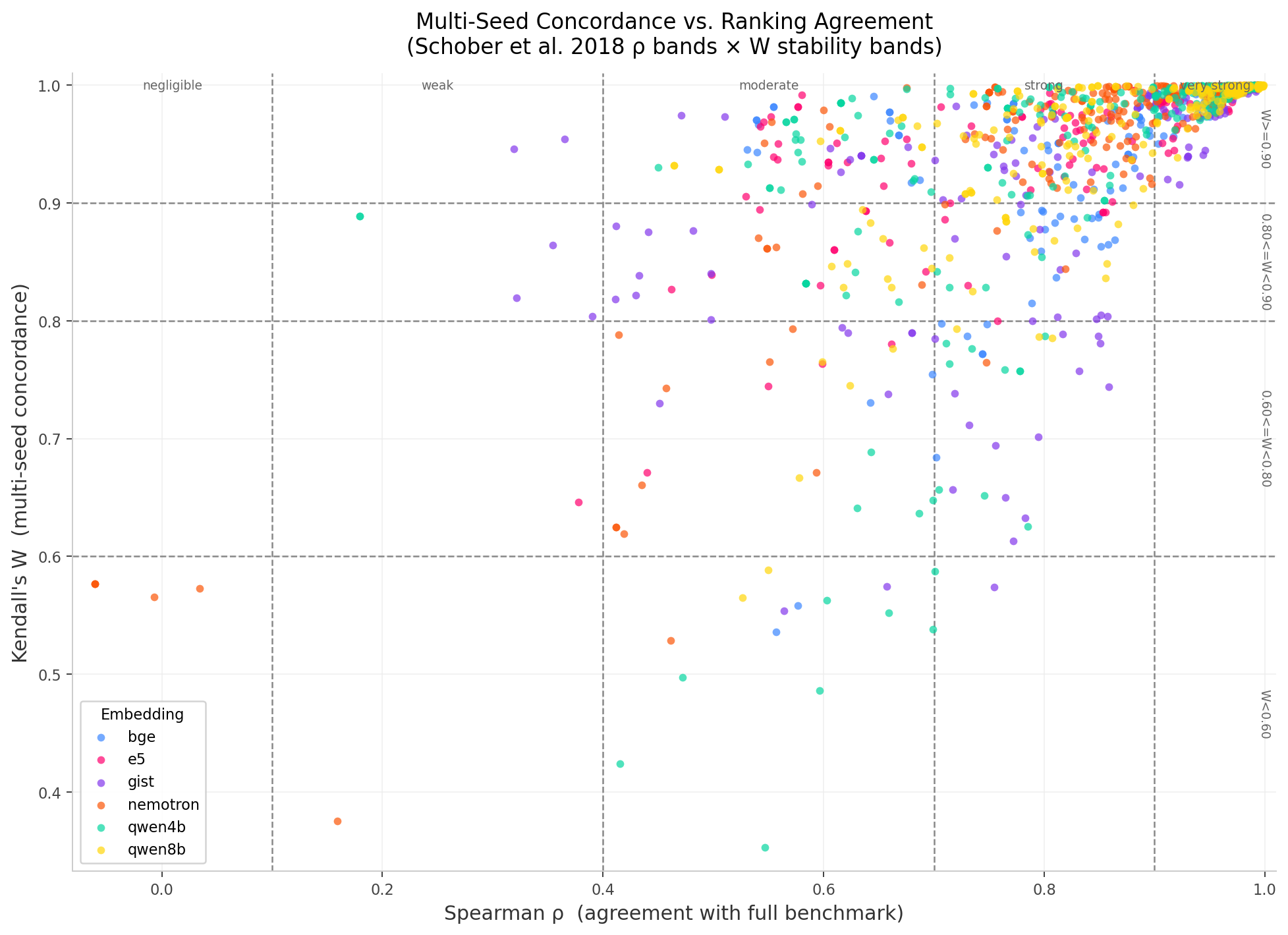}
 \caption{\textbf{Multi-seed concordance vs.\ ranking agreement.} Kendall's $W$
(concordance across seeds) against Spearman's $\rho$ (agreement with the
full-benchmark ranking) for every stochastic configuration, colored by embedding
model. Dashed grid lines are the interpretation bands defined in
Section~\ref{sec:method}: vertical lines split $\rho$ into the magnitude bands of
\citet{schober2018correlation} (negligible, weak, moderate, strong, very strong, at
$0.10/0.40/0.70/0.90$), and horizontal lines split $W$ into the concordance bands of
\citet{landis1977measurement} ($W<0.60$ poor; $0.60\le W<0.80$; $0.80\le W<0.90$
substantial; $W\ge0.90$ near-perfect). Points pile up in the top bands---$W\ge0.90$
in $84.9\%$ of configurations---so selection is reliable across seeds, while spreading
horizontally from the moderate to the very-strong $\rho$ bands: what varies under
compression is rank agreement with the full benchmark, not seed stability.}
  \label{fig:hypothesis}
\end{figure}

\section{Results}
\label{sec:results}

We organize the results into several parts. First, we examine selection consistency and the compression–fidelity trade-off, including the behavior of stochastic solvers. We then analyze where divergence is concentrated and compare the proposed methods with the baselines. \textbf{Consistency and the compression--fidelity trade-off.}
Figure~\ref{fig:hypothesis} places every stochastic configuration (embedding $\times$ distance $\times$ threshold $\times$ stochastic solver across four benchmarks) on the $W$--$\rho$
plane, gridded by the $\rho$-magnitude bands of \citet{schober2018correlation} and the
$W$-concordance bands of \citet{landis1977measurement} (the master results file records
each configuration's quadrant; see our repository). The two axes move almost
independently. Vertically, random seed concordance of the stochastic MIS is high nearly everywhere (mean
$W=0.950\pm0.089$; $84.9\%$ near-perfect at $W\ge0.90$, only $1.7\%$ poor), so a new seed
reproduces essentially the same ranking. Horizontally, agreement with the full benchmark
spreads across bands, from $51.6\%$ very strong ($\rho\ge0.90$) down to $1.1\%$ weak or
negligible. The two combine most often in the reliable-and-faithful corner ($\rho\ge0.70$ and $W\ge0.90$): $77.3\%$ of
configurations pair near-perfect seed concordance with strong-or-better baseline
agreement. Which quadrant a configuration lands in is set primarily by the
threshold---the compression level the practitioner chooses---so reshaping is neither
automatically better nor worse than preserving. We report Kendall's $W$ without tie correction, which can slightly underestimate concordance when ranks are tied, and show in Table~\ref{tab:rev_ties} in Appendix~\ref{app:distribution_bench} that applying the standard correction leaves our conclusions unchanged. \textbf{Where divergence concentrates.} Splitting the plane at $\rho=0.70$ and
$W=0.90$ gives four quadrants (Table~\ref{tab:quadrant_reco}), and divergence from the
reliable-and-faithful quadrant is unevenly distributed. \emph{By benchmark}, the reliable-and-faithful quadrant holds $86.7\%$ of Omni-MATH and $80.6\%$ of MMLU-Pro configurations but only $67.2\%$ of
GPQA and $73.8\%$ of IFEval; the two hardest diverge in complementary ways---IFEval into
reliable-but-reshaped ($15.4\%$: stable seeds, $\rho<0.70$), GPQA into doubly-degraded
($11.7\%$), consistent with its being the smallest, most discretized benchmark.
\emph{By threshold}, divergence is a high-percentile effect: $\ge98.6\%$ of configurations
stay reliable-and-faithful through $p_{50}$, but only $36.9\%$ do at $p_{95}$, and $100\%$
of $\rho<0.70$ and $98.4\%$ of $W<0.90$ cases occur at $p_{80}$/$p_{90}$/$p_{95}$.
\emph{By embedding}, large decoders stay on target far more than small encoders
($82.5\%$ vs.\ $67.4\%$; \texttt{nemotron} $88.5\%$ and \texttt{qwen8b} $83.8\%$ vs.\
\texttt{gist} $60.4\%$). These regularities make the trade-off predictable: benchmark,
threshold, and embedding together anticipate which quadrant a compression level falls in.
Appendix~\ref{app:distribution_bench} breaks these outcomes down per benchmark, embedding model, solver, and distance measure (Tables~\ref{tab:distribution_bench}--\ref{tab:algorithms}).

\textbf{Operating-point recommendations.} These patterns turn the plane into a
recipe (Table~\ref{tab:quadrant_reco}): the threshold decides whether a configuration
leaves the reliable-and-faithful quadrant, and the embedding decides \emph{how} it fails
once it does. Staying at $p_{20}$ or below reaches the target quadrant almost regardless of
other choices ($100\%$ of $p_{10}$/$p_{20}$ configurations); when harder compression is
required, a large decoder embedding with standardized-Euclidean distance keeps
configurations on target ($82.5\%$ vs.\ $67.4\%$ for small encoders, $82.9\%$ vs.\ $75.2\%$
for cosine; $100\%$ combined with a mild threshold). Our recommended point is
\texttt{qwen8b}/seuclidean/$p_{20}$.
 When a configuration does leave the quadrant, the
failure mode is diagnostic and fixable: aggressive thresholds reshape the ranking (lower
toward $p_{50}$ if a faithful proxy was intended), weak or small embeddings destabilize
seeds (switch to a larger embedding, prefer ReduMIS solver, or average more seeds), and pairing
aggressive pruning with weak embeddings on small benchmarks fails on both axes---avoid it
(e.g.\ \texttt{gist} @ $p_{90}$/$p_{95}$ is off-target $93.8\%$ of the time).  \textbf{Top-$N$ overlap across quadrants.} Because practitioners often care about the \emph{head}
of the leaderboard, we also compute the Jaccard overlap between the top-$5$ and top-$10$ models of the full
benchmark and of each subset (Table~\ref{tab:jaccard}, Appendix~\ref{app:distribution_bench}). Overlap declines
monotonically across the quadrants, from $0.561$/$0.678$ (top-$5$/top-$10$) in the reliable-and-faithful
quadrant to $0.163$/$0.240$ in the doubly-degraded one, where $21\%$ of configurations share no top-$5$ model
with the full benchmark. The reliable-but-reshaped quadrant keeps more of the head than the
faithful-but-seed-unstable one ($0.336$ vs.\ $0.254$ on top-$5$), so seed instability corrupts the top of the
leaderboard more than mild whole-ranking reshaping. Overlap falls with the threshold, from $0.684$/$0.809$ at
$p_{10}$ to $0.292$/$0.418$ at $p_{95}$.

\begin{table}[t]
\centering
\caption{Configuration patterns behind each quadrant of Figure~\ref{fig:hypothesis}, and the
resulting recommendation. Enrichments ($\times$) are relative to the base rate across all
$1{,}262$ stochastic configurations.}
\label{tab:quadrant_reco}
\small
\begin{tabular}{@{}p{2.9cm}p{5.0cm}p{4.4cm}@{}}
\toprule
\textbf{Quadrant} & \textbf{Dominant drivers} & \textbf{Recommendation} \\
\midrule
Reliable \& faithful \newline ($\rho\ge0.70,W\ge0.90$; $77.3\%$)
 & Mild thresholds ($p_{10}$/$p_{20}$: $\times1.29$ each, $100\%$ on target); large decoders ($\times1.07$;
 \texttt{nemotron} $\times1.14$); seuclidean ($\times1.07$)
 & Default target. Use $p_{20}$ or below; for harder compression, large\,+\,seuclidean ($89\%$ on target across thresholds) \\
\addlinespace
Reliable but reshaped \newline ($\rho<0.70,W\ge0.90$; $7.5\%$)
 & $p_{95}$ ($\times2.86$), $p_{90}$ ($\times1.96$); IFEval ($\times2.05$); \texttt{e5} ($\times1.85$),
 \texttt{qwen4b} ($\times1.44$); ReduMIS ($\times1.39$)
 & Use deliberately for coverage reshaping\,+\,category diagnostic; else lower threshold \\
\addlinespace
Faithful but seed-unstable \newline ($\rho\ge0.70,W<0.90$; $7.4\%$)
 & Weak/small embeddings: \texttt{gist} ($\times2.42$), \texttt{bge} ($\times2.24$); $p_{90}$ ($\times2.56$);
 MMLU-Pro ($\times1.62$); Online-MIS ($\times1.34$)
 & Stronger/larger embedding, prefer ReduMIS, or average more seeds \\
\addlinespace
Doubly-degraded \newline ($\rho<0.70,W<0.90$; $7.7\%$)
 & $p_{95}$ ($\times3.65$), \texttt{gist} ($\times1.81$), GPQA ($\times1.53$); Online-MIS ($\times1.40$)
 & Avoid; \texttt{gist} @ $p_{90}$/$p_{95}$ off-target $93.8\%$. Cap threshold on small benchmarks \\
\bottomrule
\end{tabular}
\end{table}

\begin{figure}[t]
  \centering
  \includegraphics[width=\textwidth]{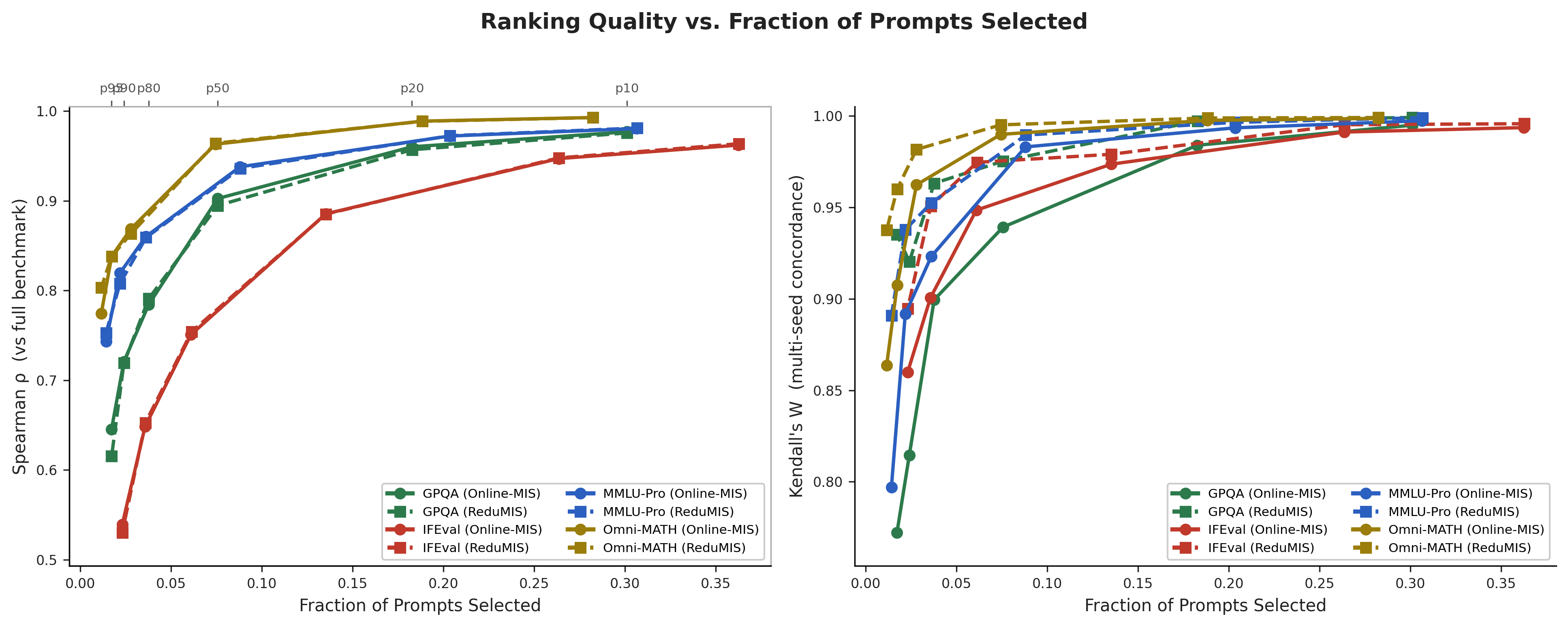}
  \caption{Agreement with the full benchmarks ($\rho$, left) and seed consistency ($W$, right) vs.\ the
  fraction of prompts selected, per benchmark and stochastic solver. Higher-percentile thresholds sit
  at the left (small fractions, aggressive pruning). Omni-MATH and MMLU-Pro preserve quality at small
  fractions; IFEval requires larger subsets.}
  \label{fig:reduction}
\end{figure}

\noindent\textbf{Reduction vs.\ fidelity.} Figure~\ref{fig:reduction} shows $\rho$ and $W$ as a function of the fraction of prompts selected, per benchmark and stochastic solver. All benchmarks show fidelity rising with subset size, but with benchmark-specific slopes: Omni-MATH and MMLU-Pro reach high $\rho$ with small fractions, whereas IFEval needs larger subsets for comparable fidelity. The knee of each curve sits near $p_{20}$, reinforcing it as a robust default.  Table~\ref{tab:threshold} from Appendix~\ref{app:distribution_bench} quantifies the central trade-off. As the percentile rises from $p_{10}$ to $p_{95}$, the density of the graph increases, pruning becomes more aggressive, and the mean subset shrinks from $\sim$231 to $\sim$11 prompts, increasing the reduction from $64$--$72\%$ to $98$--$99\%$. Fidelity moves in the opposite direction: $\rho$ falls monotonically from $0.978$ ($p_{10}$) to $0.672$ ($p_{95}$) and $W$ from $0.997$ to $0.869$. The practical consequence is favorable: at $p_{20}$, the subsets preserve the full ranking well ($\rho=0.967$, $W=0.994$) while still removing $74$--$82\%$ of the prompts. Mild pruning thus captures most of the efficiency benefit with little fidelity cost, while aggressive pruning ($p_{80}$+) buys marginal extra compression at a steep fidelity price. \textbf{Recommended configuration.} We recommend Qwen8b with standardized Euclidean distance, the
\texttt{p20} threshold, and ReduMIS for its balance between prompt
reduction and fidelity to the full-benchmark model ranking. In the
leave-one-benchmark-out analysis, configuration rankings were positively
associated in every fold. The recommended configuration achieved
Spearman's $\rho$ between $0.975$ and $0.998$ and Kendall's
$W \geq 0.999$ across all held-out benchmarks and both chunking
variants (Appendix~\ref{app:configuration-transfer}). 

\textbf{Baselines.} We compare maximal independent set (MIS) selection with sampling,
clustering, and response-based baselines. The sampling baselines use
subset sizes aggregated across configurations at each threshold. We
also evaluate random sampling at the exact subset size produced by the
recommended MIS configuration. Both sampling baselines are MIS-informed, since the set size is the result of the MIS optimization. One clustering baseline determines
the subset size from the cluster structure, while the other uses the
exact subset size produced by MIS to compare the methods at the same
level of compression. The response-based baselines, added as references, are anchor points
and item response theory (IRT).

\noindent\textbf{Comparison to size-matched sampling baselines.} We compare MIS with three baselines drawn at the
subset size MIS selected for each threshold\,$\times$\,benchmark cell (averaged over embedding, distance, solver,
and chunking): random, stratified (proportional to the prompt categories), and uniform-stratified (an equal share
per category via water-filling), each repeated five times and scored by Spearman $\rho$ and Kendall $W$ across draws. On the same $\rho=0.70$,
$W=0.90$ plane (Figure~\ref{fig:hypothesis_baselines}, Appendix~\ref{app:baseline}), all three
baselines stay almost entirely in the two faithful ($\rho{\ge}0.70$) quadrants but split by
repetition stability: all three are $100\%$ reliable-and-faithful at $p_{10}$/$p_{20}$, but by
$p_{90}$/$p_{95}$ their mean $W$ across repetitions falls to $0.73$--$0.77$. This $W$ measures variation
between independent draws, whereas for MIS it measures solver-seed variation on a fixed graph; it
therefore describes reproducibility rather than fidelity. Leaderboard-head overlap follows the same
threshold-driven decline: averaged over the five repetitions, top-$5$ Jaccard falls from $0.79$ at
$p_{10}$ to $0.31$ at $p_{95}$ (top-$10$: $0.84\to0.46$; Appendix~\ref{app:baseline}). Unlike the
baselines, MIS never sees category labels: any redistribution it produces is an emergent byproduct of
where semantic redundancy concentrates in embedding space. Full quadrant occupancy, the
per-benchmark/per-threshold comparison (Table~\ref{tab:full_baseline_comparison}), and
category-coverage figures are given in Appendix~\ref{app:baseline}. The mean absolute error (MAE) of model accuracy (Appendix~\ref{app:mae}) grows smoothly
with compression, from $3.5$ accuracy points at $p_{10}$/$p_{20}$ to $11.2$ at $p_{90}$/$p_{95}$
(Table~\ref{tab:mae_comparison}), near-identically across the four solvers; the error is unbiased (mean signed
shift $+0.8$ points) and shrinks with subset size (Fig.~\ref{fig:mae_panel}). Random subsets attain
marginally lower MAE at every threshold ($2.4$ vs.\ $3.5$ at $p_{10}$/$p_{20}$, $9.2$ vs.\ $11.2$ at
$p_{90}$/$p_{95}$), and stratified and uniform-stratified sampling sit between random and MIS ($10.2$ and $10.7$ at
$p_{90}$/$p_{95}$). MIS rebalances prompt categories, which can shift accuracy estimates in the same direction across models, increasing MAE while largely preserving the ranking. \textbf{Comparison to random sampling at the exact subset size produced by MIS configuration.} To compare a single, deployable subset with
sampling, we drew $1{,}000$ random, stratified and uniform-stratified subsets at exactly the size of the
recommended configuration (Table~\ref{tab:rev_main_equal}), and $1{,}000$ random subsets at the size of each of
the $2{,}558$ configurations (Appendix~\ref{app:rev_random}). At $p_{20}$ the recommended configuration is within
$0.007$ of the median random subset on all four benchmarks while removing $63$--$74\%$ of the prompts. Random
sampling is nevertheless slightly closer on average: the median MIS configuration lies at the $15.6$th percentile
of same-size random draws, although $93.2\%$ of configurations exceed the worst of $1{,}000$ draws. Where MIS
differs clearly is coverage: at equal size its subsets are more category-balanced than $96\%$, $92\%$ and over
$99\%$ of random subsets on GPQA, MMLU-Pro and Omni-MATH, without using labels. Its MAE is higher, because the
redistribution also changes the difficulty mix (e.g.\ a lower official difficulty rating on Omni-MATH at
$p_{80}$--$p_{95}$: $4.3$--$4.6$ vs.\ $4.9$).

\begin{table}[t]
\centering
\caption{Recommended MIS configuration (ReduMIS, \texttt{qwen8b}, standardized Euclidean, chunked) vs.\
random, stratified and uniform-stratified subsets of \emph{exactly the same size} at $p_{20}$
($1{,}000$ draws per baseline). $\rho$ = Spearman correlation with the full ranking of the $66$ LLMs
(random: mean and 5th percentile of the same draws as Table~\ref{tab:rev_equal_size}); MAE in accuracy points; coverage = normalized entropy
of the within-subset category shares ($1$ = all categories equally represented; full benchmark in
brackets); last column: share of same-size random subsets less balanced than MIS. IFEval has no
category labels.}
\label{tab:rev_main_equal}
\small
\resizebox{\textwidth}{!}{%
\begin{tabular}{lr cccc c cc cc c}
\toprule
 & & \multicolumn{4}{c}{$\rho$} & & \multicolumn{2}{c}{MAE} & \multicolumn{2}{c}{Coverage (entropy)} & \\
\cmidrule(lr){3-6}\cmidrule(lr){8-9}\cmidrule(lr){10-11}
Benchmark & $n$ & MIS & Random (5th pct) & Strat. & Unif.-strat. & & MIS & Random & MIS & Random [full] & Random less balanced \\
\midrule
GPQA      & 116 & 0.975 & 0.980 (0.972) & 0.980 & 0.979 & & 4.8 & 3.0 & 0.974 & 0.936 [0.942] & 96\% \\
IFEval    & 200 & 0.976 & 0.976 (0.961) & --    & --    & & 2.7 & 1.5 & --    & --            & --    \\
MMLU-Pro  & 284 & 0.985 & 0.992 (0.988) & 0.992 & 0.990 & & 3.6 & 1.7 & 0.978 & 0.969 [0.975] & 92\% \\
Omni-MATH & 332 & 0.996 & 0.996 (0.995) & 0.996 & 0.996 & & 4.8 & 1.6 & 0.810 & 0.775 [0.778] & $>$99\% \\
\bottomrule
\end{tabular}}
\end{table}
\textbf{Comparison to clustering baseline.} The clustering baseline applies $k$-means to unit-normalized embeddings
reduced to 50 dimensions by PCA, selects $k \in \{2,\ldots,20\}$ by
maximizing the silhouette score, and samples uniformly from each cluster
using the size of the smallest cluster as the per-cluster quota.
The baseline achieved high agreement with the full-benchmark ranking
(mean Spearman's $\rho = 0.993$) and retained at least 85\% of the instances
in 15 of 36 configurations. On all four benchmarks, MIS reached
$\rho \geq 0.95$ with substantially fewer instances than the median
clustering subset (Appendix~\ref{app:rev_selectors}). Clustering-based
selection is less parameter-free at the subset construction stage because
it requires separate rules for choosing the number of clusters, allocating
instances across clusters, and sampling within each cluster. For MIS,
once the distance metric, graph construction threshold, and solver are
specified, the solver determines both the subset size and which instances
to retain. \textbf{Comparison to MIS-informed clustering baselines.} We compare the recommended MIS configuration with $k$-center,
$k$-medoids, $k$-means, and SemDeDup, at the
subset sizes produced by MIS for each benchmark and threshold. MIS,
$k$-center, and SemDeDup showed similar ranking fidelity up to
\texttt{p90}, with mean Spearman's $\rho$ of $0.983$, $0.982$, and
$0.983$, respectively, at \texttt{p20}. At this threshold, $k$-medoids
achieved slightly higher ranking fidelity ($\rho = 0.987$) and $k$-means
similar fidelity ($\rho = 0.984$), both close to random sampling
($\rho = 0.986$), and both yielded lower mean absolute error than MIS. A practical benefit
of MIS is that, given the prompt similarity graph, it determines
both which prompts to retain and how many, and in most cases the stochastic solvers provide the exact optimal solution as the deterministic solvers. The matched-size baselines
use that subset size as an input
(Appendix~\ref{app:rev_selectors}). \textbf{Response-based baselines as references.} We also evaluated anchor points~\citep{vivek2024anchor} over 20
random splits of the 66 LLMs into 33 training and 33 held-out models. At \texttt{p20}, anchor points achieved a mean held-out $\rho$ of
$0.971$, compared with $0.979$ for MIS and $0.983$ for matched-size
random sampling on the same held-out models. Anchor points achieved
higher mean fidelity only at the most aggressive compression level,
\texttt{p95}, with $\rho=0.856$, compared with $0.833$ for MIS and
$0.845$ for random sampling. The advantage varied substantially
across model splits and was driven mainly by GPQA. In addition, on held-out LLMs, 2PL IRT estimated full-benchmark scores with a mean
absolute error of at most 3 percentage points using 7.7\% of the prompts,
and at most 2 percentage points using 20.9\%, with Kendall's
$\tau \geq 0.82$ at the stricter error target (Appendix~\ref{app:rev_selectors}).

\begin{table}[t]
\centering
\caption{Do embedding neighbors behave alike? Each discarded prompt is paired with its nearest retained
prompt (all ReduMIS configurations at $p_{20}$). Difficulty $r$: correlation of the share of the $66$ LLMs
solving each prompt, for neighbors vs.\ random pairs and random same-category pairs; annotation $r$: the same
for the benchmarks' own difficulty ratings (GPQA: writer rating; Omni-MATH: official rating). Agreement: share
of LLMs with the same outcome on both prompts, for neighbors vs.\ control pairs matched on category and
difficulty.}
\label{tab:rev_main_redundancy}
\small
\resizebox{\textwidth}{!}{%
\begin{tabular}{l ccc c cc c cc}
\toprule
 & \multicolumn{3}{c}{Difficulty $r$} & Annotation $r$ & \multicolumn{2}{c}{Same category} & & \multicolumn{2}{c}{Agreement} \\
\cmidrule(lr){2-4}\cmidrule(lr){6-7}\cmidrule(lr){9-10}
Benchmark & Neighbors & Random & Same-cat. & Neighbors & Neighbors & Random & & Neighbors & Matched control \\
\midrule
GPQA      & 0.17 & 0.00 & 0.11 & 0.22 & 95\% & 37\% & & 58.2\% & 57.7\% \\
IFEval    & 0.10 & $-0.01$ & --   & --   & --    & --    & & --      & --      \\
MMLU-Pro  & 0.10 & 0.00 & 0.09 & --   & 59\% & 8\%  & & 64.2\% & 64.0\% \\
Omni-MATH & 0.21 & 0.00 & 0.09 & 0.44 & 60\% & 23\% & & 62.5\% & 62.5\% \\
\bottomrule
\end{tabular}}
\end{table}

\begin{figure}[t]
  \centering
  \includegraphics[width=\textwidth]{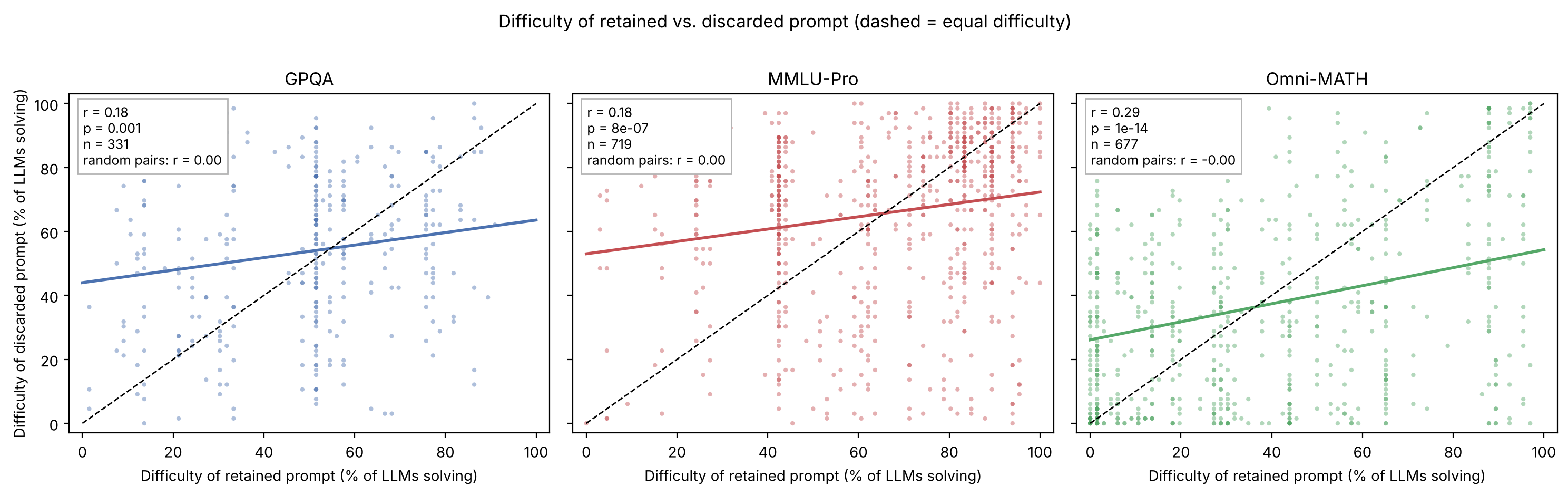}
  \caption{\textbf{Difficulty of retained vs.\ discarded prompt.} For each pair, the discarded prompt's
  difficulty (\% of the $66$ LLMs solving it) against its nearest retained neighbor's, per benchmark
  (\texttt{qwen8b}, standardized Euclidean, $p_{20}$, ReduMIS, no chunking); the dashed line marks equal
  difficulty. Pairs are positively correlated on all three benchmarks, whereas random prompt pairs show no
  correlation ($r=0.00$). Each panel reports Pearson $r$, its $p$-value, and $n$.}
  \label{fig:redundancy_difficulty}
\end{figure}

\noindent\textbf{Graph topology as a diagnostic.} The degree distribution of the full-benchmark graph
quantifies how unevenly prompts are distributed: benchmarks with high-degree hubs (many near-duplicate
prompts) are where MIS changes coverage most, so a practitioner can read this off the graph before
selecting. Clustering all $648$ graphs (benchmark $\times$ embedding $\times$ distance $\times$ threshold)
over nine graph-theoretic features yields a clean $2\times2$ grid (silhouette $0.503$): a density regime
set by the threshold and a size regime set by the benchmark, while embedding model, embedding size and
chunking leave no detectable signature and among the distances only \texttt{seuclidean} perturbs the
topology (Appendix~\ref{app:topology}). \textbf{Does embedding proximity imply evaluative redundancy?}
We paired each prompt discarded by ReduMIS with its nearest retained prompt and measured difficulty as the fraction of the 66 LLMs that solved it. Figure~\ref{fig:redundancy_difficulty} illustrates the relationship between paired difficulties for the recommended configuration. Across 631 configurations on four benchmarks, difficulty correlations were positive in 89\% of configurations. At $p_{20}$, mean correlations ranged from $r=0.10$ to $0.21$ across benchmarks, compared with $r=0.00$ for random pairs. In addition, the official Omni-MATH difficulty ratings (presented as difficulty levels by the design of the benchmark) also correlated between neighbors ($r=0.44$). On the three category-labeled benchmarks, embedding neighbors shared a category more often than random pairs. At $p_{20}$, their agreement in model outcomes was within 1.3 percentage points of controls matched by category and difficulty for every embedding. This comparison suggests that category and difficulty account for the observed behavioral similarity (Table~\ref{tab:rev_main_redundancy}). MIS thus reduces semantic redundancy by identifying prompts that tend to share a category and have similar difficulty, using prompt text alone (Appendix~\ref{app:redundancy}). \noindent\textbf{Category redistribution.} Figure~\ref{fig:coverage} grounds the redistribution effect in MMLU-Pro's ground-truth
categories rather than a synthetic proxy. Coverage is far from uniform: across $14$
categories a uniform split gives $7.1\%$ each, yet math and physics alone account for
$12.5\%$ and $11.8\%$, while history and computer science fall to $3.5\%$ and $3.4\%$
(panel~(a)). ReduMIS selection with \texttt{qwen8b}, standardized
Euclidean distance, and $p_{20}$ retains $281$ of $1{,}000$ prompts (panel~(b)) and moves the
distribution toward uniform: within the subset, math and physics shrink from $12.5\%$ and $11.8\%$ to
$6.4\%$, close to the uniform share of $7.1\%$, while biology and health grow ($5.9\%\to11.7\%$,
$5.6\%\to9.6\%$; panel~(c)). This is not a size effect: at exactly the same subset size, MIS subsets are
more category-balanced than the average random and stratified subset and than $92$--$99\%$ of individual random subsets (Table~\ref{tab:rev_main_equal};
Appendix~\ref{app:baseline}), and Appendix~\ref{app:coverage} finds the same pattern on GPQA and Omni-MATH.
This concretely measures \emph{what} changes when a subset's ranking diverges from the full one: prompt
mass is redistributed toward balanced coverage. We report this as a characterization. The evaluation goal determines whether to favor flatter coverage or retain the full-benchmark ranking.

\section{Discussion and Limitations}
\label{sec:discussion}
\noindent\textbf{Cost and when MIS pays off.} MIS selection is one-time preprocessing: embedding all four
benchmarks with the three large decoders (both chunking variants, $24$ runs) took $53$ minutes on $2\times$
NVIDIA L40S GPUs, and solving takes under one second (GREEDY) to $30$--$120$\,s (CPLEX), with ReduMIS matching
the CPLEX subset size for the recommended configuration (Table~\ref{tab:rev_solvers}). For a one-off evaluation, random sampling with a pre-defined set size is cheaper and slightly closer to the full ranking, but its subset size must be set in advance. For a curated benchmark reused across many models, the upfront cost of MIS is spread across evaluations, and the same canonical, reproducible subset can be reused with guaranteed non-redundancy in embedding space, a size determined automatically from the graph, and balanced coverage without category labels.

\textbf{Agentic and multi-turn benchmarks.} The framework extends to agentic benchmarks by treating
tasks or trajectories as nodes, provided they can be represented in a common feature space (e.g.\ actions, tool
calls and outcomes); agent-dependent trajectories would require task-level features or reference agents. Whether
proximity in such a space reflects evaluative redundancy is an empirical question, testable by whether the subset
preserves agent rankings. \textbf{Scope and scalability.} MIS is NP-hard, but solved \emph{once} as offline preprocessing,
after which the reduced benchmark is reused at no cost; solvers such as ReduMIS handle large sparse graphs
well. Benchmarks with tens of thousands of prompts may need approximate or streaming MIS, which we leave to
future work. Redundancy is defined in the chosen embedding space: it captures category and difficulty
structure but not prompt interchangeability (Section~\ref{sec:results}), and biases of the embedding carry
over to the selected subset.

\section{Conclusion}
\label{sec:conclusion}
We introduced an MIS framework for LLM benchmark compression, evaluated across four benchmarks, 66 LLMs, and 2{,}558 configurations. Selection is highly seed-consistent ($W \geq 0.90$ in 84.9\% of configurations), and agreement with the full ranking follows a controllable efficiency--fidelity trade-off: mild thresholds ($p_{10}$--$p_{20}$) preserve rankings ($\rho = 0.97$--$0.98$) while removing 64--82\% of prompts. 
Compared with random subsets drawn at the subset size determined by MIS, MIS is close in ranking fidelity at mild compression but not better on average, and diversity-oriented selectors in general share this behavior; its distinctive contributions are a reproducible subset, an automatic subset size result from the graph optimization, and label-free coverage redistribution that captures category and difficulty structure without model evaluations. We recommend ReduMIS with \texttt{qwen8b}/\texttt{seuclidean} at $p_{20}$ as a default when a reproducible, coverage-balanced subset is the goal.

\subsubsection*{Ethics Statement}
This work studies benchmark subset selection over publicly reported evaluation scores and involves no human subjects. We identify three risks. \textbf{(1)~Adversarial subset selection:} the configurability of the pipeline (embedding, distance, threshold, solver) could be exploited post hoc to favor a particular model; we recommend that benchmark curators fix and publicly disclose the full configuration before evaluation. \textbf{(2)~Embedding-bias amplification:} MIS redistributes coverage only as captured by the chosen embedding, so cultural, linguistic, or domain biases in the embedding may be amplified rather than reduced, especially for domain-specific or multilingual benchmarks. \textbf{(3)~Reduced capability coverage:} aggressive pruning may omit rare but important capabilities, which is particularly consequential for safety-critical evaluation; under-represented edge cases should not be discarded for the sake of coverage uniformity. We use only public leaderboard data and release no personal or sensitive information.

\subsubsection*{Reproducibility Statement}
The methodology is fully specified in Section~\ref{sec:method}, the experimental grid and hyperparameters in Section~\ref{sec:setup}, the embedding-generation and hardware details in Appendix~\ref{app:compute}, and the evaluated models in Appendix~\ref{app:models}. All benchmark scores are drawn from the public HELM Capabilities leaderboard, and all reported tables and figures are computed from the released per-run statistics; the additional analyses (same-size random draws, other selectors, control pairs) are reproduced by a single script with fixed random seeds. Anonymised source code and the derived per-run data are provided as supplementary material to reproduce every table and figure. Code available at: https://anonymous.4open.science/r/BenchCompress-B6BD/.

\subsubsection*{Use of AI Assistants}
AI assistance was used for grammar and language polishing, as well as auxiliary coding and implementation support that has been checked and verified by the authors; all scientific content, experimental design, and conclusions are the authors' own.

\input{main.bbl}
\appendix
\newpage

\appbarrier
\section{Positioning Among Benchmark-Reduction Methods}
\label{app:related_table}

\begin{table}[!htbp]
\centering
\caption{Where our method sits among benchmark-reduction approaches. Methods differ mainly in
the \emph{signal} they consume for selection and the \emph{granularity} at which they select; only
embedding-driven methods avoid needing prior model evaluations.}
\label{tab:related_appendix}
\small
\begin{tabular}{@{}p{5.3cm}lll@{}}
\toprule
\textbf{Approach (examples)} & \textbf{Granularity} & \textbf{Selection signal} & \textbf{Model evals?} \\
\midrule
IRT / anchor items~\citep{polo2024tinybenchmarks,vivek2024anchor,kipnis2025metabench} & sample  & model responses  & yes \\
Response-correlation selection~\citep{huang2025minilongbench}, hidden-state representations~\citep{zhang2026internal}, or anchor-LLM + intrinsic features~\citep{saranathan2025sublime} & sample  & model responses / hidden states  & yes \\
Score-based dataset selection~\citep{subramani2025simba}          & dataset & evaluation scores & yes \\
Task-similarity selection~\citep{surkov2024vygotsky}                                    & dataset & response patterns & yes \\
Diversity coresets~\citep{abbas2023semdedup}                           & sample & embeddings        & no \\
\textbf{MIS on prompt proximity graphs (ours)}                                                & sample  & embeddings        & no$^{\dagger}$ \\
\bottomrule
\end{tabular}
\\[3pt]
{\footnotesize $^{\dagger}$Model scores are used only to \emph{measure} ranking fidelity after selection, never to drive it.}
\end{table}

Table~\ref{tab:related_appendix} summarizes how representative benchmark-reduction methods differ
in the signal they consume and the granularity at which they operate, and locates our
embedding-only, evaluation-free MIS selection within that landscape.
Along the first axis, most prior work is \emph{response-driven}: IRT and anchor-point methods
\citep{polo2024tinybenchmarks,vivek2024anchor} fit item difficulty and discrimination to a matrix of
model responses; methods that instead learn from item-level statistics or model-internal signals ---
response-correlation matrices \citep{huang2025minilongbench}, hidden-state representations
\citep{zhang2026internal}, or a small set of anchor LLMs combined with intrinsic sample features
\citep{saranathan2025sublime} --- similarly identify the most informative items; score-based dataset selection
\citep{subramani2025simba} compares whole benchmarks using their raw evaluation-score matrices; and
task-similarity selection \citep{surkov2024vygotsky} instead compares them by the response patterns
they induce across models. All of these require running a pool of models on the full benchmark before a
subset can be chosen, and the resulting subset is tied to that pool. Diversity coresets such as SemDeDup~\citep{abbas2023semdedup} and our MIS approach instead use only
the prompts' embeddings, so the subset can be constructed once, offline, with no model evaluations,
and reused for any future model. SemDeDup has been applied to web-scale image and text training
datasets to remove examples based on semantic similarity, using cosine similarity between embeddings
combined with a clustering step that requires choosing the number of clusters and a rule for which
examples to retain from each group. Unlike clustering, which requires specifying the number of clusters and a rule for selecting examples, MIS instead determines the subset directly from the graph, with no cluster count or retention rule to fix in advance; SELECTOR~\citep{selector} presents a direct
comparison of MIS-style selection with clustering. Along the second axis, methods operate either at the level of individual samples (selecting
prompts within a benchmark) or at the level of whole datasets (selecting benchmarks within a suite);
ours is sample-level. What distinguishes MIS from other embedding-only selectors is that the
non-redundancy constraint is explicit and the subset size is a by-product of the graph rather than a
hyperparameter: coreset and clustering methods must fix a budget or cluster count in advance, whereas
the independent-set property guarantees no two retained prompts are within $\theta$ and lets the
benchmark's own redundancy structure determine how much can be removed. Model scores enter our
pipeline only afterward, to measure fidelity ($\rho$, $W$, MAE), never to drive selection.

\appbarrier
\section{Maximum Independent Set: Formulation and Algorithm}
\label{app:MIS_formulation}

Let $G=(V,E)$ be the undirected proximity graph,
with $V=\{v_1,\dots,v_N\}$ the prompt nodes and
$(v_i,v_j)\in E \iff d(e_i,e_j)\le\theta$, for a given distance $d$ and a threshold value $\theta \in \mathbb{R}^{+}$. A set $S\subseteq V$ is
\emph{independent} if it contains no edge,
\begin{equation}
  \forall\, v_i,v_j\in S,\ i\neq j:\quad (v_i,v_j)\notin E
  \;\Longleftrightarrow\; d(e_i,e_j)>\theta .
\end{equation}
The \emph{maximum independent set} (MIS) is an independent set of maximum
cardinality,
\begin{equation}
  S^\star \;=\; \argmax_{S\subseteq V}\ \lvert S\rvert
  \quad\text{s.t.}\quad (v_i,v_j)\notin E \ \ \forall\, v_i,v_j\in S .
  \label{eq:mis-def}
\end{equation}
Its cardinality $\alpha(G)=\lvert S^\star\rvert$ is the independence number
of $G$. Equivalently, $S^\star$ is a maximum clique in the complement graph
$\bar G$, and $V\setminus S^\star$ is a minimum vertex cover of $G$.

Introducing a binary indicator $x_i\in\{0,1\}$ for whether $v_i\in S$,
Eq.~\ref{eq:mis-def} is represented by the binary linear program
\begin{align}
  \max_{x\in\{0,1\}^N}\ & \sum_{i=1}^{N} x_i \label{eq:mis-ilp-obj}\\
  \text{s.t.}\quad & x_i + x_j \le 1 \qquad \forall\,(v_i,v_j)\in E.
  \label{eq:mis-ilp-con}
\end{align}
The edge constraint Eq.~\ref{eq:mis-ilp-con} forbids selecting both endpoints
of any edge, so any feasible $x$ encodes an independent set and the objective
counts its size. Solving Eqs.~\ref{eq:mis-ilp-obj}--\ref{eq:mis-ilp-con}
exactly yields $S^\star$.

MIS is NP-hard, and hard to approximate within $N^{1-\epsilon}$ for any
$\epsilon>0$ unless \textup{P}=\textup{NP}~\citep{hastad1999clique,zuckerman2007linear}. Exact
solution via Eqs.~\ref{eq:mis-ilp-obj}--\ref{eq:mis-ilp-con} is therefore
tractable only for moderate $N$ or sparse $G$; for larger or denser graphs we
use heuristic solvers: the greedy algorithm below, Online-MIS and ReduMIS.

When an exact solution is generally impractical to obtain,  we may use  various approximation techniques, and, among them, standard greedy algorithms take a special place due to their simplicity and wide applicability.  The common greedy
criterion for solving MIS problem 
repeatedly selects a minimum-degree vertex and deletes it together with its
neighbors (Algorithm~\ref{alg:mis-greedy}). Choosing the lowest-degree vertex removes as few candidates as possible at each step. The heuristic at the end returns a \emph{maximal} independent set (no vertex can be added without violating independence) and guarantees $\lvert S\rvert \ge \sum_{v\in V} 1/(\deg(v)+1)$,
although not necessarily the maximum $S^\star$.

\begin{algorithm}[t]
\caption{Minimum-degree greedy heuristic for  MIS}
\label{alg:mis-greedy}
\begin{algorithmic}[1]
\Require Graph $G=(V,E)$
\Ensure Independent set $S\subseteq V$
\State $S \gets \emptyset$
\State $H \gets G$ \Comment{working copy}
\While{$V(H)\neq\emptyset$}
  \State $v \gets \argmin_{u\in V(H)} \deg_H(u)$ \Comment{ties broken arbitrarily/by index}
  \State $S \gets S \cup \{v\}$
  \State remove $v$ and all its neighbors $N_H(v)$ from $H$
\EndWhile
\State \Return $S$
\end{algorithmic}
\end{algorithm}

The algorithm~\ref{alg:mis-greedy} is deterministic once the tie-breaking rule is
fixed; randomized tie-breaking (or randomized solvers generally) induces a
distribution over solutions, which motivates the across-seed concordance
analysis (Kendall's $W$).

\appbarrier
\section{Evaluated LLMs}
\label{app:models}

\begin{table}[!htbp]
  \centering
  \caption{Benchmark statistics. $|P|$ = prompts; $|M|$ = number of LLMs; Acc.\ min/max = accuracy range on the full benchmark.}
  \label{tab:benchmarks}
  \small
  \begin{tabular}{lrrrrl}
    \toprule
    \textbf{Dataset} & $|\mathbf{P}|$ & $|\mathbf{M}|$ & \textbf{Acc.\ min} & \textbf{Acc.\ max} & \textbf{Top-1 model} \\
    \midrule
    GPQA      & 446   & 66 & 0.168 & 0.803 & Gemini~3~Pro (thinking) \\
    IFEval    & 541   & 66 & 0.567 & 0.951 & Grok-3 Mini \\
    MMLU-Pro  & 1{,}000 & 66 & 0.169 & 0.903 & Gemini~3~Pro (thinking) \\
    Omni-MATH & 1{,}000 & 66 & 0.072 & 0.722 & GPT-5 Mini (thinking) \\
    \bottomrule
  \end{tabular}
\end{table}
Table~\ref{tab:benchmarks} summarizes the four benchmarks by prompt count, number of evaluated LLMs, full-benchmark accuracy range, and top-ranked model, showing that GPQA and IFEval are roughly half the size of MMLU-Pro and Omni-MATH and that IFEval has the narrowest accuracy spread.

\begin{table}[!htbp]
\centering
\caption{The 66 evaluated LLMs and their providers.}
\label{tab:models}
\resizebox{0.95\textwidth}{!}{
\begin{tabular}{rll@{\hspace{2em}}rll}
\toprule
\textbf{\#} & \textbf{Model Key} & \textbf{Provider} & \textbf{\#} & \textbf{Model Key} & \textbf{Provider}\\
\midrule
1 & \texttt{amazon\_nova\_lite} & Amazon & 34 & \texttt{gpt\_oss\_20b} & OpenAI \\
2 & \texttt{amazon\_nova\_micro} & Amazon & 35 & \texttt{grok\_3\_beta} & xAI \\
3 & \texttt{amazon\_nova\_premier} & Amazon & 36 & \texttt{grok\_3\_mini\_beta} & xAI \\
4 & \texttt{amazon\_nova\_pro} & Amazon & 37 & \texttt{grok\_4\_0709} & xAI \\
5 & \texttt{claude\_3\_5\_haiku\_20241022} & Anthropic & 38 & \texttt{ibm\_granite\_3\_3\_8b\_instruct} & IBM \\
6 & \texttt{claude\_3\_5\_sonnet\_20241022} & Anthropic & 39 & \texttt{ibm\_granite\_4\_0\_micro} & IBM \\
7 & \texttt{claude\_3\_7\_sonnet\_20250219} & Anthropic & 40 & \texttt{ibm\_granite\_4\_0\_small} & IBM \\
8 & \texttt{claude\_4\_5\_haiku\_20251001} & Anthropic & 41 & \texttt{kimi\_k2\_instruct} & Moonshot \\
9 & \texttt{claude\_4\_5\_sonnet\_20250929} & Anthropic & 42 & \texttt{llama\_3\_1\_instruct\_turbo\_405b} & Meta \\
10 & \texttt{claude\_4\_opus\_20250514} & Anthropic & 43 & \texttt{llama\_3\_1\_instruct\_turbo\_70b} & Meta \\
11 & \texttt{claude\_4\_opus\_..\_thinking} & Anthropic & 44 & \texttt{llama\_3\_1\_instruct\_turbo\_8b} & Meta \\
12 & \texttt{claude\_4\_sonnet\_20250514} & Anthropic & 45 & \texttt{llama\_4\_maverick\_17bx128e\_fp8} & Meta \\
13 & \texttt{claude\_4\_sonnet\_..\_thinking} & Anthropic & 46 & \texttt{llama\_4\_scout\_17bx16e\_instruct} & Meta \\
14 & \texttt{deepseek\_r1\_0528\_with\_thinking} & DeepSeek & 47 & \texttt{marin\_8b\_instruct} & Marin \\
15 & \texttt{deepseek\_v3} & DeepSeek & 48 & \texttt{mistral\_instruct\_v0\_3\_7b} & Mistral AI \\
16 & \texttt{gemini\_1\_5\_flash\_002} & Google & 49 & \texttt{mistral\_large\_2411} & Mistral AI \\
17 & \texttt{gemini\_2\_0\_flash} & Google & 50 & \texttt{mistral\_small\_3\_1\_2503} & Mistral AI \\
18 & \texttt{gemini\_2\_0\_flash\_lite\_preview} & Google & 51 & \texttt{mixtral\_instruct\_8x7b} & Mistral AI \\
19 & \texttt{gemini\_2\_5\_flash\_04\_17\_preview} & Google & 52 & \texttt{o3\_2025\_04\_16} & OpenAI \\
20 & \texttt{gemini\_2\_5\_flash\_lite} & Google & 53 & \texttt{o4\_mini\_2025\_04\_16} & OpenAI \\
21 & \texttt{gemini\_2\_5\_pro\_03\_25\_preview} & Google & 54 & \texttt{olmo\_2\_13b\_instruct\_nov\_2024} & AllenAI \\
22 & \texttt{gemini\_3\_pro\_preview\_thinking} & Google & 55 & \texttt{olmo\_2\_32b\_instruct\_mar\_2025} & AllenAI \\
23 & \texttt{glm\_4\_5\_air\_fp8\_with\_thinking} & Zhipu AI & 56 & \texttt{olmo\_2\_7b\_instruct\_nov\_2024} & AllenAI \\
24 & \texttt{gpt\_4\_1\_2025\_04\_14} & OpenAI & 57 & \texttt{olmoe\_1b\_7b\_instruct\_jan\_2025} & AllenAI \\
25 & \texttt{gpt\_4\_1\_mini\_2025\_04\_14} & OpenAI & 58 & \texttt{palmyra\_fin} & Writer \\
26 & \texttt{gpt\_4\_1\_nano\_2025\_04\_14} & OpenAI & 59 & \texttt{palmyra\_med} & Writer \\
27 & \texttt{gpt\_4o\_2024\_11\_20} & OpenAI & 60 & \texttt{palmyra\_x5} & Writer \\
28 & \texttt{gpt\_4o\_mini\_2024\_07\_18} & OpenAI & 61 & \texttt{palmyra\_x\_004} & Writer \\
29 & \texttt{gpt\_5\_1\_2025\_11\_13} & OpenAI & 62 & \texttt{qwen2\_5\_instruct\_turbo\_72b} & Alibaba \\
30 & \texttt{gpt\_5\_2025\_08\_07\_with\_thinking} & OpenAI & 63 & \texttt{qwen2\_5\_instruct\_turbo\_7b} & Alibaba \\
31 & \texttt{gpt\_5\_mini\_..\_with\_thinking} & OpenAI & 64 & \texttt{qwen3\_235b\_a22b\_fp8\_thinking} & Alibaba \\
32 & \texttt{gpt\_5\_nano\_..\_with\_thinking} & OpenAI & 65 & \texttt{qwen3\_235b\_a22b\_instruct\_2507} & Alibaba \\
33 & \texttt{gpt\_oss\_120b\_with\_thinking} & OpenAI & 66 & \texttt{qwen3\_next\_80b\_a3b\_thinking} & Alibaba \\
\bottomrule
\end{tabular}
}
\end{table}
Table~\ref{tab:models} lists all $66$ LLMs evaluated, drawn from the HELM Capabilities leaderboard.
\appbarrier
\section{How Benchmarks Differ in Subset Characteristics}
\label{app:distribution_bench}

Table~\ref{tab:jaccard} gives the full per-quadrant top-$N$ Jaccard breakdown summarized in
Section~\ref{sec:results}.

\begin{table}[!htbp]
  \centering
  \caption{Top-$N$ Jaccard overlap between the full-benchmark and subset leaderboards, by quadrant
  of Figure~\ref{fig:hypothesis} ($\rho=0.70$, $W=0.90$). Overlap is the fraction of shared models
  among the top $5$ (or top $10$); ``$\ge0.5$'' is the share of configurations preserving at least
  half of the head.}
  \label{tab:jaccard}
  \small
  \begin{tabular}{lrrrr}
    \toprule
    Quadrant & top-5 (mean) & top-5 $\ge0.5$ & top-10 (mean) & top-10 $\ge0.5$ \\
    \midrule
    Reliable \& faithful ($\rho{\ge}.70,W{\ge}.90$)    & 0.561 & 59\% & 0.678 & 83\% \\
    Reliable but reshaped ($\rho{<}.70,W{\ge}.90$)     & 0.336 & 22\% & 0.447 & 42\% \\
    Faithful but seed-unstable ($\rho{\ge}.70,W{<}.90$)& 0.254 &  5\% & 0.373 & 15\% \\
    Doubly-degraded ($\rho{<}.70,W{<}.90$)             & 0.163 &  4\% & 0.240 &  1\% \\
    \bottomrule
  \end{tabular}
\end{table}

Table~\ref{tab:distribution_bench} partitions the stochastic configurations of each benchmark across the four quadrants of Section~\ref{sec:results} ($\rho=0.70$, $W=0.90$; Figure~\ref{fig:hypothesis}), and Table~\ref{tab:embedding_breakdown} does the same by embedding model.

\begin{table}[!htbp]
\centering
\caption{Distribution of stochastic subset configurations across the four quadrants of Section~\ref{sec:results} ($\rho=0.70$, $W=0.90$; Figure~\ref{fig:hypothesis}).}
\label{tab:distribution_bench}
\small
\begin{tabular}{l p{2.1cm} p{2.1cm} p{2.1cm} p{1.8cm}}
\toprule
\textbf{Benchmark} & 
\textbf{Reliable \& faithful} ($\rho{\ge}.70,W{\ge}.90$) & 
\textbf{Reliable but reshaped} ($\rho{<}.70,W{\ge}.90$) & 
\textbf{Faithful but seed-unstable} ($\rho{\ge}.70,W{<}.90$) & 
\textbf{Doubly-degraded} ($\rho{<}.70,W{<}.90$) \\
\midrule
GPQA      & 67.2\% & 11.4\% &  9.7\% & 11.7\% \\
IFEval    & 73.8\% & 15.4\% &  0.6\% & 10.2\% \\
MMLU-Pro  & 80.6\% &  0.9\% & 12.0\% &  6.5\% \\
Omni-MATH & 86.7\% &  2.8\% &  7.7\% &  2.8\% \\
\bottomrule
\end{tabular}
\end{table}

\begin{table}[!htbp]
\centering
\caption{Per-embedding-model breakdown of stochastic configuration outcomes across the four quadrants of Section~\ref{sec:results} ($\rho=0.70$, $W=0.90$).}
\label{tab:embedding_breakdown}
\small
\begin{tabular}{l p{2.3cm} p{2.3cm} p{2.3cm} p{2.0cm}}
\toprule
\textbf{Embedding} & 
\textbf{Reliable \& faithful} & 
\textbf{Reliable but reshaped} & 
\textbf{Faithful but seed-unstable} & 
\textbf{Doubly-degraded} \\
\midrule
\emph{BGE}       & 70.1\% & 10.4\% & 16.7\% &  2.8\% \\
\emph{E5}        & 71.5\% & 13.9\% &  4.2\% & 10.4\% \\
\emph{GIST}      & 60.4\% &  7.6\% & 18.1\% & 13.9\% \\
\emph{Nemotron}  & 88.5\% &  2.5\% &  1.4\% &  7.6\% \\
\emph{Qwen-4B}   & 75.0\% & 10.8\% &  6.3\% &  7.8\% \\
\emph{Qwen-8B}   & 83.8\% &  4.6\% &  6.0\% &  5.6\% \\
\bottomrule
\end{tabular}
\end{table}

\noindent\textbf{Per-dataset analysis.} Table~\ref{tab:datasets} reveals clear variation across benchmarks for the stochastic solvers, aggregated across embeddings, distances, and thresholds. Omni-MATH and MMLU-Pro preserve rankings best ($\rho\approx0.88$--$0.91$, $W\approx0.93$--$0.98$), while GPQA and IFEval are hardest ($\rho\approx0.79$--$0.83$), reflecting their smaller size and greater score discretizations. Even on the hardest benchmarks, ReduMIS keeps high seed concordance ($W\ge0.964$), so subsets remain internally reliable even where they diverge more from the full ranking.

\begin{table}[!htbp]
  \centering
  \caption{Per-dataset, per-algorithm results for the stochastic solvers, aggregated across embeddings, distances, and thresholds. $\rho$ = mean Spearman correlation with the full benchmark; $\rho{\ge}0.90$ and $0.70{\le}\rho{<}0.90$ = share of seed runs falling in the ``very strong'' and ``strong'' Schober bands, respectively; $W$ = mean Kendall concordance; $\tau{\ge}0.9$ = fraction of seed pairs with Kendall $\tau\ge0.9$; $|S|$ = mean subset size.}
  \label{tab:datasets}
  \small
  \resizebox{\textwidth}{!}{%
  \begin{tabular}{llrrrrrrr}
    \toprule
    Dataset & Algorithm & $\rho$ (mean) & $\rho_{\mathrm{std}}$ & $\rho{\ge}0.90$ (\%) & $0.70{\le}\rho{<}0.90$ (\%) & $W$ & $\tau{\ge}0.9$ (\%) & $|S|$ \\
    \midrule
    GPQA      & Online-MIS & 0.826 & 0.146 & 43.2 & 35.8 & 0.8973 & 40.5 & 45.4 \\
    GPQA      & ReduMIS    & 0.820 & 0.164 & 42.6 & 33.9 & 0.9640 & 89.8 & 45.4 \\
    \midrule
    IFEval    & Online-MIS & 0.789 & 0.196 & 39.3 & 34.8 & 0.9445 & 62.6 & 79.4 \\
    IFEval    & ReduMIS    & 0.789 & 0.205 & 40.0 & 34.5 & 0.9649 & 86.5 & 79.5 \\
    \midrule
    MMLU-Pro  & Online-MIS & 0.886 & 0.107 & 57.8 & 35.0 & 0.9309 & 65.2 & 111.6 \\
    MMLU-Pro  & ReduMIS    & 0.885 & 0.107 & 59.4 & 33.5 & 0.9610 & 86.5 & 111.7 \\
    \midrule
    Omni-MATH & Online-MIS & 0.904 & 0.110 & 64.0 & 29.0 & 0.9532 & 64.3 & 100.3 \\
    Omni-MATH & ReduMIS    & 0.908 & 0.104 & 64.6 & 30.5 & 0.9786 & 86.0 & 100.3 \\
    \bottomrule
  \end{tabular}%
  }
\end{table}

\noindent\textbf{Effect of threshold.} Table~\ref{tab:threshold} gives the full per-threshold breakdown underlying the efficiency--fidelity trade-off discussed in Section~\ref{sec:results}.

\begin{table}[!htbp]
  \centering
  \caption{Effect of the distance threshold, averaged over all solvers and all embedding/distance/benchmark combinations. $\rho$ = mean Spearman correlation with the full benchmark; $W$ = Kendall concordance; $|S|$ = mean subset size; Reduction = relative reduction vs.\ the full benchmark (min--max across benchmarks). Higher percentiles compress more but preserve rankings less.}
  \label{tab:threshold}
  \small
  \begin{tabular}{lrrrrrr}
    \toprule
    Threshold & $\rho$ & $\rho_{\mathrm{std}}$ & $W$ & $W_{\mathrm{std}}$ & $|S|$ & Reduction (\%) \\
    \midrule
    p10 & 0.978 & 0.015 & 0.9971 & 0.0036 & 230.7 & 63.9--71.9 \\
    p20 & 0.967 & 0.024 & 0.9943 & 0.0068 & 154.1 & 73.8--81.7 \\
    p50 & 0.920 & 0.057 & 0.9788 & 0.0278 & 67.7  & 86.6--92.6 \\
    p80 & 0.815 & 0.114 & 0.9508 & 0.0575 & 28.2  & 93.9--97.2 \\
    p90 & 0.750 & 0.142 & 0.9115 & 0.1014 & 17.1  & 96.5--98.3 \\
    p95 & 0.672 & 0.184 & 0.8690 & 0.1389 & 11.3  & 97.7--98.9 \\
    \bottomrule
  \end{tabular}
\end{table}

\begin{figure}[!htbp]
  \centering
  \includegraphics[width=\textwidth]{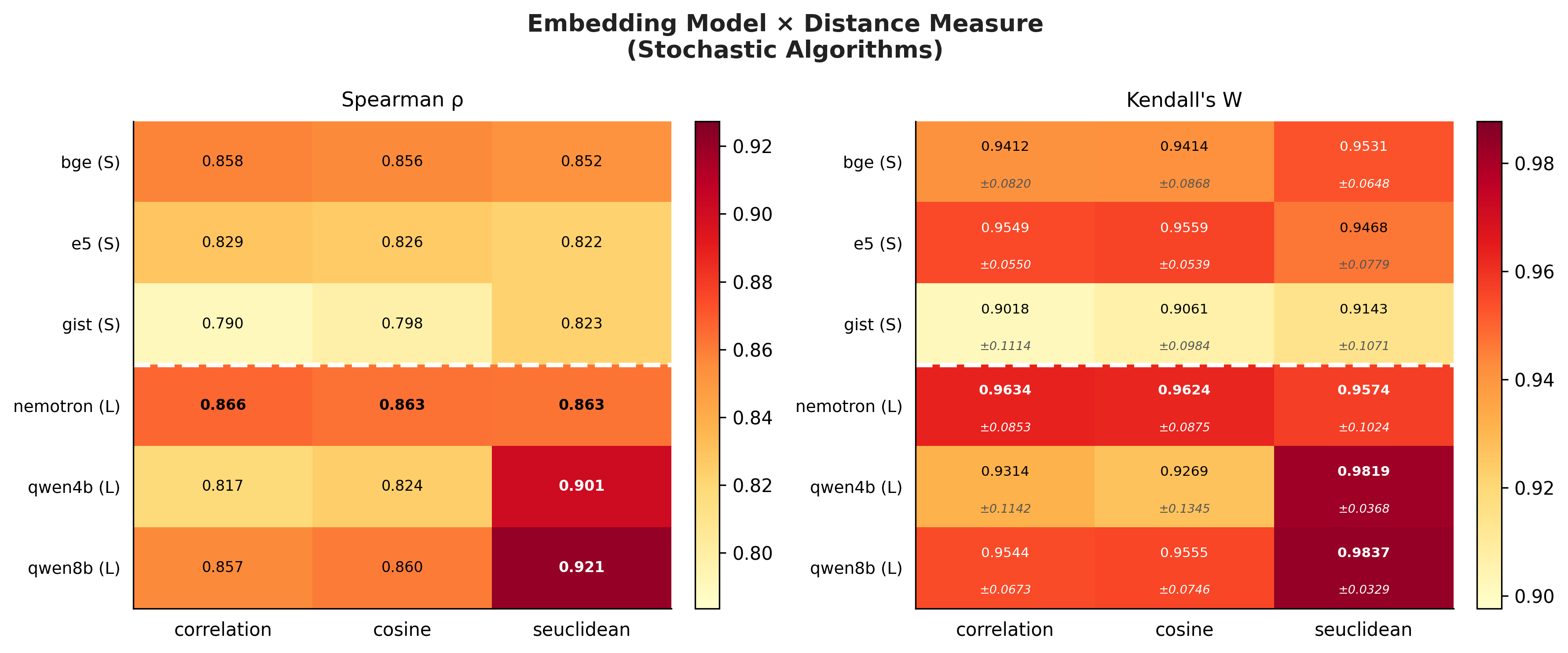}
  \caption{Spearman $\rho$ (left) and Kendall $W$ (right) by embedding model $\times$ distance measure
  (stochastic solvers). Standardized-Euclidean distance is markedly stronger for the large decoder
  models (\texttt{qwen4b}, \texttt{qwen8b}); \texttt{qwen8b}+seuclidean is best on both metrics.}
  \label{fig:heatmaps}
\end{figure}

\noindent\textbf{Embedding model and distance measure.} Table~\ref{tab:embeddings} and Figure~\ref{fig:heatmaps} compare embeddings and distances for the stochastic solvers. \texttt{qwen8b} with standardized Euclidean distance is the best configuration ($\rho=0.921$, $W=0.984$), closely followed by \texttt{qwen4b}+seuclidean ($\rho=0.901$, $W=0.982$); for these decoder models the standardized-Euclidean measure is clearly stronger than cosine or correlation. Among small encoders, \emph{e5} and \emph{bge} are the most stable ($W\approx0.94$--$0.96$), while \emph{gist} is the weakest overall. Large decoder embeddings give the highest and most consistent scores, but a small encoder (\emph{e5} or \emph{bge}) remains a reasonable choice when GPU memory is limited.

\begin{table}[!htbp]
  \centering
  \caption{Mean Spearman $\rho$ and Kendall $W$ by embedding model and distance measure (stochastic solvers). S/L denote small encoder-based and large decoder-based models. \textbf{Bold}: best $W$.}
  \label{tab:embeddings}
  \small
  \begin{tabular}{llrr}
    \toprule
    Model & Measure & $\rho$ & $W$ \\
    \midrule
    bge (S) & correlation & 0.858 & 0.9412 \\
    bge (S) & cosine      & 0.856 & 0.9414 \\
    bge (S) & seuclidean  & 0.852 & 0.9531 \\
    \midrule
    e5 (S) & correlation  & 0.829 & 0.9549 \\
    e5 (S) & cosine       & 0.826 & 0.9559 \\
    e5 (S) & seuclidean   & 0.822 & 0.9468 \\
    \midrule
    gist (S) & correlation & 0.790 & 0.9018 \\
    gist (S) & cosine      & 0.798 & 0.9061 \\
    gist (S) & seuclidean  & 0.823 & 0.9143 \\
    \midrule
    nemotron (L) & correlation & 0.866 & 0.9634 \\
    nemotron (L) & cosine      & 0.863 & 0.9624 \\
    nemotron (L) & seuclidean  & 0.863 & 0.9574 \\
    \midrule
    qwen4b (L) & correlation & 0.817 & 0.9314 \\
    qwen4b (L) & cosine      & 0.824 & 0.9269 \\
    qwen4b (L) & seuclidean  & 0.901 & 0.9819 \\
    \midrule
    qwen8b (L) & correlation & 0.857 & 0.9544 \\
    qwen8b (L) & cosine      & 0.860 & 0.9555 \\
    qwen8b (L) & seuclidean  & 0.921 & \textbf{0.9837} \\
    \bottomrule
  \end{tabular}
\end{table}

\noindent\textbf{Overall performance by algorithm.} Table~\ref{tab:algorithms} shows that all four solvers achieve nearly identical Spearman correlation with the full-benchmark ordering ($\rho\approx0.84$--$0.85$, aggregated across all benchmarks, embeddings, distances, and thresholds), so global ranking structure is largely invariant to solver choice. Among stochastic solvers, ReduMIS attains the strongest and most stable seed concordance ($W=0.967$ vs.\ $0.932$ for Online-MIS) with much smaller variance, making it the recommended solver.

\noindent\textbf{Tie-corrected concordance.} Kendall's $W$ in the main text is computed without tie correction, which slightly underestimates concordance when rankings contain ties, as is common for very small subsets (identical rankings with ties give $W$ slightly below $1$, e.g.\ $0.9993$). Table~\ref{tab:rev_ties} reports the effect of the standard tie correction. It can only increase $W$: the share of stochastic configurations with $W\ge0.90$ rises from $84.9\%$ to $89.1\%$, and $54$ of the $1{,}262$ configurations move from a seed-unstable to the corresponding seed-stable quadrant; among the $60$ baseline configurations, one changes quadrant. None of the conclusions change.

\begin{table}[!htbp]
\centering
\caption{Effect of the tie correction on Kendall's $W$ (stochastic MIS configurations, $n=1{,}262$).}
\label{tab:rev_ties}
\small
\begin{tabular}{lrr}
\toprule
Quantity & Without tie correction & With tie correction \\
\midrule
Configurations with $W\ge0.90$ & 84.9\% & 89.1\% \\
Mean $W$ & 0.950 & 0.967 \\
Reliable \& faithful & 77.3\% & 79.2\% \\
Reliable but reshaped & 7.5\% & 9.9\% \\
Faithful but seed-unstable & 7.4\% & 5.5\% \\
Doubly-degraded & 7.7\% & 5.3\% \\
\bottomrule
\end{tabular}
\end{table}

\begin{table}[!htbp]
  \centering
  \caption{Per-algorithm summary, aggregated across benchmarks, embeddings, distances, and thresholds. $\rho$ = mean Spearman correlation with the full-benchmark ranking; $\rho{\ge}0.95$ = fraction of configurations with high rank agreement; $W$ = mean Kendall concordance across seeds; $W_{\mathrm{std}}$ = its standard deviation; $|S|$ = mean selected subset size. $W$ is undefined (---) for the deterministic solvers CPLEX and GREEDY. \textbf{Bold}: best stochastic concordance.}
  \label{tab:algorithms}
  \small
  \begin{tabular}{lrrrrr}
    \toprule
    Algorithm & $\rho$ (mean) & $\rho{\ge}0.95$ (\%) & $W$ & $W_{\mathrm{std}}$ & $|S|$ \\
    \midrule
    CPLEX      & 0.851 & 38.1 & ---             & ---             & 84.8 \\
    GREEDY     & 0.845 & 36.4 & ---             & ---             & 82.5 \\
    Online-MIS & 0.852 & 37.1 & 0.9324          & 0.1046          & 85.2 \\
    ReduMIS    & 0.851 & 36.9 & \textbf{0.9672} & \textbf{0.0645} & 85.3 \\
    \bottomrule
  \end{tabular}
\end{table}

\section{Selection and Cross-Benchmark Transfer of the Recommended Configuration}
\label{app:configuration-transfer}

\paragraph{Configuration selection.}
We evaluated 216 configurations spanning embedding models, distance metrics,
thresholds, and two stochastic maximal independent set (MIS) solvers,
Online-MIS and ReduMIS. For each configuration, we averaged Spearman's
$\rho$ and Kendall's $W$ across the four benchmarks and both chunking
variants. Qwen8b with standardized Euclidean distance, the \texttt{p20}
threshold, and ReduMIS ranked third overall and first among \texttt{p20}
configurations. It ranked fifth by mean $\rho$ ($0.984$) and fourth by
mean $W$ ($0.9997$). The two configurations ranked higher overall used
the milder \texttt{p10} threshold, retained more prompts, and improved
mean $\rho$ by at most $0.002$. We therefore selected \texttt{p20} for
its balance between prompt reduction and fidelity to the full-benchmark
model ranking. With Online-MIS, the same embedding, distance metric, and
threshold achieved similar ranking fidelity (mean $\rho = 0.983$) and
lower agreement across solver seeds (mean $W = 0.997$, ranking 30th by
$W$).

\paragraph{Leave-one-benchmark-out analysis.}
We evaluated 290 stochastic configurations with results on all four
benchmarks, treating the chunked and unchunked variants as separate
configurations. In each fold, we ranked configurations separately by
mean $\rho$ and mean $W$ across the three retained benchmarks. For each
metric, we used Kendall's $\tau_b$ to compare the resulting ranking with
the corresponding configuration ranking on the held-out benchmark.
Agreement was positive in all four folds (all $p < 10^{-29}$).
As shown in Table~\ref{tab:configuration-transfer}, Kendall's $\tau_b$
ranged from $0.603$ to $0.757$ for rankings by $\rho$ and from $0.453$
to $0.689$ for rankings by $W$. Agreement for rankings by $W$ was lowest
on IFEval.

\begin{table}[t]
  \centering
  \caption{Agreement between configuration rankings computed on the
  three retained benchmarks and on the held-out benchmark.}
  \label{tab:configuration-transfer}
  \begin{tabular}{lcc}
    \hline
    \textbf{Held-out benchmark} &
    \textbf{$\tau_b$ for rankings by $\rho$} &
    \textbf{$\tau_b$ for rankings by $W$} \\
    \hline
    GPQA      & 0.603 & 0.592 \\
    IFEval    & 0.621 & 0.453 \\
    MMLU-Pro  & 0.714 & 0.689 \\
    Omni-MATH & 0.757 & 0.677 \\
    \hline
  \end{tabular}
\end{table}

\paragraph{Performance of the recommended configuration.}
Across the four folds, the two chunking variants of the recommended
configuration ranked 10th--16th by mean $\rho$ and 5th--11th by mean
$W$ on the three retained benchmarks. On the held-out benchmarks, they
ranked 5th--16th by $\rho$ on IFEval, MMLU-Pro, and Omni-MATH (top 6\%)
and 36th--37th on GPQA (top 13\%). By $W$, they ranked 1st--17th on
every held-out benchmark (top 6\%). Across the four held-out benchmarks
and both chunking variants, the recommended configuration achieved
$\rho$ values between $0.975$ and $0.998$ and $W \geq 0.999$.
These results indicate that configuration rankings partially transfer
across benchmarks and that the recommended configuration consistently
achieves high fidelity to the full-benchmark model ranking and high
agreement across solver seeds.

\appbarrier
\section{Category Coverage on Additional Benchmarks}
\label{app:coverage}
\begin{figure}[!htbp]
  \centering
  \includegraphics[width=\textwidth]{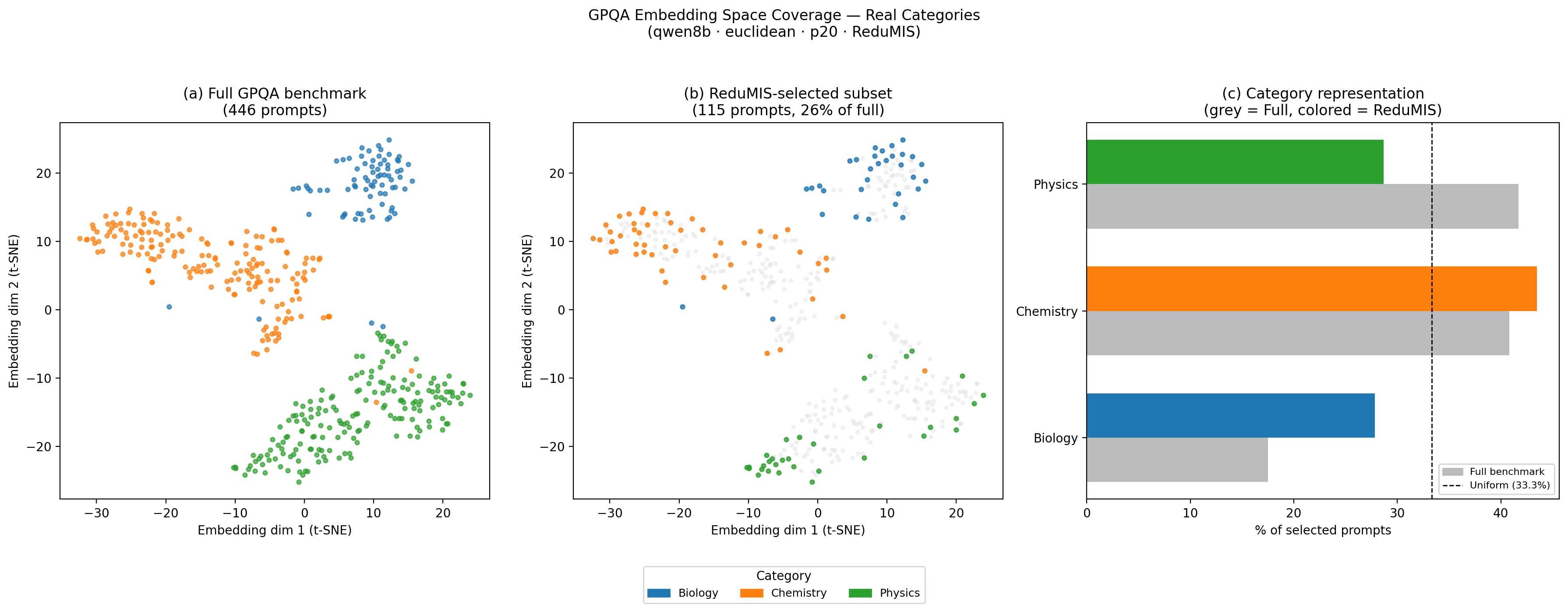}
  \caption{Real-category embedding-space coverage for GPQA (\texttt{qwen8b}, standardized Euclidean, $p_{20}$, ReduMIS, no chunking; $115$ of $446$ prompts). Within the subset, Biology rises from $17.5\%$ to $27.8\%$ and Physics falls from $41.7\%$ to $28.7\%$, toward the uniform share of $33.3\%$.}
  \label{fig:coverage_gpqa}
\end{figure}

\begin{figure}[!htbp]
  \centering
  \includegraphics[width=\textwidth]{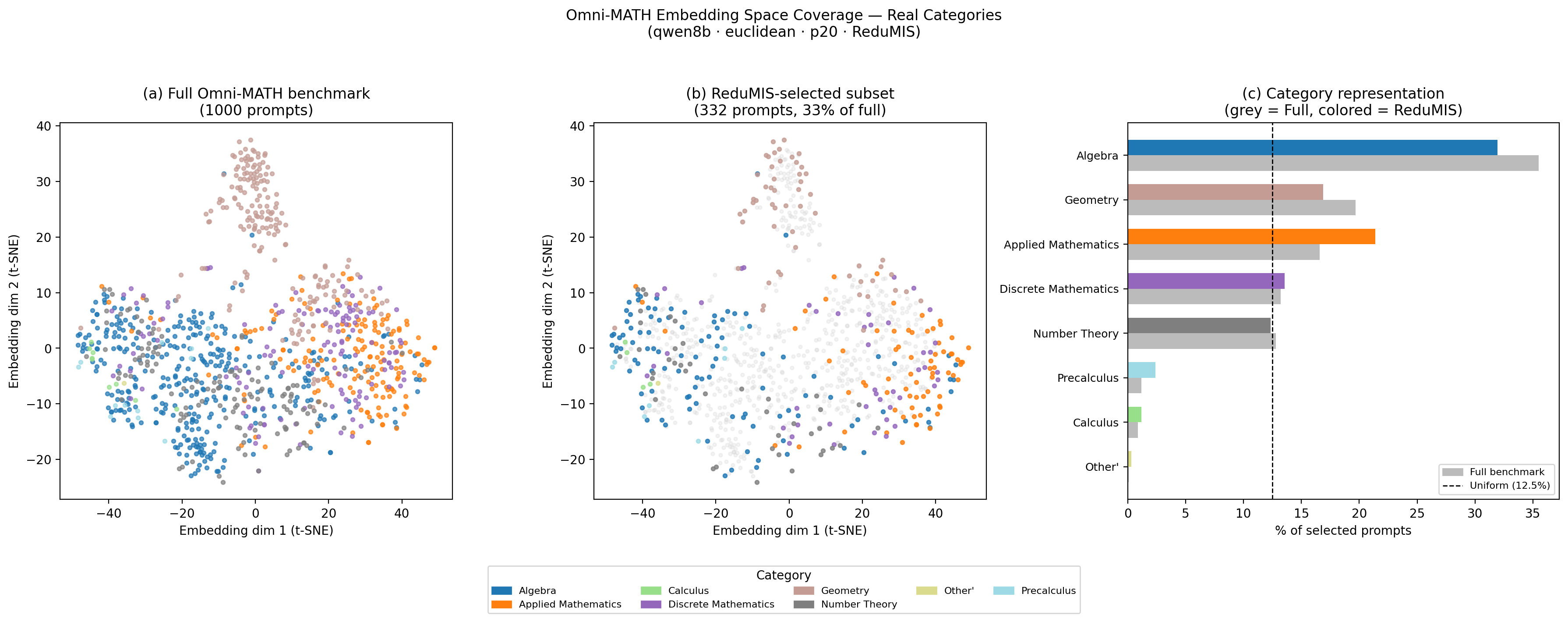}
  \caption{Real-category embedding-space coverage for Omni-MATH (\texttt{qwen8b}, standardized Euclidean, $p_{20}$, ReduMIS, chunked; $332$ of $1{,}000$ prompts). Algebra falls from $35.5\%$ to $31.9\%$ and Applied Mathematics rises from $16.6\%$ to $21.4\%$; the redistribution is milder than on GPQA and MMLU-Pro.}
  \label{fig:coverage_omni}
\end{figure}

Figures~\ref{fig:coverage_gpqa} and~\ref{fig:coverage_omni} repeat the real-category coverage analysis of Figure~\ref{fig:coverage} for GPQA and Omni-MATH.

\begin{figure}[!htbp]
  \centering
  \includegraphics[width=\textwidth]{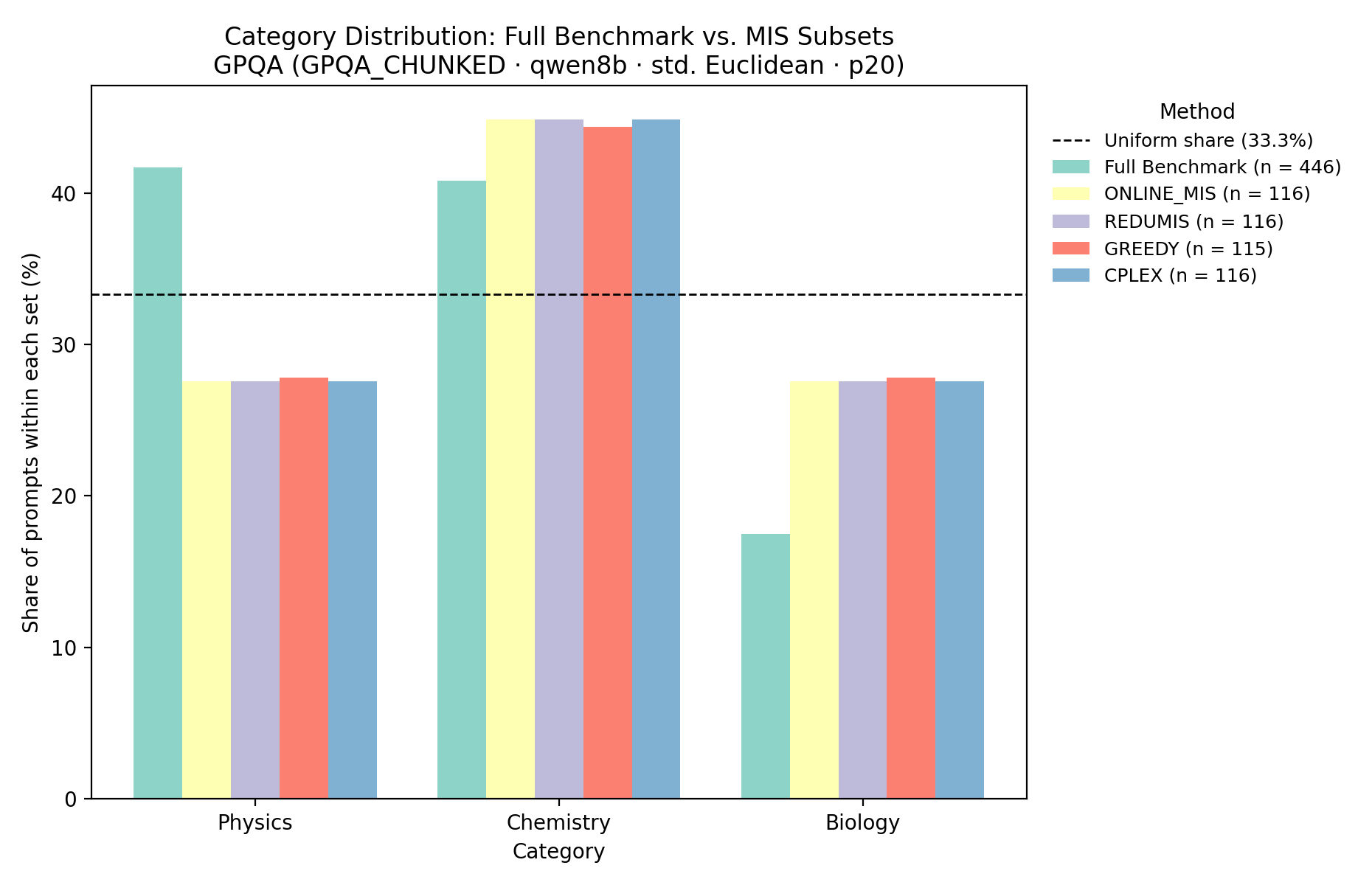}
  \caption{Category shares on GPQA (\texttt{qwen8b}, standardized Euclidean, $p_{20}$, chunked): full benchmark vs.\ the four MIS solvers, as shares within each set (dashed line: uniform share, $33.3\%$). All solvers select nearly identical compositions ($115$--$116$ prompts): Biology rises from $17.5\%$ to $27.6\%$ and Physics falls from $41.7\%$ to $27.6\%$.}
  \label{fig:catdist_gpqa}
\end{figure}

\begin{figure}[!htbp]
  \centering
  \includegraphics[width=\textwidth]{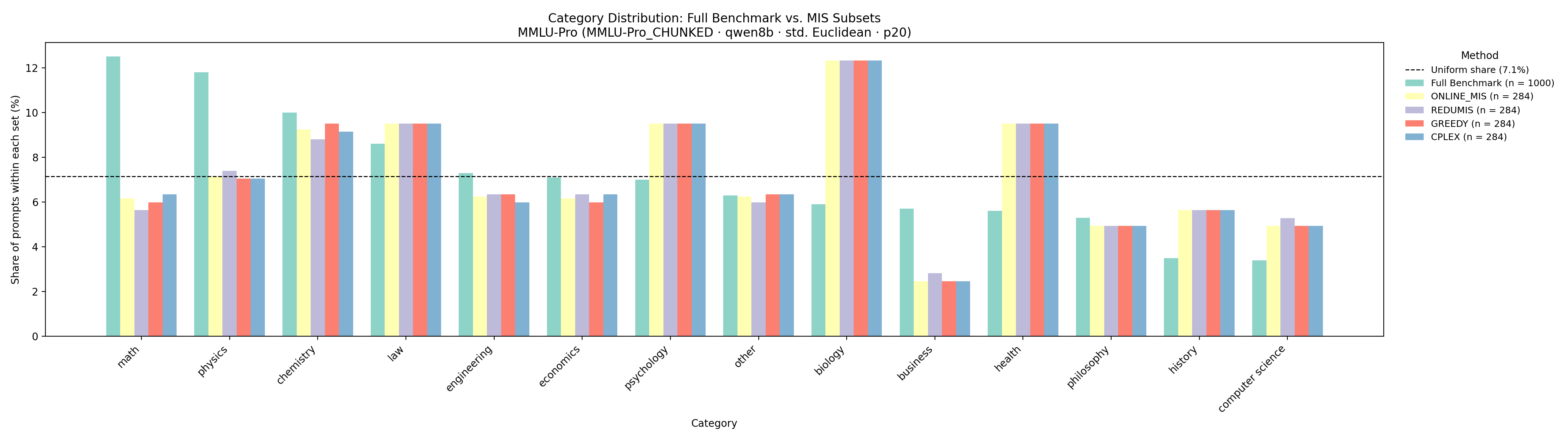}
  \caption{Category shares on MMLU-Pro (\texttt{qwen8b}, standardized Euclidean, $p_{20}$, chunked): full
  benchmark vs.\ the four MIS solvers, as shares within each set (dashed line: uniform share, $7.1\%$).
  All solvers select $284$ prompts with nearly identical compositions: math and physics fall from $12.5\%$
  and $11.8\%$ to $5.6$--$6.3\%$ and $7.0$--$7.4\%$, while biology ($5.9\%\to12.3\%$), health
  ($5.6\%\to9.5\%$) and psychology ($7.0\%\to9.5\%$) increase; business decreases ($5.7\%\to2.5$--$2.8\%$).}
  \label{fig:catdist_mmlu}
\end{figure}

\begin{figure}[!htbp]
  \centering
  \includegraphics[width=\textwidth]{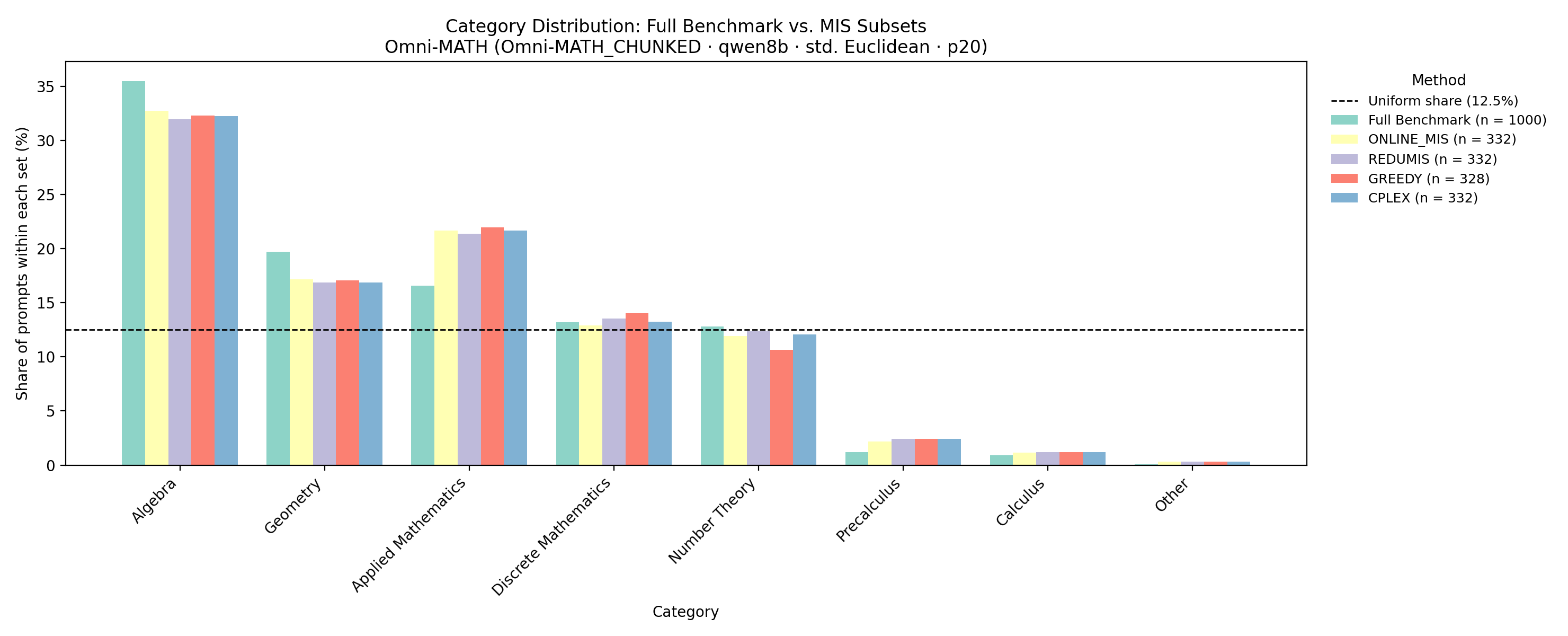}
  \caption{Category shares on Omni-MATH (\texttt{qwen8b}, standardized Euclidean, $p_{20}$, chunked): full benchmark vs.\ the four MIS solvers, as shares within each set (dashed line: uniform share, $12.5\%$). All solvers ($328$--$332$ prompts) reduce the share of Algebra ($35.5\%\to31.9$--$32.7\%$) and increase Applied Mathematics ($16.6\%\to21.4$--$22.0\%$) and the small Precalculus category ($1.2\%\to2.2$--$2.4\%$).}
  \label{fig:catdist_omni}
\end{figure}

Figures~\ref{fig:catdist_gpqa}--\ref{fig:catdist_omni} show, for GPQA, MMLU-Pro and Omni-MATH, how each solver redistributes category mass relative to the full benchmark. All four solvers select nearly identical compositions on every benchmark, so the redistribution is a property of the proximity graph rather than of the solver.

\appbarrier
\section{Baselines}
\label{app:baseline}
\noindent\textbf{Baseline construction.} For each tested configuration, defined by the threshold and
benchmark dataset, we average across embedding, distance measure, solver, and chunking variant and
use that subset size in our baseline calculations. We construct three size-matched baselines: (i)
random selection, in which prompts are sampled uniformly from the entire benchmark; (ii) stratified
selection, in which prompts are sampled proportionally to each prompt category represented in that
benchmark; and (iii) uniform-stratified selection, in which prompts are instead split as equally as
possible \emph{across} categories via a water-filling allocation---each category is capped at its own
available pool size, and any shortfall from categories that run out is redistributed to categories
that still have room---so small and large categories are targeted for comparable representation
rather than their natural share of the benchmark. Each baseline is repeated five times to account for
sampling stochasticity, and evaluated by Spearman $\rho$ (agreement with the full ranking) and Kendall
$W$ across repetitions.

\begin{figure}[!htbp]
  \centering
  \includegraphics[width=\textwidth]{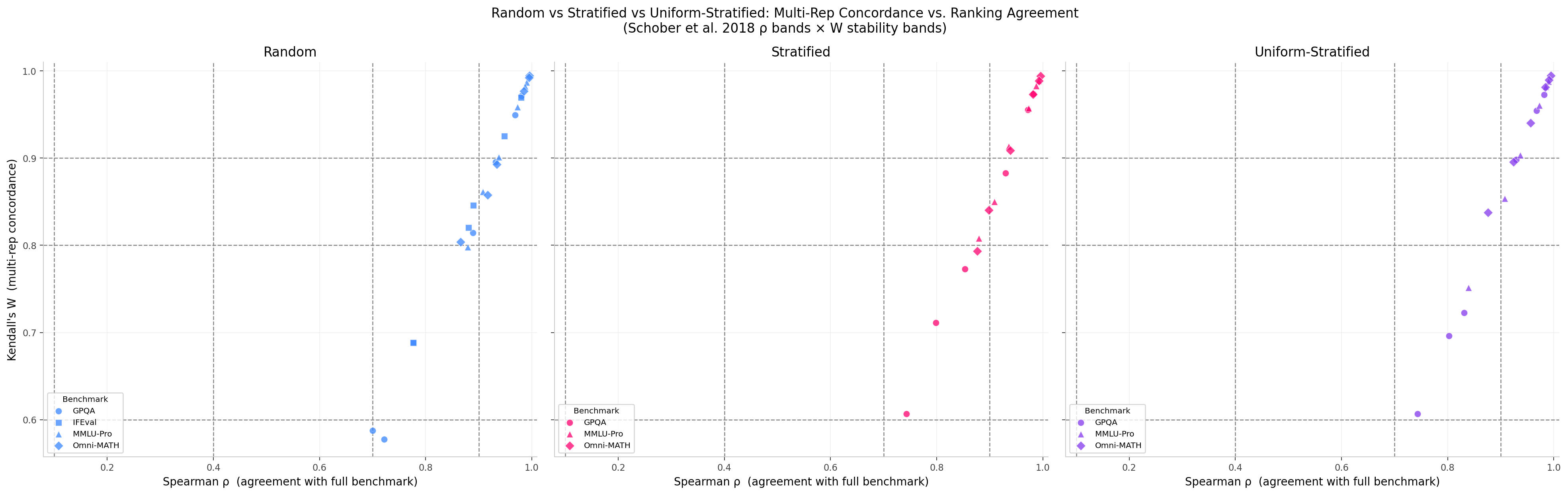}
  \caption{\textbf{Size-matched baselines on the concordance--agreement plane.} Multi-rep Kendall's
  $W$ (concordance across the five repetitions) against Spearman's $\rho$ (agreement with the full
  benchmark) for the random (left), stratified (middle), and uniform-stratified (right) baselines,
  one point per (benchmark, threshold) cell, marker-coded by benchmark. Grid lines are the same
  interpretation bands as Figure~\ref{fig:hypothesis}: the $\rho$-magnitude bands of
  \citet{schober2018correlation} ($0.10/0.40/0.70/0.90$) and the $W$-concordance bands of
  \citet{landis1977measurement} ($0.60/0.80/0.90$). All three baselines occupy almost exclusively
  the faithful ($\rho\ge0.70$) side of the plane and slide \emph{downward} in $W$ as the threshold
  grows---from near-perfect concordance at $p_{10}$/$p_{20}$ (top-right) into the seed-unstable band
  ($W<0.90$) by $p_{80}$--$p_{95}$---so their divergence from the full ranking mostly takes the form
  of \emph{repetition instability}, not reshaping. This is the mirror image of MIS
  (Figure~\ref{fig:hypothesis}), which retains $W\ge0.90$ and instead drifts leftward in $\rho$.
  The only point below $\rho=0.70$ is random sampling on GPQA at $p_{90}$ ($\rho=0.6999$, essentially on the
  boundary).}
  \label{fig:hypothesis_baselines}
\end{figure}

\noindent\textbf{Quadrant occupancy.} Placing the size-matched baselines on the same $\rho=0.70$,
$W=0.90$ plane as Figure~\ref{fig:hypothesis} exposes the difference from MIS structurally (Figure~\ref{fig:hypothesis_baselines}): the baselines mostly stay within the two
$\rho{\ge}0.70$ (faithful) quadrants, but they split between them by \emph{repetition stability}. Random
selection lands in reliable-and-faithful for $45.8\%$ of cells, in faithful-but-seed-unstable
($\rho{\ge}0.70$, $W{<}0.90$) for $50.0\%$, and in doubly-degraded for $4.2\%$; stratified and
uniform-stratified behave identically to each other ($55.6\%$ / $44.4\%$). The only baseline cell below
$\rho=0.70$ is random sampling on GPQA at $p_{90}$ ($\rho=0.6999$, essentially on the boundary), so baseline
divergence takes the form of instability rather than reshaping. The split is threshold-driven: at
$p_{10}$/$p_{20}$ all three baselines are $100\%$ reliable-and-faithful; at $p_{50}$ and $p_{80}$ stratified
and uniform-stratified are reliable-and-faithful in $67\%$ of cells, random in $50\%$ and $25\%$; all three
reach $0\%$ at $p_{90}$/$p_{95}$, where mean $W$ falls to $0.73$--$0.77$ (random $0.73$, stratified $0.77$,
uniform-stratified $0.77$). Note that this $W$ measures variation between independent draws, whereas for MIS
it measures solver-seed variation on a fixed graph, so the two describe reproducibility of different
procedures rather than fidelity. Head overlap is statistically indistinguishable between the three baselines
(mean top-$5$/top-$10$ Jaccard: random $0.545$/$0.649$, stratified $0.532$/$0.639$, uniform-stratified
$0.543$/$0.641$).

\noindent\textbf{Quadrant distribution, by benchmark and by method.} Table~\ref{tab:baseline_distribution_bench}
and Table~\ref{tab:baseline_method_breakdown} give the same four-quadrant breakdown used for MIS in
Appendix~\ref{app:distribution_bench} (Tables~\ref{tab:distribution_bench}--\ref{tab:embedding_breakdown}),
computed from the per-configuration $\rho$ and $W$ values underlying Figure~\ref{fig:hypothesis_baselines}
($60$ configurations total: random $n{=}24$; stratified and uniform-stratified $n{=}18$ each, undefined
for IFEval). Across all $60$ configurations, not one lands in the reliable-but-reshaped quadrant
($\rho{<}0.70,W{\ge}0.90$)---baseline divergence, when it occurs, is exclusively seed instability, never
reshaping, confirming the mirror-image relationship with MIS reported above. \emph{By benchmark}
(Table~\ref{tab:baseline_distribution_bench}), GPQA is hardest for the baselines ($33.3\%$
reliable-and-faithful, with $5.6\%$ doubly-degraded---the random $p_{90}$ boundary case), while MMLU-Pro
($66.7\%$) and Omni-MATH ($61.1\%$) are easiest; this mirrors the benchmark-difficulty ordering seen for MIS
(Table~\ref{tab:distribution_bench}), but at systematically lower occupancy---even the easiest benchmark keeps
only $67\%$ of baseline configurations reliable-and-faithful, versus $81$--$87\%$ for MIS. \emph{By method}
(Table~\ref{tab:baseline_method_breakdown}), stratified and uniform-stratified tie at $55.6\%$
reliable-and-faithful and both avoid the doubly-degraded quadrant, while random is weakest ($45.8\%$).

\begin{table}[!htbp]
\centering
\caption{Distribution of baseline configurations across the four quadrants of Section~\ref{sec:results}
($\rho=0.70$, $W=0.90$; Figure~\ref{fig:hypothesis_baselines}), by benchmark, pooled across the three
baseline methods. Stratified and uniform-stratified are undefined for IFEval, so its $n$ reflects
random only.}
\label{tab:baseline_distribution_bench}
\small
\begin{tabular}{l p{2.1cm} p{2.1cm} p{2.1cm} p{1.8cm} c}
\toprule
\textbf{Benchmark} & 
\textbf{Reliable \& faithful} ($\rho{\ge}.70,W{\ge}.90$) & 
\textbf{Reliable but reshaped} ($\rho{<}.70,W{\ge}.90$) & 
\textbf{Faithful but seed-unstable} ($\rho{\ge}.70,W{<}.90$) & 
\textbf{Doubly-degraded} ($\rho{<}.70,W{<}.90$) & 
\textbf{$n$} \\
\midrule
GPQA      & 33.3\% & 0.0\% & 61.1\% &  5.6\% & 18 \\
IFEval    & 33.3\% & 0.0\% & 66.7\% &  0.0\% &  6 \\
MMLU-Pro  & 66.7\% & 0.0\% & 33.3\% &  0.0\% & 18 \\
Omni-MATH & 61.1\% & 0.0\% & 38.9\% &  0.0\% & 18 \\
\bottomrule
\end{tabular}
\end{table}

\begin{table}[!htbp]
\centering
\caption{Per-baseline-method breakdown of configuration outcomes across the four quadrants of
Section~\ref{sec:results} ($\rho=0.70$, $W=0.90$), pooled across benchmarks.}
\label{tab:baseline_method_breakdown}
\small
\begin{tabular}{l p{2.2cm} p{2.2cm} p{2.2cm} p{2.0cm} c}
\toprule
\textbf{Method} & 
\textbf{Reliable \& faithful} & 
\textbf{Reliable but reshaped} & 
\textbf{Faithful but seed-unstable} & 
\textbf{Doubly-degraded} & 
\textbf{$n$} \\
\midrule
Random             & 45.8\% & 0.0\% & 50.0\% & 4.2\% & 24 \\
Stratified         & 55.6\% & 0.0\% & 44.4\% & 0.0\% & 18 \\
Uniform-stratified & 55.6\% & 0.0\% & 44.4\% & 0.0\% & 18 \\
\bottomrule
\end{tabular}
\end{table}

\noindent\textbf{Quadrant drivers.} Table~\ref{tab:baseline_quadrant_reco} gives the baseline
analog of Table~\ref{tab:quadrant_reco}, characterizing what drives each quadrant. As already
noted, the reliable-but-reshaped quadrant is empty for every baseline---unlike MIS, which reshapes
deliberately under aggressive pruning, a baseline that diverges from the full ranking never does so
while staying seed-stable; its divergence is instability. The reliable-and-faithful
quadrant is dominated by mild thresholds ($p_{10}$/$p_{20}$ alone are $65\%$ of it, each enriched
$1.94\times$ over the base rate), with MMLU-Pro ($1.29\times$) and Omni-MATH ($1.18\times$)
over-represented and no method favored. The faithful-but-seed-unstable quadrant mirrors this at the other
end of the threshold range ($p_{90}$ and $p_{95}$ together are $68\%$ of it, enriched $1.93\times$ and
$2.14\times$), with IFEval ($1.43\times$) and GPQA ($1.31\times$) over-represented. The doubly-degraded
quadrant ($1.7\%$) contains a single configuration: random sampling on GPQA at $p_{90}$ ($\rho=0.6999$).

\begin{table}[!htbp]
\centering
\caption{Configuration patterns behind each quadrant for the size-matched baselines
(Figure~\ref{fig:hypothesis_baselines}), analogous to Table~\ref{tab:quadrant_reco}. Enrichments
($\times$) are relative to the base rate across all $60$ baseline configurations (random
$n{=}24$; stratified/uniform-stratified $n{=}18$ each, undefined for IFEval).}
\label{tab:baseline_quadrant_reco}
\small
\begin{tabular}{@{}p{2.9cm}p{5.0cm}p{4.4cm}@{}}
\toprule
\textbf{Quadrant} & \textbf{Dominant drivers} & \textbf{Takeaway} \\
\midrule
Reliable \& faithful \newline ($\rho\ge0.70,W\ge0.90$; $51.7\%$)
 & Mild thresholds ($p_{10}$/$p_{20}$: $\times1.94$ each, $65\%$ of quadrant); MMLU-Pro ($\times1.29$),
 Omni-MATH ($\times1.18$); no method favored
 & Baselines are usable proxies at mild compression \\
\addlinespace
Reliable but reshaped \newline ($\rho<0.70,W\ge0.90$; $0.0\%$)
 & Not observed ($0/60$)
 & Baselines never reshape the ranking while staying seed-stable---divergence, when it happens,
 is always instability, the mirror image of MIS \\
\addlinespace
Faithful but seed-unstable \newline ($\rho\ge0.70,W<0.90$; $46.7\%$)
 & Aggressive thresholds ($p_{90}$: $\times1.93$, $p_{95}$: $\times2.14$; $68\%$ of quadrant); IFEval
 ($\times1.43$), GPQA ($\times1.31$)
 & Expect instability once compression is aggressive; repeat draws (especially for random) rather
 than trusting a single sample \\
\addlinespace
Doubly-degraded \newline ($\rho<0.70,W<0.90$; $1.7\%$)
 & A single configuration: random, GPQA, $p_{90}$ ($\rho=0.6999$)
 & Boundary case on the smallest benchmark at aggressive compression \\
\bottomrule
\end{tabular}
\end{table}

\noindent\textbf{Leaderboard-head overlap (Jaccard).} Leaderboard-head overlap declines the same way: averaged over the five repetitions, top-$5$ Jaccard falls from $0.79$ at $p_{10}$ to $0.31$ at $p_{95}$ (top-$10$: $0.84\to0.46$), with all three baselines indistinguishable on this measure ($0.545$ vs.\ $0.532$ vs.\ $0.543$ on top-$5$ for random, stratified and uniform-stratified). Grouped by quadrant (Table~\ref{tab:baseline_jaccard}), the faithful quadrants separate by stability---reliable-and-faithful preserves the head best (top-$5$ mean $0.705$, top-$10$ $0.784$; $90\%$ of these configurations share at least half their top-$5$ and $100\%$ at least half their top-$10$), whereas seed-unstable degrades it (top-$5$ mean $0.372$, top-$10$ $0.502$; only $14\%$ reach half-overlap on top-$5$)---so even at high $\rho$, an unstable subset ($W<0.90$) still scrambles the top. Within a quadrant the baselines edge MIS on head overlap (e.g.\ $0.705$ vs.\ $0.561$ on top-$5$ in reliable-and-faithful), consistent with their small $\rho$ advantage. Forcing equal rather than proportional category representation (uniform-stratified) is no more helpful for seed stability, and it departs further from the full benchmark's own category composition than stratified does, at no gain in ranking fidelity or head overlap.

\begin{table}[!htbp]
  \centering
  \caption{Top-$N$ Jaccard overlap between the full-benchmark and size-matched baseline subset
  leaderboards, by quadrant of Figure~\ref{fig:hypothesis_baselines} ($\rho=0.70$, $W=0.90$).
  Overlap is the fraction of shared models among the top $5$ (or top $10$); ``$\ge0.5$'' is the
  share of configurations preserving at least half of the head.}
  \label{tab:baseline_jaccard}
  \small
  \begin{tabular}{lrrrr}
    \toprule
    Quadrant & top-5 (mean) & top-5 $\ge0.5$ & top-10 (mean) & top-10 $\ge0.5$ \\
    \midrule
    Reliable \& faithful ($\rho{\ge}.70,W{\ge}.90$)     & 0.705 & 90\% & 0.784 & 100\% \\
    Reliable but reshaped ($\rho{<}.70,W{\ge}.90$)      & -- & -- & -- & -- \\
    Faithful but seed-unstable ($\rho{\ge}.70,W{<}.90$) & 0.372 & 14\% & 0.502 & 46\% \\
    Doubly-degraded ($\rho{<}.70,W{<}.90$)              & 0.172 &  0\% & 0.273 &  0\% \\
    \bottomrule
  \end{tabular}
\end{table}

\begin{figure}[!htbp]
  \centering
  \includegraphics[width=\textwidth]{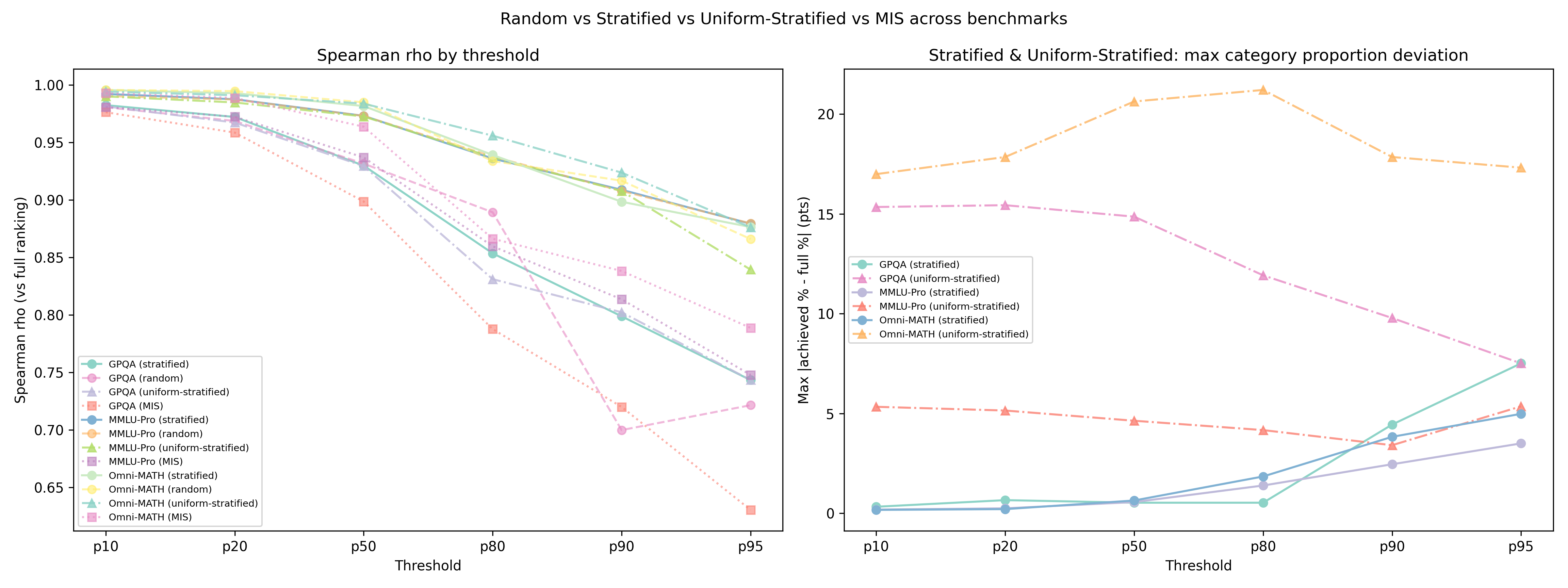}
  \caption{\textbf{Random vs.\ stratified vs.\ uniform-stratified vs.\ MIS, across benchmarks and
  thresholds.} \emph{Left:} Spearman $\rho$ vs.\ the full ranking, one line per (benchmark, method)
  pair. \emph{Right:} maximum $|$achieved category \% $-$ full-benchmark category \%$|$ for the two
  stratified variants only. Uniform-stratified's deviation-from-full line sits well above stratified's
  on MMLU-Pro and Omni-MATH at every threshold (on GPQA the two meet at $p_{95}$, where the allocation of
  $8$ prompts to three categories is the same for both), because uniform-stratified is not
  targeting the full benchmark's proportions in the first place; stratified's line stays near the
  bottom throughout, since matching those proportions is exactly what it is built to do.}
  \label{fig:cross_baseline_compression}
\end{figure}

\noindent\textbf{Compression vs.\ quality, all four methods.} Figure~\ref{fig:cross_baseline_compression}
plots Spearman $\rho$ (left) and the stratified/uniform-stratified category-deviation metric (right)
against threshold for all benchmarks with category labels, extending
Figure~\ref{fig:hypothesis_baselines} with the same random/stratified/uniform-stratified/MIS grouping
used in Table~\ref{tab:full_baseline_comparison}. The left panel shows the four methods follow each
other closely at mild thresholds ($\rho\ge0.93$ for all methods at $p_{10}$/$p_{20}$ on every
benchmark) and separate only once pruning becomes aggressive, where MIS is the method that drifts
lowest in $\rho$ at the most aggressive threshold (its divergence is reshaping, not instability, as established in the main text). The
right panel makes explicit what the two stratified variants are optimizing for: stratified's deviation
from the full benchmark's own category proportions stays small and roughly flat across thresholds
(under $2$ points up to $p_{80}$, rising to $4$--$8$ points at $p_{90}$--$p_{95}$ as subsets become very small),
whereas uniform-stratified's deviation from the full proportions is larger and does not shrink with
threshold on MMLU-Pro and Omni-MATH, because its allocation target is deliberately \emph{not} the full benchmark's proportions
but an equal split across categories. The size of that gap tracks how imbalanced each benchmark's
categories are: on MMLU-Pro, whose $14$ categories are relatively similar in size, the deviation stays
modest ($3$--$5$ points); on GPQA and Omni-MATH, whose few categories are far more imbalanced
(GPQA is $42\%$/$41\%$/$17\%$; Omni-MATH's largest category alone is $35\%$), the deviation is much
larger ($8$--$16$ points on GPQA, $17$--$21$ points on Omni-MATH)---a large deviation from full there
is expected behavior, not degradation.

\begin{table}[!htbp]
\centering
\caption{Full comparison of MIS against the three size-matched baselines (random, stratified, uniform-stratified), by benchmark and threshold. $\rho$ = mean Spearman correlation with the full ranking; $W$ = mean Kendall concordance across repetitions (5 seeds for MIS/random, 5 reps for stratified/uniform-stratified). Stratified and uniform-stratified require category labels and are undefined for IFEval (--).}
\label{tab:full_baseline_comparison}
\resizebox{\textwidth}{!}{%
\begin{tabular}{llc cc cc cc cc}
\toprule
 & & & \multicolumn{2}{c}{MIS} & \multicolumn{2}{c}{Random} & \multicolumn{2}{c}{Stratified} & \multicolumn{2}{c}{Uniform-Stratified} \\
\cmidrule(lr){4-5}\cmidrule(lr){6-7}\cmidrule(lr){8-9}\cmidrule(lr){10-11}
Dataset & Threshold & $|S|$ & $\rho$ & $W$ & $\rho$ & $W$ & $\rho$ & $W$ & $\rho$ & $W$ \\
\midrule
GPQA & $p_{10}$ & 134 & 0.976 & 0.997 & 0.981 & 0.971 & 0.982 & 0.974 & 0.981 & 0.973 \\
 & $p_{20}$ & 82 & 0.959 & 0.991 & 0.969 & 0.949 & 0.972 & 0.956 & 0.967 & 0.955 \\
 & $p_{50}$ & 34 & 0.899 & 0.957 & 0.932 & 0.896 & 0.930 & 0.883 & 0.929 & 0.898 \\
 & $p_{80}$ & 17 & 0.788 & 0.931 & 0.889 & 0.815 & 0.854 & 0.773 & 0.831 & 0.723 \\
 & $p_{90}$ & 11 & 0.720 & 0.867 & 0.700 & 0.587 & 0.799 & 0.711 & 0.802 & 0.696 \\
 & $p_{95}$ & 8 & 0.630 & 0.854 & 0.722 & 0.578 & 0.743 & 0.607 & 0.743 & 0.607 \\
\midrule
IFEval & $p_{10}$ & 196 & 0.963 & 0.995 & 0.980 & 0.970 & -- & -- & -- & -- \\
 & $p_{20}$ & 142 & 0.947 & 0.993 & 0.948 & 0.925 & -- & -- & -- & -- \\
 & $p_{50}$ & 73 & 0.885 & 0.976 & 0.890 & 0.846 & -- & -- & -- & -- \\
 & $p_{80}$ & 33 & 0.752 & 0.961 & 0.881 & 0.820 & -- & -- & -- & -- \\
 & $p_{90}$ & 19 & 0.650 & 0.926 & 0.777 & 0.688 & -- & -- & -- & -- \\
 & $p_{95}$ & 12 & 0.535 & 0.877 & 0.776 & 0.689 & -- & -- & -- & -- \\
\midrule
MMLU-Pro & $p_{10}$ & 307 & 0.981 & 0.998 & 0.990 & 0.987 & 0.992 & 0.989 & 0.990 & 0.988 \\
 & $p_{20}$ & 204 & 0.972 & 0.995 & 0.987 & 0.982 & 0.988 & 0.983 & 0.985 & 0.982 \\
 & $p_{50}$ & 89 & 0.937 & 0.986 & 0.973 & 0.959 & 0.973 & 0.957 & 0.973 & 0.961 \\
 & $p_{80}$ & 36 & 0.860 & 0.938 & 0.938 & 0.901 & 0.936 & 0.913 & 0.937 & 0.903 \\
 & $p_{90}$ & 22 & 0.814 & 0.915 & 0.908 & 0.861 & 0.909 & 0.850 & 0.908 & 0.854 \\
 & $p_{95}$ & 14 & 0.748 & 0.844 & 0.880 & 0.798 & 0.879 & 0.808 & 0.839 & 0.752 \\
\midrule
Omni-MATH & $p_{10}$ & 281 & 0.993 & 0.999 & 0.996 & 0.995 & 0.996 & 0.994 & 0.994 & 0.995 \\
 & $p_{20}$ & 187 & 0.989 & 0.998 & 0.995 & 0.992 & 0.993 & 0.989 & 0.991 & 0.990 \\
 & $p_{50}$ & 74 & 0.964 & 0.992 & 0.985 & 0.977 & 0.982 & 0.973 & 0.984 & 0.982 \\
 & $p_{80}$ & 28 & 0.866 & 0.972 & 0.934 & 0.893 & 0.939 & 0.909 & 0.956 & 0.940 \\
 & $p_{90}$ & 17 & 0.838 & 0.934 & 0.917 & 0.858 & 0.898 & 0.841 & 0.924 & 0.896 \\
 & $p_{95}$ & 11 & 0.789 & 0.901 & 0.866 & 0.804 & 0.876 & 0.793 & 0.876 & 0.838 \\
\bottomrule
\end{tabular}%
}
\end{table}

\begin{figure}[!htbp]
  \centering
  \includegraphics[width=\textwidth]{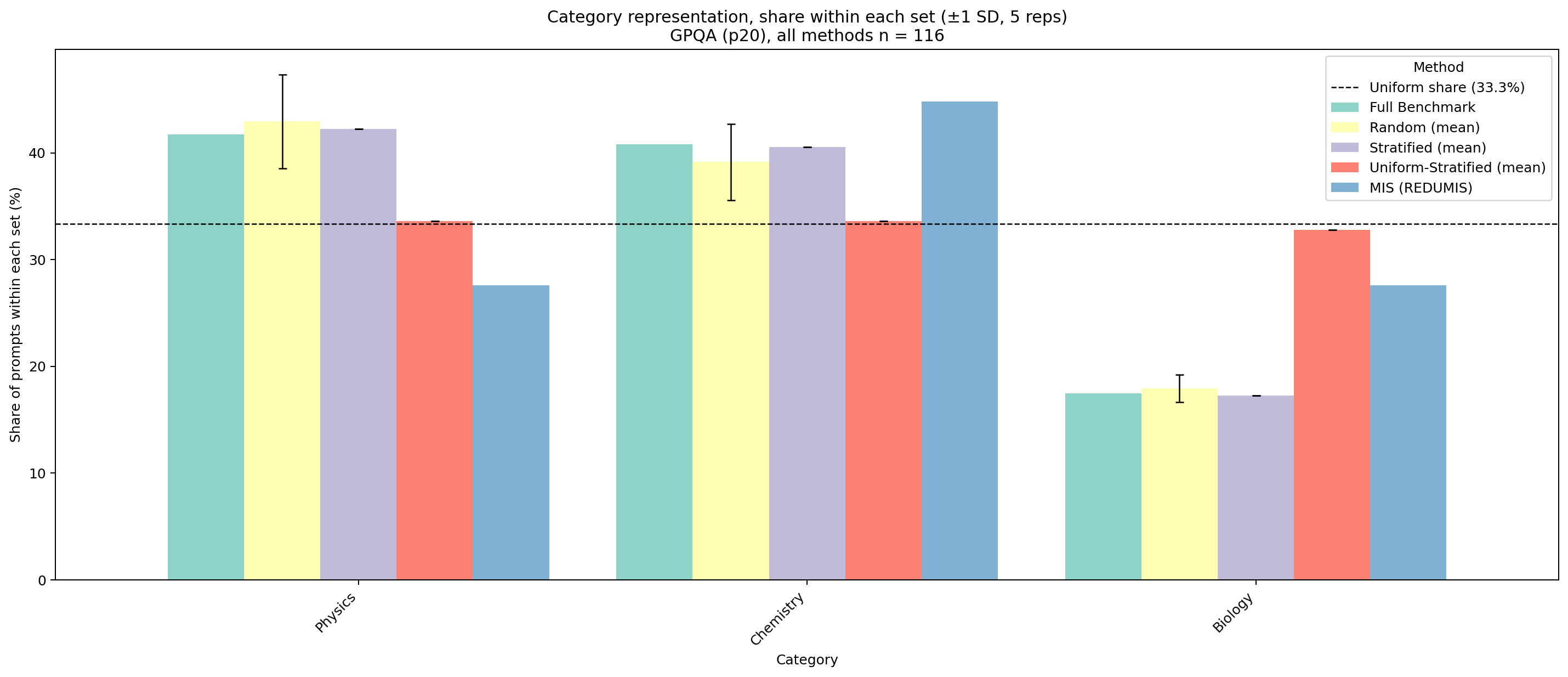}
  \caption{Category shares at equal subset size for GPQA ($p_{20}$; MIS: ReduMIS, \texttt{qwen8b}, standardized Euclidean, chunked; $116$ prompts). Random, stratified and uniform-stratified subsets are drawn at the same size (mean $\pm1$ SD over $5$ repetitions); every bar is the share of a category within its own set; dashed line: uniform share ($33.3\%$).}
  \label{fig:samen_gpqa}
\end{figure}

\begin{figure}[!htbp]
  \centering
  \includegraphics[width=\textwidth]{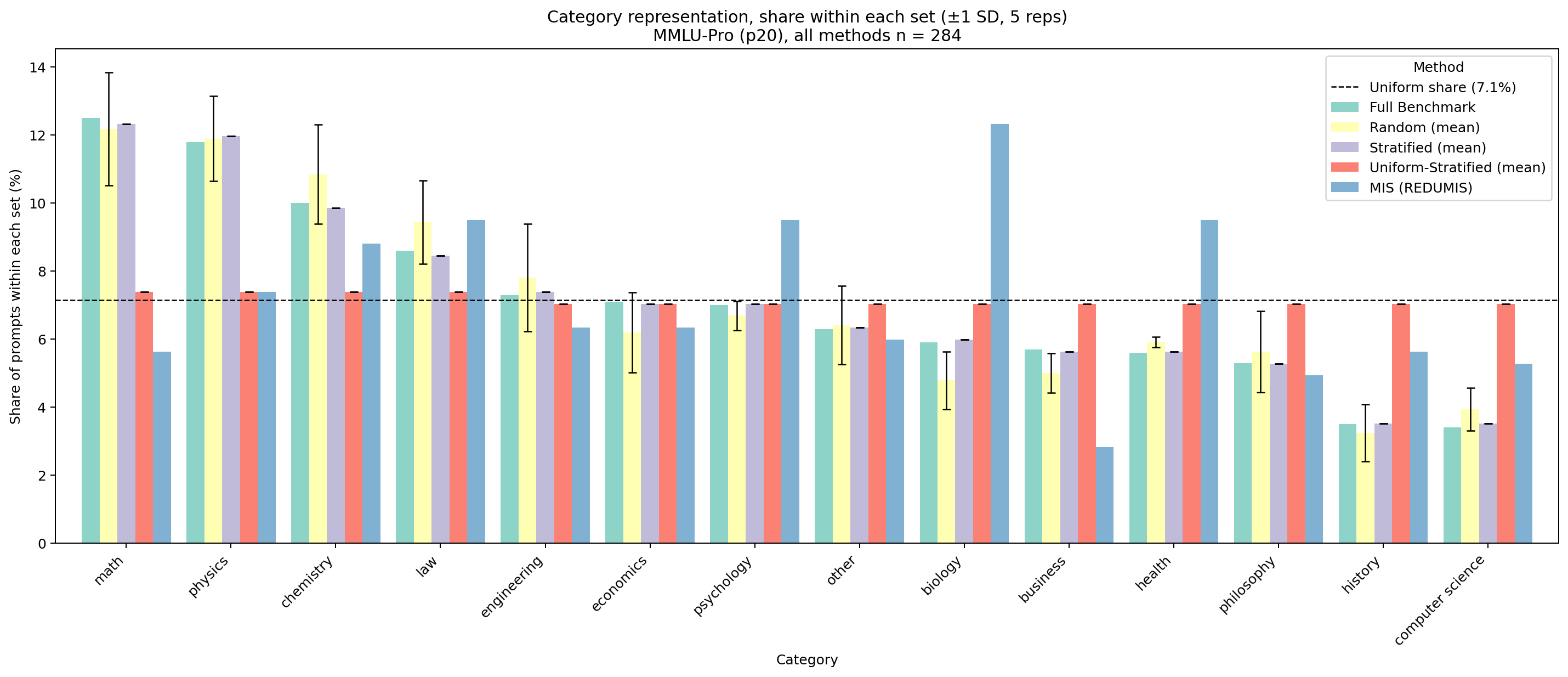}
  \caption{Category shares at equal subset size for MMLU-Pro ($p_{20}$; $284$ prompts; as in Figure~\ref{fig:samen_gpqa}); dashed line: uniform share ($7.1\%$).}
  \label{fig:samen_mmlu}
\end{figure}

\begin{figure}[!htbp]
  \centering
  \includegraphics[width=\textwidth]{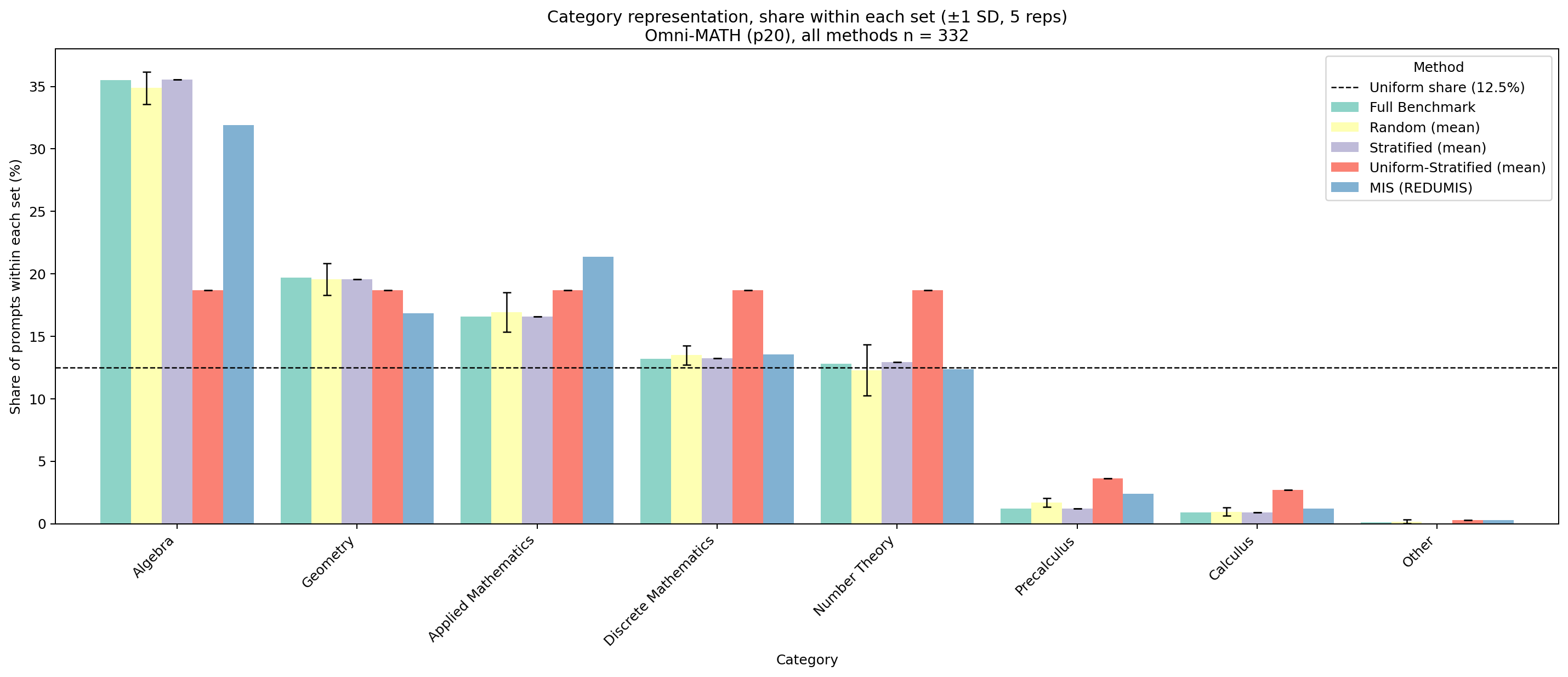}
  \caption{Category shares at equal subset size for Omni-MATH ($p_{20}$; $332$ prompts; as in Figure~\ref{fig:samen_gpqa}); dashed line: uniform share ($12.5\%$).}
  \label{fig:samen_omni}
\end{figure}

\noindent\textbf{Category coverage at equal subset size.} Figures~\ref{fig:samen_gpqa}--\ref{fig:samen_omni}
repeat the real-category coverage comparison of Figure~\ref{fig:coverage} at $p_{20}$ for GPQA, MMLU-Pro and
Omni-MATH, contrasting the three baselines with MIS. All baselines are drawn at exactly the size of the MIS
subset shown ($116$, $284$ and $332$ prompts), and every bar is the share of a category within its own set,
so the comparison is free of size effects. Stratified sampling reproduces the full benchmark's composition,
and uniform-stratified sampling flattens it toward an equal share per category, using the labels. Without
using labels, MIS lifts Biology on GPQA ($17.5\%\to27.6\%$) and several mid-sized MMLU-Pro categories
(biology $5.9\%\to12.3\%$, health $5.6\%\to9.5\%$, psychology $7.0\%\to9.5\%$) while compressing the largest
(math $12.5\%\to5.6\%$, physics $11.8\%\to7.4\%$); on Omni-MATH the redistribution is milder (Algebra
$35.5\%\to31.9\%$, Applied Mathematics $16.6\%\to21.4\%$). Table~\ref{tab:rev_counts} reports retained counts
and within-subset shares side by side, Table~\ref{tab:rev_equal_size} the corresponding fidelity, MAE, $W$
and entropy for all four benchmarks, and Table~\ref{tab:rev_entropy} the entropy at every threshold: MIS is more
balanced than the median random subset in $17$ of $18$ benchmark--threshold cells. This redistribution is an
emergent byproduct of where semantic redundancy concentrates in embedding space, not a targeted allocation, so
its shape does not reduce to either baseline's rule.

\begin{table}[!htbp]
\centering
\caption{Recommended MIS configuration vs. same-size sampling baselines at $p_{20}$ (1,000 draws per baseline).}
\label{tab:rev_equal_size}
\small
\resizebox{\textwidth}{!}{%
\begin{tabular}{llrrrrrr}
\toprule
Benchmark & Method & n & $\rho$ & MAE & W & Category entropy & Max. deviation from equal share (pts) \\
\midrule
GPQA & MIS (recommended) & 116 & 0.975 & 4.8 & 0.999 & 0.974 & 11.5 \\
 & Random & 116 & 0.980 & 3.0 & 0.970 & 0.936 & 15.9 \\
 & Stratified & 116 & 0.980 & 3.0 & 0.970 & 0.940 & 16.1 \\
 & Uniform-stratified & 116 & 0.979 & 3.3 & 0.970 & 1.000 & 0.6 \\
\midrule
IFEval & MIS (recommended) & 200 & 0.976 & 2.7 & 1.000 & -- & -- \\
 & Random & 200 & 0.976 & 1.5 & 0.964 & -- & -- \\
\midrule
MMLU-Pro & MIS (recommended) & 284 & 0.985 & 3.6 & 0.999 & 0.978 & 5.2 \\
 & Random & 284 & 0.992 & 1.7 & 0.988 & 0.969 & 6.1 \\
 & Stratified & 284 & 0.992 & 1.7 & 0.988 & 0.975 & 5.2 \\
 & Uniform-stratified & 284 & 0.990 & 2.0 & 0.987 & 1.000 & 0.3 \\
\midrule
Omni-MATH & MIS (recommended) & 332 & 0.996 & 4.8 & 1.000 & 0.810 & 19.4 \\
 & Random & 332 & 0.996 & 1.6 & 0.995 & 0.775 & 23.0 \\
 & Stratified & 332 & 0.996 & 1.5 & 0.995 & 0.776 & 23.0 \\
 & Uniform-stratified & 332 & 0.996 & 2.0 & 0.995 & 0.867 & 12.2 \\
\bottomrule
\end{tabular}}
\\[3pt]{\footnotesize Category entropy of within-subset proportions, normalized to [0, 1]. W: across the five MIS seeds, or across groups of five baseline draws. IFEval has no category labels.}
\end{table}

\begin{table}[!htbp]
\centering
\caption{Retained prompts per category at $p_{20}$: count (share within the subset), all methods at the same subset size.}
\label{tab:rev_counts}
\small
\resizebox{\textwidth}{!}{%
\begin{tabular}{llrrrrr}
\toprule
Benchmark & Category & Full benchmark & MIS & Random & Stratified & Uniform-stratified \\
\midrule
GPQA & Physics & 186 (41.7\%) & 32.0 (27.6\%) & 48.4 (41.7\%) & 49.0 (42.2\%) & 39.0 (33.6\%) \\
 & Chemistry & 182 (40.8\%) & 52.0 (44.8\%) & 47.4 (40.8\%) & 47.0 (40.5\%) & 39.0 (33.6\%) \\
 & Biology & 78 (17.5\%) & 32.0 (27.6\%) & 20.3 (17.5\%) & 20.0 (17.2\%) & 38.0 (32.8\%) \\
\midrule
MMLU-Pro & math & 125 (12.5\%) & 16.0 (5.6\%) & 35.4 (12.5\%) & 35.0 (12.3\%) & 21.0 (7.4\%) \\
 & physics & 118 (11.8\%) & 21.0 (7.4\%) & 33.3 (11.7\%) & 34.0 (12.0\%) & 21.0 (7.4\%) \\
 & chemistry & 100 (10.0\%) & 25.0 (8.8\%) & 28.4 (10.0\%) & 28.0 (9.9\%) & 21.0 (7.4\%) \\
 & law & 86 (8.6\%) & 27.0 (9.5\%) & 24.6 (8.7\%) & 24.0 (8.5\%) & 21.0 (7.4\%) \\
 & engineering & 73 (7.3\%) & 18.0 (6.3\%) & 20.5 (7.2\%) & 21.0 (7.4\%) & 20.0 (7.0\%) \\
 & economics & 71 (7.1\%) & 18.0 (6.3\%) & 20.1 (7.1\%) & 20.0 (7.0\%) & 20.0 (7.0\%) \\
 & psychology & 70 (7.0\%) & 27.0 (9.5\%) & 20.0 (7.0\%) & 20.0 (7.0\%) & 20.0 (7.0\%) \\
 & other & 63 (6.3\%) & 17.0 (6.0\%) & 18.1 (6.4\%) & 18.0 (6.3\%) & 20.0 (7.0\%) \\
 & biology & 59 (5.9\%) & 35.0 (12.3\%) & 16.8 (5.9\%) & 17.0 (6.0\%) & 20.0 (7.0\%) \\
 & business & 57 (5.7\%) & 8.0 (2.8\%) & 16.2 (5.7\%) & 16.0 (5.6\%) & 20.0 (7.0\%) \\
 & health & 56 (5.6\%) & 27.0 (9.5\%) & 15.9 (5.6\%) & 16.0 (5.6\%) & 20.0 (7.0\%) \\
 & philosophy & 53 (5.3\%) & 14.0 (4.9\%) & 15.1 (5.3\%) & 15.0 (5.3\%) & 20.0 (7.0\%) \\
 & history & 35 (3.5\%) & 16.0 (5.6\%) & 10.0 (3.5\%) & 10.0 (3.5\%) & 20.0 (7.0\%) \\
 & computer science & 34 (3.4\%) & 15.0 (5.3\%) & 9.7 (3.4\%) & 10.0 (3.5\%) & 20.0 (7.0\%) \\
\midrule
Omni-MATH & Algebra & 355 (35.5\%) & 106.0 (31.9\%) & 117.8 (35.5\%) & 118.0 (35.5\%) & 62.0 (18.7\%) \\
 & Geometry & 197 (19.7\%) & 56.0 (16.9\%) & 65.2 (19.6\%) & 65.0 (19.6\%) & 62.0 (18.7\%) \\
 & Applied Mathematics & 166 (16.6\%) & 71.0 (21.4\%) & 55.0 (16.6\%) & 55.0 (16.6\%) & 62.0 (18.7\%) \\
 & Discrete Mathematics & 132 (13.2\%) & 45.0 (13.6\%) & 44.1 (13.3\%) & 44.0 (13.3\%) & 62.0 (18.7\%) \\
 & Number Theory & 128 (12.8\%) & 41.0 (12.3\%) & 42.5 (12.8\%) & 43.0 (13.0\%) & 62.0 (18.7\%) \\
 & Precalculus & 12 (1.2\%) & 8.0 (2.4\%) & 4.0 (1.2\%) & 4.0 (1.2\%) & 12.0 (3.6\%) \\
 & Calculus & 9 (0.9\%) & 4.0 (1.2\%) & 3.0 (0.9\%) & 3.0 (0.9\%) & 9.0 (2.7\%) \\
 & Other & 1 (0.1\%) & 1.0 (0.3\%) & 0.3 (0.1\%) & 0.0 (0.0\%) & 1.0 (0.3\%) \\
\bottomrule
\end{tabular}}
\\[3pt]{\footnotesize Baseline counts are averages over 1,000 draws.}
\end{table}

\begin{table}[!htbp]
\centering
\caption{Normalized category entropy of within-subset proportions by threshold.}
\label{tab:rev_entropy}
\small
\resizebox{\textwidth}{!}{%
\begin{tabular}{llrrrrrr}
\toprule
Benchmark & Thr. & n & MIS & Random & Stratified & Uniform-stratified & Random subsets less balanced than MIS \\
\midrule
GPQA & $p_{10}$ & 153 & 0.978 & 0.939 & 0.944 & 1.000 & 99\% \\
 & $p_{20}$ & 116 & 0.974 & 0.936 & 0.940 & 1.000 & 96\% \\
 & $p_{50}$ & 56 & 0.994 & 0.929 & 0.945 & 1.000 & 100\% \\
 & $p_{80}$ & 21 & 0.994 & 0.899 & 0.953 & 1.000 & 95\% \\
 & $p_{90}$ & 10 & 0.991 & 0.849 & 0.960 & 0.991 & 100\% \\
 & $p_{95}$ & 8 & 0.985 & 0.816 & 0.985 & 0.985 & 91\% \\
\midrule
MMLU-Pro & $p_{10}$ & 410 & 0.981 & 0.971 & 0.974 & 1.000 & 98\% \\
 & $p_{20}$ & 284 & 0.978 & 0.969 & 0.975 & 1.000 & 92\% \\
 & $p_{50}$ & 141 & 0.971 & 0.960 & 0.974 & 1.000 & 82\% \\
 & $p_{80}$ & 64 & 0.947 & 0.937 & 0.972 & 0.998 & 66\% \\
 & $p_{90}$ & 46 & 0.894 & 0.918 & 0.971 & 0.997 & 20\% \\
 & $p_{95}$ & 30 & 0.929 & 0.884 & 0.978 & 0.995 & 86\% \\
\midrule
Omni-MATH & $p_{10}$ & 441 & 0.804 & 0.776 & 0.774 & 0.851 & 100\% \\
 & $p_{20}$ & 332 & 0.810 & 0.775 & 0.776 & 0.867 & 100\% \\
 & $p_{50}$ & 150 & 0.789 & 0.770 & 0.774 & 0.920 & 79\% \\
 & $p_{80}$ & 64 & 0.782 & 0.757 & 0.759 & 0.960 & 75\% \\
 & $p_{90}$ & 42 & 0.802 & 0.746 & 0.771 & 0.967 & 91\% \\
 & $p_{95}$ & 26 & 0.768 & 0.726 & 0.745 & 0.974 & 78\% \\
\bottomrule
\end{tabular}}
\end{table}

\noindent\textbf{Single configurations against $1{,}000$ same-size random draws.}\label{app:rev_random} The size-matched baselines above use the subset size averaged over configurations and five repetitions. To compare individual configurations with sampling, we drew $1{,}000$ random subsets at exactly the size of each of the $2{,}558$ configurations, using a separate random generator so that the five-repetition baselines are unchanged. Table~\ref{tab:rev_rec_vs_random} compares the recommended configuration with these draws in $\rho$, MAE and top-$5$ overlap, Table~\ref{tab:rev_best_vs_random} does the same for the best configuration in each cell (selected post hoc, hence optimistic), Table~\ref{tab:rev_tail} reports the distribution of the random draws, and Table~\ref{tab:rev_all_configs} summarizes all configurations. The recommended configuration is within $0.007$ of the random median at $p_{20}$ on all four benchmarks and above the 5th percentile of random draws in all eight $p_{90}$/$p_{95}$ cells; it preserves the top-$5$ models better than random in $12$ of $24$ cells (5 of the 8 $p_{90}$/$p_{95}$ cells; on Omni-MATH at $p_{10}$ both round to $1.00$). The best configuration per cell is at or above the random median in $16$ of $24$ cells; in the other eight it is below by at most $0.012$, and by less than $0.001$ in three of them.

\begin{table}[!htbp]
\centering
\caption{Recommended MIS configuration (ReduMIS / qwen8b / std. Euclidean / chunked) vs. 1,000 random subsets of exactly the same size.}
\label{tab:rev_rec_vs_random}
\small
\resizebox{\textwidth}{!}{%
\begin{tabular}{llrrrrrrrr}
\toprule
Benchmark & Thr. & n & $\rho$ MIS & $\rho$ random (median) & $\rho$ random (5th pct) & MAE MIS & MAE random & Top-5 MIS & Top-5 random \\
\midrule
GPQA & $p_{10}$ & 153 & 0.980 & 0.986 & 0.980 & 2.9 & 2.3 & 0.43 & 0.74 \\
 & $p_{20}$ & 116 & 0.975 & 0.980 & 0.971 & 4.8 & 2.8 & 0.43 & 0.69 \\
 & $p_{50}$ & 56 & 0.959 & 0.958 & 0.933 & 6.2 & 4.4 & 0.43 & 0.56 \\
 & $p_{80}$ & 21 & 0.794 & 0.895 & 0.817 & 13.0 & 7.5 & 0.43 & 0.38 \\
 & $p_{90}$ & 10 & 0.685 & 0.810 & 0.623 & 14.6 & 11.2 & 0.11 & 0.26 \\
 & $p_{95}$ & 8 & 0.668 & 0.781 & 0.547 & 15.4 & 12.6 & 0.43 & 0.21 \\
\midrule
IFEval & $p_{10}$ & 256 & 0.981 & 0.984 & 0.974 & 1.9 & 1.1 & 0.43 & 0.76 \\
 & $p_{20}$ & 200 & 0.976 & 0.977 & 0.962 & 2.7 & 1.3 & 0.43 & 0.73 \\
 & $p_{50}$ & 117 & 0.955 & 0.955 & 0.926 & 3.3 & 1.9 & 0.67 & 0.65 \\
 & $p_{80}$ & 57 & 0.875 & 0.907 & 0.845 & 4.6 & 2.9 & 0.67 & 0.53 \\
 & $p_{90}$ & 35 & 0.834 & 0.858 & 0.749 & 5.8 & 3.8 & 0.67 & 0.45 \\
 & $p_{95}$ & 20 & 0.772 & 0.786 & 0.605 & 5.0 & 5.1 & 0.67 & 0.34 \\
\midrule
MMLU-Pro & $p_{10}$ & 410 & 0.985 & 0.995 & 0.992 & 2.9 & 1.2 & 0.43 & 0.69 \\
 & $p_{20}$ & 284 & 0.985 & 0.992 & 0.988 & 3.6 & 1.6 & 0.67 & 0.63 \\
 & $p_{50}$ & 141 & 0.971 & 0.983 & 0.973 & 2.7 & 2.5 & 0.67 & 0.53 \\
 & $p_{80}$ & 64 & 0.955 & 0.964 & 0.943 & 4.3 & 3.8 & 0.43 & 0.45 \\
 & $p_{90}$ & 46 & 0.949 & 0.951 & 0.914 & 4.7 & 4.6 & 0.43 & 0.41 \\
 & $p_{95}$ & 30 & 0.931 & 0.928 & 0.875 & 4.7 & 5.7 & 0.43 & 0.36 \\
\midrule
Omni-MATH & $p_{10}$ & 441 & 0.998 & 0.997 & 0.996 & 2.4 & 1.2 & 1.00 & 1.00 \\
 & $p_{20}$ & 332 & 0.996 & 0.996 & 0.994 & 4.8 & 1.4 & 1.00 & 0.99 \\
 & $p_{50}$ & 150 & 0.994 & 0.992 & 0.986 & 6.2 & 2.5 & 0.67 & 0.92 \\
 & $p_{80}$ & 64 & 0.980 & 0.980 & 0.965 & 13.2 & 3.9 & 0.67 & 0.81 \\
 & $p_{90}$ & 42 & 0.971 & 0.970 & 0.944 & 16.3 & 4.8 & 0.67 & 0.73 \\
 & $p_{95}$ & 26 & 0.944 & 0.953 & 0.905 & 13.6 & 6.1 & 0.43 & 0.64 \\
\bottomrule
\end{tabular}}
\\[3pt]{\footnotesize $\rho$ = Spearman correlation with the full ranking of the 66 LLMs; MAE in accuracy points; Top-5 = Jaccard overlap of the top-5 models. Random: median ($\rho$, MAE) or mean (Top-5) over 1,000 draws.}
\end{table}

\begin{table}[!htbp]
\centering
\caption{Best MIS configuration in each benchmark--threshold cell (highest $\rho$ among all 2,558 configurations, selected post hoc) vs. 1,000 random subsets of the same size.}
\label{tab:rev_best_vs_random}
\small
\resizebox{\textwidth}{!}{%
\begin{tabular}{llrrrrrrrr}
\toprule
Benchmark & Thr. & Configuration & n & $\rho$ MIS & $\rho$ random & MAE MIS & MAE random & Top-5 MIS & Top-5 random \\
\midrule
GPQA & $p_{10}$ & CPLEX/nemotron/corr/C & 204 & 0.991 & 0.991 & 1.6 & 1.8 & 1.00 & 0.80 \\
 & $p_{20}$ & Redu/nemotron/corr/C & 146 & 0.988 & 0.985 & 2.1 & 2.4 & 0.67 & 0.73 \\
 & $p_{50}$ & Redu/nemotron/corr/C & 78 & 0.973 & 0.970 & 2.9 & 3.7 & 1.00 & 0.61 \\
 & $p_{80}$ & Redu/nemotron/cos/C & 35 & 0.949 & 0.933 & 5.4 & 5.8 & 0.43 & 0.48 \\
 & $p_{90}$ & CPLEX/nemotron/seucl/C & 23 & 0.904 & 0.902 & 8.9 & 7.2 & 0.67 & 0.39 \\
 & $p_{95}$ & Online/nemotron/corr/C & 14 & 0.875 & 0.859 & 8.3 & 9.6 & 0.48 & 0.31 \\
\midrule
IFEval & $p_{10}$ & Redu/qwen4b/seucl/C & 225 & 0.988 & 0.981 & 2.2 & 1.2 & 0.67 & 0.73 \\
 & $p_{20}$ & Greedy/qwen4b/seucl/C & 164 & 0.980 & 0.970 & 1.9 & 1.5 & 0.67 & 0.70 \\
 & $p_{50}$ & Redu/qwen8b/seucl/C & 117 & 0.955 & 0.955 & 3.3 & 1.9 & 0.67 & 0.65 \\
 & $p_{80}$ & CPLEX/qwen4b/seucl/NC & 49 & 0.906 & 0.892 & 2.8 & 3.1 & 0.67 & 0.51 \\
 & $p_{90}$ & Redu/qwen8b/corr/C & 25 & 0.843 & 0.820 & 3.7 & 4.5 & 0.67 & 0.39 \\
 & $p_{95}$ & Redu/qwen8b/corr/C & 17 & 0.776 & 0.763 & 4.4 & 5.6 & 0.25 & 0.33 \\
\midrule
MMLU-Pro & $p_{10}$ & Redu/nemotron/corr/C & 558 & 0.993 & 0.997 & 1.9 & 0.9 & 0.67 & 0.74 \\
 & $p_{20}$ & Redu/nemotron/seucl/C & 440 & 0.989 & 0.995 & 2.2 & 1.1 & 0.67 & 0.70 \\
 & $p_{50}$ & CPLEX/nemotron/seucl/C & 225 & 0.978 & 0.989 & 3.0 & 1.8 & 0.67 & 0.60 \\
 & $p_{80}$ & Redu/qwen8b/seucl/NC & 69 & 0.964 & 0.966 & 3.8 & 3.7 & 0.43 & 0.45 \\
 & $p_{90}$ & CPLEX/qwen8b/seucl/NC & 47 & 0.952 & 0.952 & 5.1 & 4.5 & 0.43 & 0.41 \\
 & $p_{95}$ & Online/qwen8b/seucl/NC & 31 & 0.943 & 0.931 & 5.5 & 5.6 & 0.43 & 0.36 \\
\midrule
Omni-MATH & $p_{10}$ & Redu/qwen8b/seucl/NC & 436 & 0.999 & 0.997 & 2.4 & 1.1 & 1.00 & 1.00 \\
 & $p_{20}$ & Redu/qwen8b/seucl/NC & 323 & 0.998 & 0.996 & 4.5 & 1.4 & 1.00 & 0.98 \\
 & $p_{50}$ & CPLEX/qwen8b/seucl/C & 150 & 0.995 & 0.992 & 7.2 & 2.5 & 0.67 & 0.92 \\
 & $p_{80}$ & Online/qwen8b/seucl/C & 64 & 0.980 & 0.980 & 12.8 & 3.9 & 0.62 & 0.81 \\
 & $p_{90}$ & Greedy/qwen8b/seucl/C & 39 & 0.973 & 0.968 & 14.6 & 4.9 & 0.67 & 0.71 \\
 & $p_{95}$ & Redu/qwen8b/seucl/C & 26 & 0.944 & 0.953 & 13.6 & 6.1 & 0.43 & 0.64 \\
\bottomrule
\end{tabular}}
\\[3pt]{\footnotesize Redu = ReduMIS, Online = Online-MIS; seucl = standardized Euclidean, corr = correlation, cos = cosine; C = chunked, NC = no chunking. Selected per cell, hence optimistic.}
\end{table}

\begin{table}[!htbp]
\centering
\caption{Recommended MIS configuration vs. the distribution of 1,000 random subsets of the same size.}
\label{tab:rev_tail}
\small
\resizebox{\textwidth}{!}{%
\begin{tabular}{llrrrrrrr}
\toprule
Benchmark & Thr. & n & MIS $\rho$ & Random median & Random 5th pct & Random 1st pct & Worst random draw & Random draws below MIS \\
\midrule
GPQA & $p_{10}$ & 153 & 0.980 & 0.986 & 0.980 & 0.977 & 0.971 & 6\% \\
 & $p_{20}$ & 116 & 0.975 & 0.980 & 0.971 & 0.967 & 0.958 & 15\% \\
 & $p_{50}$ & 56 & 0.959 & 0.958 & 0.933 & 0.921 & 0.901 & 53\% \\
 & $p_{80}$ & 21 & 0.794 & 0.895 & 0.817 & 0.762 & 0.544 & 2\% \\
 & $p_{90}$ & 10 & 0.685 & 0.810 & 0.623 & 0.477 & 0.191 & 12\% \\
 & $p_{95}$ & 8 & 0.668 & 0.781 & 0.547 & 0.416 & 0.244 & 17\% \\
\midrule
IFEval & $p_{10}$ & 256 & 0.981 & 0.984 & 0.974 & 0.967 & 0.953 & 27\% \\
 & $p_{20}$ & 200 & 0.976 & 0.977 & 0.962 & 0.952 & 0.935 & 42\% \\
 & $p_{50}$ & 117 & 0.955 & 0.955 & 0.926 & 0.911 & 0.890 & 49\% \\
 & $p_{80}$ & 57 & 0.875 & 0.907 & 0.845 & 0.799 & 0.754 & 18\% \\
 & $p_{90}$ & 35 & 0.834 & 0.858 & 0.749 & 0.691 & 0.511 & 33\% \\
 & $p_{95}$ & 20 & 0.772 & 0.786 & 0.605 & 0.515 & 0.405 & 43\% \\
\midrule
MMLU-Pro & $p_{10}$ & 410 & 0.985 & 0.995 & 0.992 & 0.991 & 0.989 & 0\% \\
 & $p_{20}$ & 284 & 0.985 & 0.992 & 0.988 & 0.985 & 0.981 & 1\% \\
 & $p_{50}$ & 141 & 0.971 & 0.983 & 0.973 & 0.969 & 0.953 & 3\% \\
 & $p_{80}$ & 64 & 0.955 & 0.964 & 0.943 & 0.926 & 0.912 & 23\% \\
 & $p_{90}$ & 46 & 0.949 & 0.951 & 0.914 & 0.890 & 0.861 & 47\% \\
 & $p_{95}$ & 30 & 0.931 & 0.928 & 0.875 & 0.836 & 0.740 & 54\% \\
\midrule
Omni-MATH & $p_{10}$ & 441 & 0.998 & 0.997 & 0.996 & 0.995 & 0.994 & 93\% \\
 & $p_{20}$ & 332 & 0.996 & 0.996 & 0.994 & 0.993 & 0.991 & 61\% \\
 & $p_{50}$ & 150 & 0.994 & 0.992 & 0.986 & 0.983 & 0.971 & 86\% \\
 & $p_{80}$ & 64 & 0.980 & 0.980 & 0.965 & 0.957 & 0.937 & 48\% \\
 & $p_{90}$ & 42 & 0.971 & 0.970 & 0.944 & 0.922 & 0.837 & 53\% \\
 & $p_{95}$ & 26 & 0.944 & 0.953 & 0.905 & 0.850 & 0.725 & 32\% \\
\bottomrule
\end{tabular}}
\end{table}

\begin{table}[!htbp]
\centering
\caption{All 2,558 MIS configurations vs. 1,000 random subsets at their own subset size.}
\label{tab:rev_all_configs}
\small
\resizebox{\textwidth}{!}{%
\begin{tabular}{lrrrrr}
\toprule
Solver & Configurations & Above worst random draw & Above random 5th pct & At or above random median & Median percentile among random draws \\
\midrule
All solvers & 2558 & 93.2\% & 69.0\% & 15.1\% & 15.6 \\
CPLEX & 648 & 92.9\% & 68.2\% & 15.6\% & 14.9 \\
GREEDY & 648 & 93.8\% & 70.4\% & 17.3\% & 15.6 \\
Online-MIS & 631 & 93.0\% & 69.6\% & 11.4\% & 16.5 \\
ReduMIS & 631 & 93.2\% & 68.0\% & 16.0\% & 15.2 \\
\bottomrule
\end{tabular}}
\end{table}

\appbarrier
\section{Absolute-Score Preservation (MAE)}
\label{app:mae}
\noindent\textbf{Subset accuracy error (MAE).} Beyond ranking agreement, we measure how closely a
subset reproduces each model's \emph{absolute} accuracy. For a given benchmark (GPQA, IFEval,
MMLU-Pro, Omni-MATH), the baseline is each model's mean accuracy over the \emph{full} prompt set---the
average of that model's per-prompt results. Every solver run is defined by an algorithm, chunking
variant, benchmark, threshold, seed, and distance measure, and yields a subset of retained instances.
For each model $m$ we recompute its mean accuracy on that subset, $\mathrm{acc}^{\text{subset}}_m$, and
compare it to its full-benchmark accuracy $\mathrm{acc}^{\text{full}}_m$, giving one signed error and
one absolute error per model. Averaging the per-model absolute errors over all $N_{\text{models}}$
models summarizes a single run:
\begin{equation}
\text{run MAE} \;=\; \frac{1}{N_{\text{models}}}\sum_{m=1}^{N_{\text{models}}}
\bigl|\,\mathrm{acc}^{\text{subset}}_m - \mathrm{acc}^{\text{full}}_m\,\bigr|.
\label{eq:run_mae}
\end{equation}
The signed error is retained separately to expose systematic over- or under-estimation, whereas
Eq.~\eqref{eq:run_mae} captures the typical magnitude of the discrepancy. To report a stable figure we
average the run-level MAE across all $N_{\text{runs}}$ runs that share a configuration---either all runs
with the same benchmark and threshold, or all runs of a given algorithm:
\begin{equation}
\text{mae\_mean} \;=\; \frac{1}{N_{\text{runs}}}\sum_{r=1}^{N_{\text{runs}}} \text{run MAE}_r .
\label{eq:mae_mean}
\end{equation}
Lower values indicate that the subset preserves not only the \emph{ordering} of models (captured by
$\rho$ and $W$) but also their absolute accuracy levels.

\begin{figure}[!htbp]
  \centering
  \includegraphics[width=\textwidth]{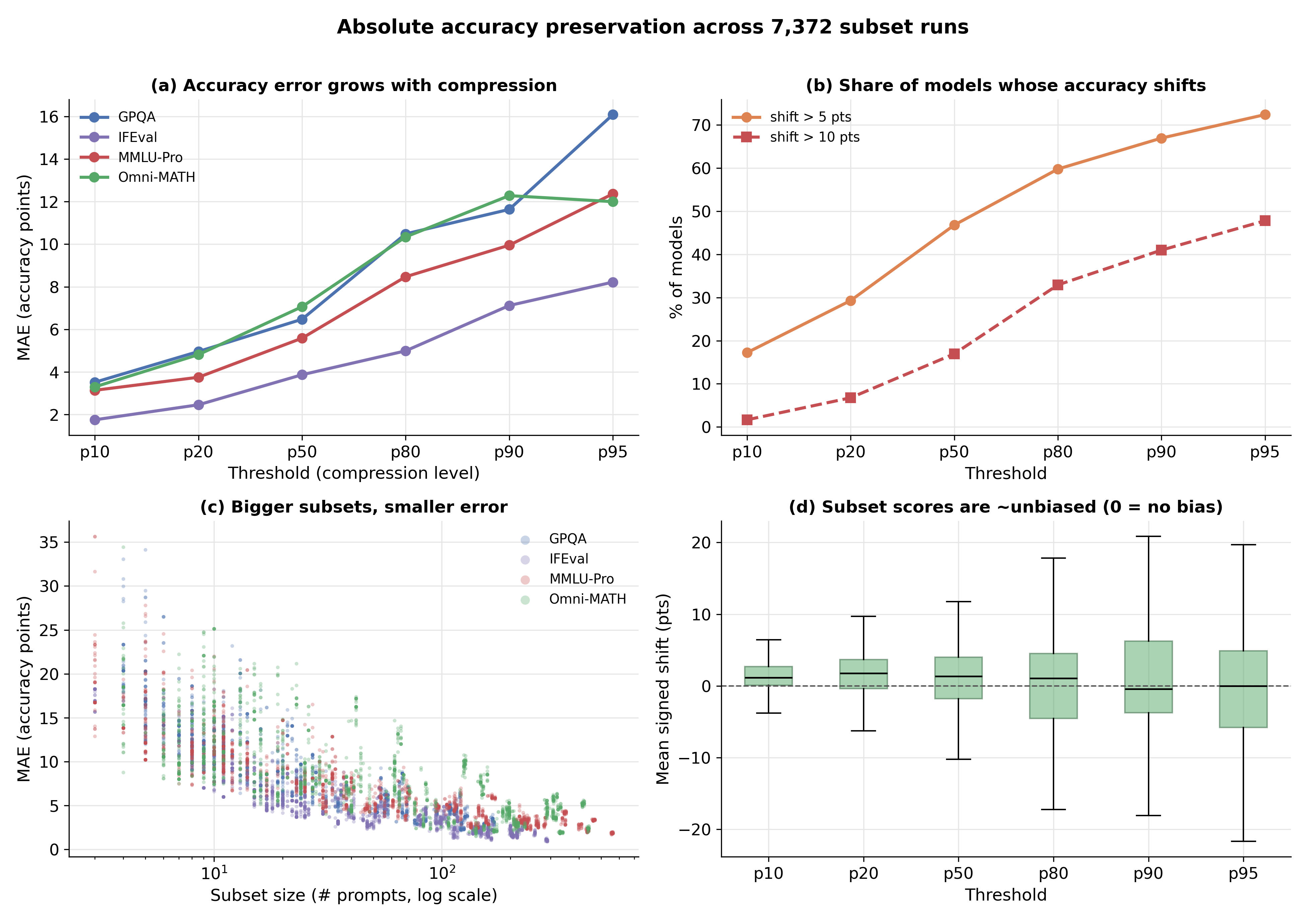}
  \caption{\textbf{Absolute accuracy preservation across $7{,}372$ subset runs.}
  Run-level MAE is the mean over models of $|\mathrm{acc}^{\text{subset}}_m-\mathrm{acc}^{\text{full}}_m|$
  (Eq.~\ref{eq:run_mae}), in accuracy points.
  \textbf{(a)}~MAE rises with the threshold (compression level) on every benchmark, from
  ${\approx}3$ points at $p_{10}$ to $8$--$16$ at $p_{95}$ (Omni-MATH dips slightly from $p_{90}$ to $p_{95}$); IFEval is the most faithful and GPQA degrades
  fastest. \textbf{(b)}~The share of models whose accuracy moves by more than $5$ and $10$ points grows
  from ${\sim}17\%/2\%$ at $p_{10}$ to ${\sim}72\%/48\%$ at $p_{95}$. \textbf{(c)}~MAE falls steeply with
  subset size (log scale): most error is incurred by very small subsets and the curve flattens once
  subsets reach the low hundreds. \textbf{(d)}~The mean signed shift stays centered on zero at every
  threshold, so subsets are approximately unbiased---the growing MAE is dispersion, not directional bias.}
  \label{fig:mae_panel}
\end{figure}

\noindent\textbf{Absolute-accuracy error.} Figure~\ref{fig:mae_panel} reports how well subsets
reproduce each model's full-benchmark accuracy, complementing the rank-based analysis. The error is
governed by the compression level, not the solver: averaged over all configurations, the four solvers
differ by less than $0.2$ accuracy points (CPLEX $7.33$, GREEDY $7.23$, Online-MIS $7.37$, ReduMIS
$7.21$), whereas moving from mild to aggressive thresholds more than triples the MAE; Table~\ref{tab:mae_per_benchmark}
gives the full per-benchmark, per-threshold breakdown. Panels~(a)--(b) show this degradation is
broad---by $p_{95}$ roughly half the models shift by more than $10$ points, and the largest single-model shift
($81.9$ points) occurs in a $3$-prompt MMLU-Pro subset (Online-MIS, \texttt{qwen4b}, cosine, $p_{95}$;
Table~\ref{tab:mae_worst})---and panel~(c) traces it to subset size, with small subsets carrying most
of the error. Panel~(d) confirms the errors are near-unbiased (overall mean signed shift $+0.79$
points), so subset scores neither systematically inflate nor deflate accuracy; the risk of aggressive
pruning is variance, which is exactly what a practitioner controls through the threshold.

\begin{table}[!htbp]
\centering
\caption{Per-benchmark MAE (accuracy points) for MIS subsets across distance thresholds, averaged over all embedding models, distances, and solvers.}
\label{tab:mae_per_benchmark}
\small
\begin{tabular}{lcccccc}
\toprule
\textbf{Benchmark} & \textbf{$p_{10}$} & \textbf{$p_{20}$} & \textbf{$p_{50}$} & \textbf{$p_{80}$} & \textbf{$p_{90}$} & \textbf{$p_{95}$} \\
\midrule
GPQA      & 3.52 & 4.96 & 6.48 & 10.48 & 11.63 & 16.08 \\
IFEval    & 1.76 & 2.46 & 3.87 &  4.99 &  7.12 &  8.22 \\
MMLU-Pro  & 3.14 & 3.75 & 5.59 &  8.47 &  9.95 & 12.36 \\
Omni-MATH & 3.30 & 4.82 & 7.07 & 10.34 & 12.28 & 12.00 \\
\bottomrule
\end{tabular}
\end{table}

\begin{table}[!htbp]
\centering
\caption{Subset runs with the largest single-model accuracy shift. Both cases occur at the most aggressive threshold ($p_{95}$) with very small subsets, where a single volatile model dominates the error.}
\label{tab:mae_worst}
\small
\begin{tabular}{lllccccl}
\toprule
\textbf{Benchmark} & \textbf{Solver} & \textbf{Embed.} & \textbf{Dist.} & \textbf{$|S|$} & \textbf{MAE} & \textbf{Max shift} & \textbf{Worst model} \\
\midrule
MMLU-Pro & Online-MIS & qwen4b & cosine                & 3 & 35.7 & 81.9 & Kimi-K2-Instruct \\
GPQA     & ReduMIS    & qwen4b & cosine/correlation     & 4 & 20.4 & 72.7 & Grok-4-0709 \\
\bottomrule
\end{tabular}
\end{table}

\begin{figure}[!htbp]
  \centering
  \includegraphics[width=\textwidth]{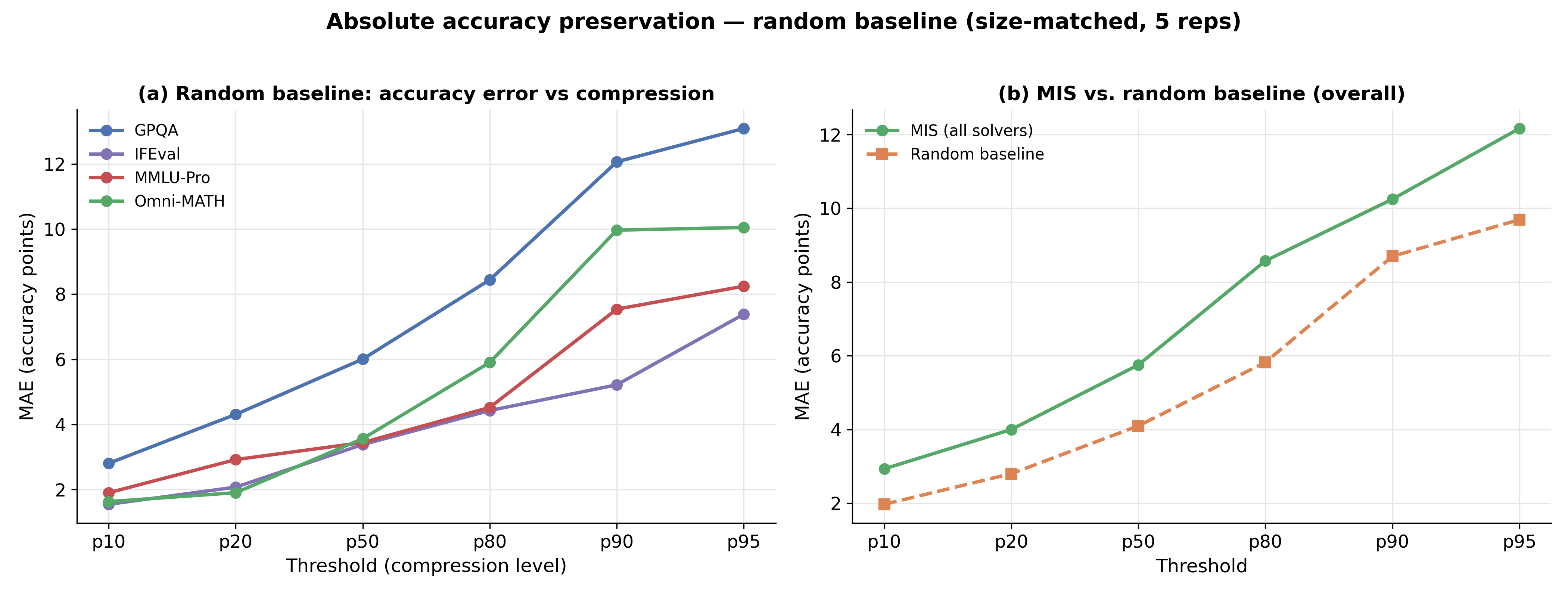}
  \caption{\textbf{Absolute accuracy preservation for the random baseline (size-matched, $5$ reps).}
  \textbf{(a)}~Random-baseline MAE vs.\ threshold per benchmark: error rises monotonically with
  compression, from ${\sim}2$ points at $p_{10}$ to $7$--$13$ at $p_{95}$, with the same benchmark
  ordering as MIS (GPQA hardest, IFEval/MMLU-Pro easiest). \textbf{(b)}~Overall MAE of MIS (all solvers)
  vs.\ the random baseline: random attains marginally lower MAE at every threshold, mirroring its small
  $\rho$ advantage; both are near-unbiased (random mean signed shift $-0.27$ points). Together with the
  results in Sec.~\ref{sec:results}, this locates MIS's distinctive contribution in reproducibility and coverage rather than absolute-accuracy fidelity. See Figure~\ref{fig:mae_all_baselines} for the comparison against all three baselines together.}
  \label{fig:mae_random}
\end{figure}

\noindent\textbf{Absolute-accuracy error of the random baseline.} Figure~\ref{fig:mae_random} repeats
the MAE analysis for the size-matched random baseline, averaged over five repetitions. The random
baseline behaves like MIS in every qualitative respect: its MAE rises monotonically with the threshold
(panel~a), follows the same benchmark ordering (GPQA hardest at $7.78$ points on average; IFEval and
MMLU-Pro easiest at $4.00$ and $4.76$), and is near-unbiased overall (mean signed shift $-0.27$
points), so its subset scores neither systematically inflate nor deflate accuracy. Quantitatively,
however, random attains \emph{marginally lower} MAE than MIS at every threshold (panel~b;
Table~\ref{tab:mae_comparison}): $1.97$ vs.\ $2.93$ points at $p_{10}$, rising to $9.69$ vs.\ $12.16$
at $p_{95}$, a gap of roughly one to three accuracy points ($0.96$ at $p_{10}$, $2.75$ at $p_{80}$) that is
largest under aggressive pruning. This mirrors the small Spearman-$\rho$
advantage of the baselines reported in Section~\ref{sec:results}: i.i.d.\ sampling reproduces both the
ranking and the absolute accuracy of the full benchmark slightly better on average. The higher MAE of MIS reflects a change in the subset's difficulty mix that accompanies the coverage
redistribution: on Omni-MATH, for example, the recommended configuration selects problems with a lower mean
official difficulty rating than same-size random subsets at $p_{80}$--$p_{95}$ ($4.3$--$4.6$ vs.\ $4.9$). In
short, a practitioner who needs an unbiased absolute-accuracy estimate at a fixed budget should use random
sampling, whereas MIS is preferable when a single reproducible subset and label-free coverage balancing matter.

\begin{figure}[!htbp]
  \centering
  \includegraphics[width=\textwidth]{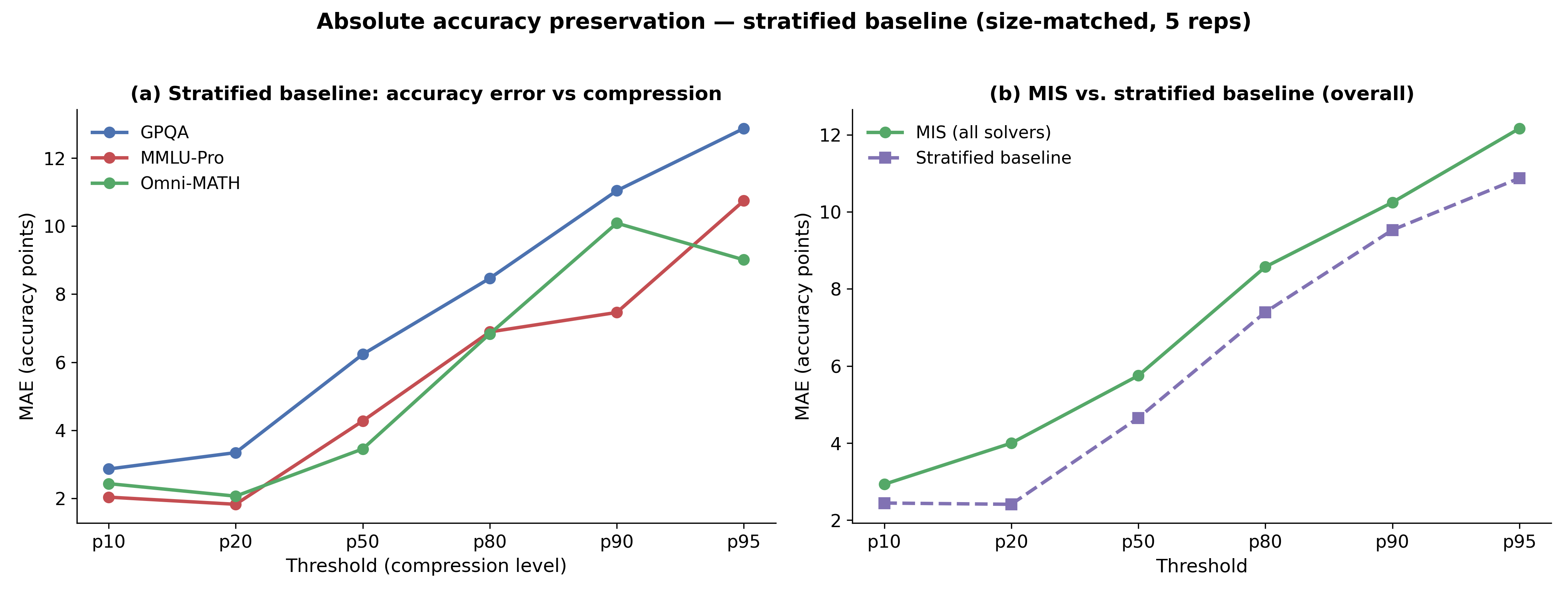}
  \caption{\textbf{Absolute accuracy preservation for the stratified baseline (size-matched, $5$ reps).}
  Stratified sampling draws prompts so that each category keeps its \emph{proportion} in the full
  benchmark (proportional allocation), and is defined only for benchmarks with category labels, so
  \textbf{IFEval is absent} (GPQA, MMLU-Pro, Omni-MATH only).
  \textbf{(a)}~MAE vs.\ threshold per benchmark; error rises with compression overall (with small dips at
  $p_{20}$ on MMLU-Pro and Omni-MATH and at $p_{95}$ on Omni-MATH), GPQA hardest.
  \textbf{(b)}~Overall MAE of MIS (all solvers) vs.\ stratified: stratified stays below MIS at every
  threshold and is near-unbiased (mean signed shift $-0.23$ points). Because stratified omits IFEval,
  its overall average is not strictly comparable to MIS's; the per-threshold ordering is nonetheless
  stable. See Figure~\ref{fig:mae_all_baselines} for the comparison against all three baselines
  together.}
  \label{fig:mae_stratified}
\end{figure}

\noindent\textbf{Absolute-accuracy error of the stratified baseline.} Figure~\ref{fig:mae_stratified}
repeats the MAE analysis for the size-matched stratified baseline, which allocates the subset across a
benchmark's categories \emph{in proportion to their frequency in the full benchmark} (rather than
uniformly), and is therefore defined only for the three benchmarks with category labels (GPQA, MMLU-Pro,
Omni-MATH; IFEval is excluded). Its behavior matches the other selectors qualitatively: MAE rises
with the threshold overall (with small dips at $p_{20}$ on MMLU-Pro and Omni-MATH and at $p_{95}$ on
Omni-MATH), keeps the same benchmark ordering (GPQA hardest at $7.47$ points on
average), and remains near-unbiased (mean signed shift $-0.23$ points). Quantitatively it stays below MIS at every threshold ($2.44$ points vs.\ MIS's $2.93$ at $p_{10}$,
widening to $10.88$ vs.\ $12.16$ at $p_{95}$). Preserving the benchmark's category proportions thus yields no absolute-accuracy
advantage over i.i.d.\ sampling---expected, since proportional stratification only reduces variance from
category-frequency imbalance, which i.i.d.\ sampling already matches in expectation---consistent with its
negligible effect on ranking agreement and seed stability (Section~\ref{sec:results}).

\begin{figure}[!htbp]
  \centering
  \includegraphics[width=\textwidth]{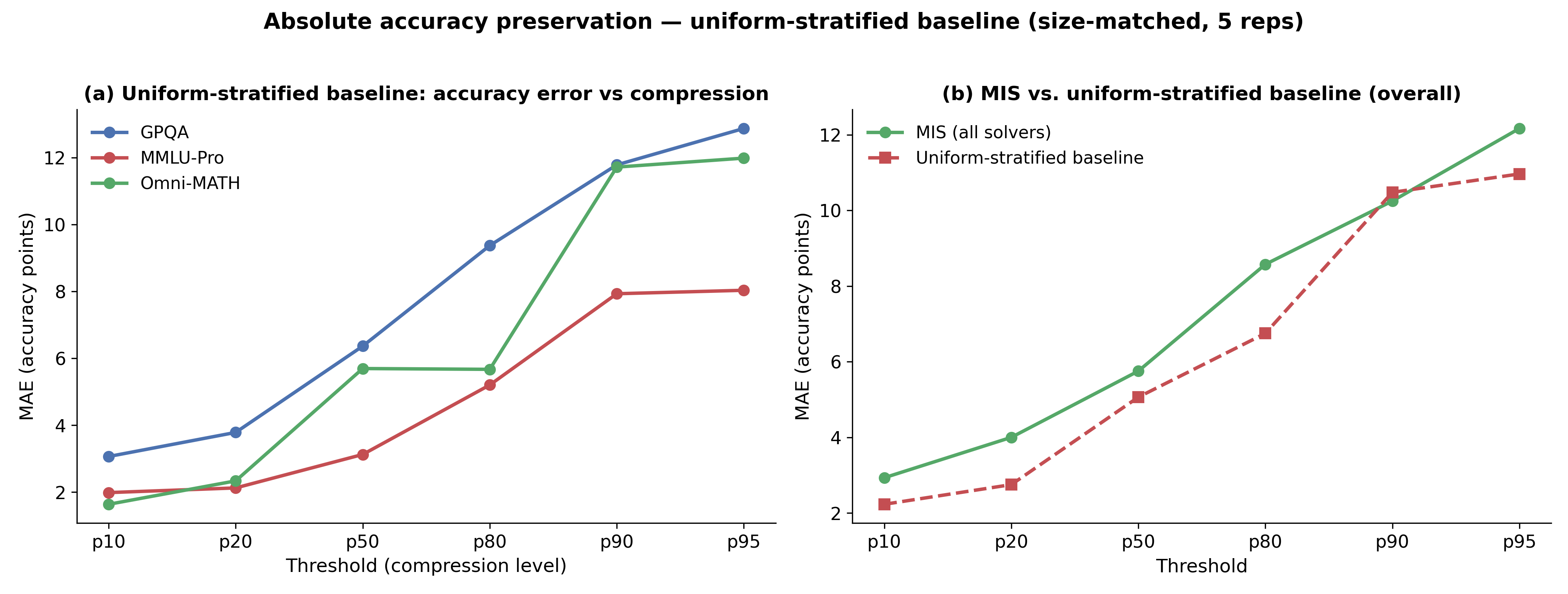}
  \caption{\textbf{Absolute accuracy preservation for the uniform-stratified baseline (size-matched,
  $5$ reps).} Uniform-stratified sampling draws prompts so that every category receives an
  \emph{equal} share of the subset via water-filling, rather than a share proportional to its
  frequency in the full benchmark, and like stratified sampling is defined only for benchmarks with
  category labels, so \textbf{IFEval is absent} (GPQA, MMLU-Pro, Omni-MATH only).
  \textbf{(a)}~MAE vs.\ threshold per benchmark: error rises with compression, GPQA hardest
  throughout, with a pronounced late-threshold spike on GPQA and Omni-MATH once their few, imbalanced
  categories leave very small per-category budgets.
  \textbf{(b)}~Overall MAE of MIS (all solvers) vs.\ uniform-stratified: uniform-stratified stays
  below MIS through $p_{80}$ but is noisier under aggressive pruning and briefly exceeds MIS at
  $p_{90}$ ($10.48$ vs.\ $10.25$) before dropping back below it at $p_{95}$. See
  Figure~\ref{fig:mae_all_baselines} for the comparison against all three baselines together.}
  \label{fig:mae_uniform_stratified}
\end{figure}

\noindent\textbf{Absolute-accuracy error of the uniform-stratified baseline.} Figure~\ref{fig:mae_uniform_stratified}
repeats the MAE analysis for the size-matched uniform-stratified baseline, which allocates the subset
\emph{equally} across a benchmark's categories via water-filling rather than proportionally to their
frequency, and like stratified sampling is defined only for the three benchmarks with category labels
(GPQA, MMLU-Pro, Omni-MATH; IFEval is excluded). baselines: MAE rises with the threshold (on Omni-MATH it stays flat from $p_{50}$ to $p_{80}$), GPQA is hardest ($7.87$ points on average, versus
$4.73$ for MMLU-Pro and $6.51$ for Omni-MATH), and the errors are near-unbiased (mean signed shift
$+1.13$ points, the largest of the three baselines but still small relative to the MAE itself).
Quantitatively it stays below MIS at mild thresholds ($2.23$/$2.75$ vs.\ MIS's $2.93$/$4.00$ at
$p_{10}$/$p_{20}$), but under aggressive pruning it briefly exceeds MIS at $p_{90}$ ($10.48$ vs.\ $10.25$)
before falling back below it at $p_{95}$ ($10.96$ vs.\ $12.16$), driven by GPQA and Omni-MATH's few, highly
imbalanced categories, where forcing equal representation leaves very few prompts per category once the
subset shrinks to single digits. Forcing equal rather than proportional category coverage therefore buys
no consistent absolute-accuracy advantage over MIS, and its higher variance under aggressive pruning is
the accuracy-side counterpart of the seed-instability result reported in Section~\ref{sec:results}.

\begin{table}[!htbp]
\centering
\caption{Mean Absolute Error (MAE, accuracy points) across distance thresholds for MIS (averaged over all four solvers and four benchmarks), the size-matched Random baseline, and the Stratified and Uniform-Stratified baselines (GPQA, MMLU-Pro, Omni-MATH; IFEval undefined for the latter two). Lower is better.}
\label{tab:mae_comparison}
\small
\begin{tabular}{lcccccc}
\toprule
\textbf{Method} & \textbf{$p_{10}$} & \textbf{$p_{20}$} & \textbf{$p_{50}$} & \textbf{$p_{80}$} & \textbf{$p_{90}$} & \textbf{$p_{95}$} \\
\midrule
MIS (Ours)           & 2.93 & 4.00 & 5.75 & 8.57 & 10.25 & 12.16 \\
Random Baseline       & 1.97 & 2.80 & 4.10 & 5.82 &  8.70 &  9.69 \\
Stratified Baseline   & 2.44 & 2.41 & 4.65 & 7.40 &  9.53 & 10.88 \\
Uniform-Stratified    & 2.23 & 2.75 & 5.06 & 6.75 & 10.48 & 10.96 \\
\bottomrule
\end{tabular}
\end{table}

\begin{figure}[!htbp]
  \centering
  \includegraphics[width=0.7\linewidth]{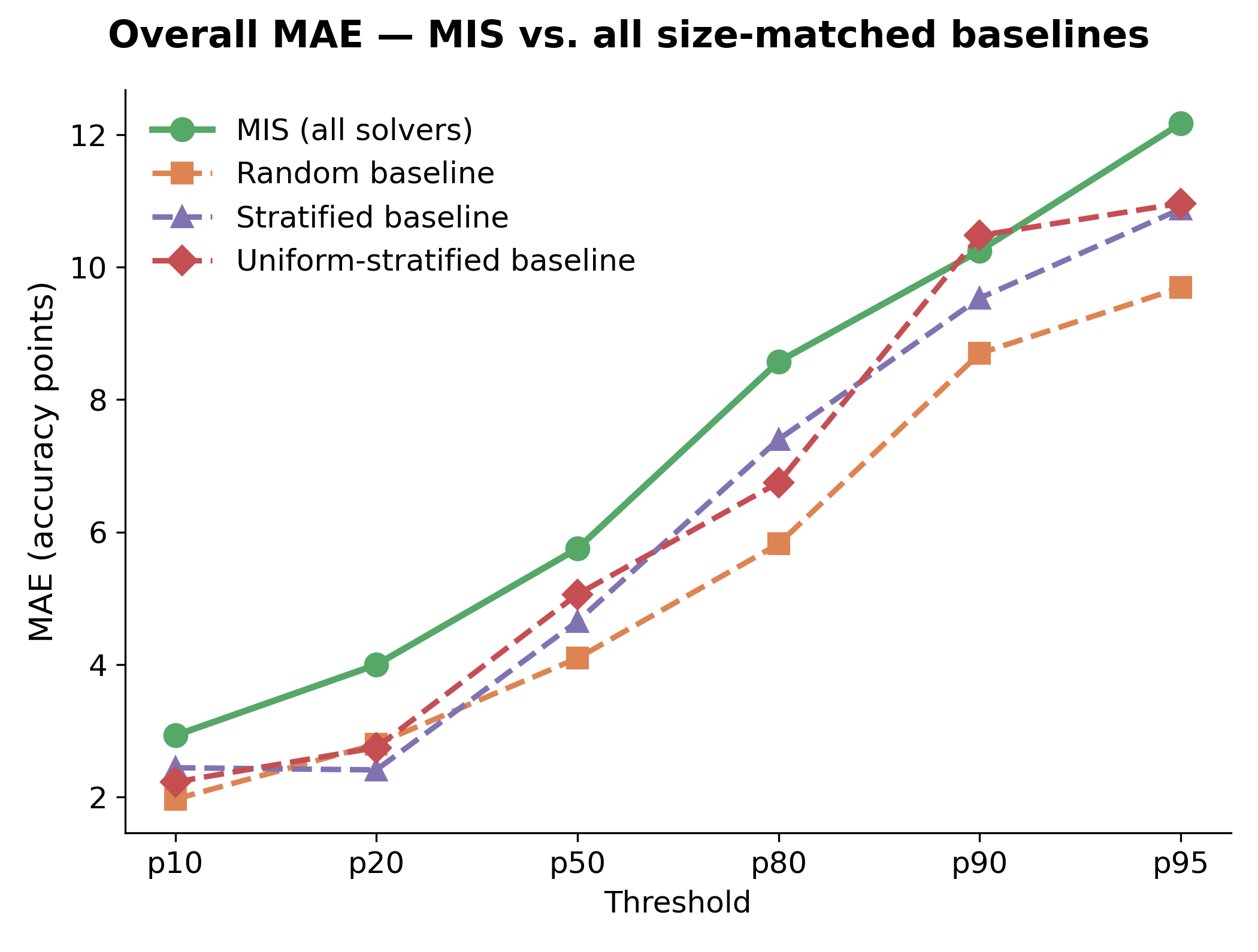}
  \caption{\textbf{Overall MAE across all four methods.} Mean absolute error (accuracy points),
  averaged over benchmarks, vs.\ distance threshold, for MIS (all solvers) and the three
  size-matched baselines. Random and stratified track closely and stay below MIS at every threshold;
  uniform-stratified matches them at mild thresholds but is noisier under aggressive pruning.}
  \label{fig:mae_all_baselines}
\end{figure}

\noindent\textbf{Comparison across all baselines.} Table~\ref{tab:mae_comparison} and
Figure~\ref{fig:mae_all_baselines} bring together the four methods. All three size-matched baselines
beat MIS on raw MAE at nearly every threshold, but the margin and its stability differ: random is the
strongest and most consistent baseline overall, attaining the lowest MAE at five of six thresholds
(losing only at $p_{20}$, to stratified) and needing no category information, which also makes it the
only baseline defined on all four benchmarks. Stratified and uniform-stratified both trail random by a
widening margin under aggressive pruning -- stratified reaches $10.88$ points at $p_{95}$ and
uniform-stratified $10.96$, versus random's $9.69$---though both still stay below MIS's $12.16$ at that
threshold. Uniform-stratified is the least consistent of the three: comparable to the other baselines
at $p_{10}$/$p_{20}$, it briefly loses its advantage over MIS entirely at $p_{90}$ ($10.48$ vs.\ MIS's
$10.25$) before falling back below MIS at $p_{95}$, reflecting the non-monotonic behavior noted above.
Overall, size-matched sampling preserves absolute accuracy better than MIS, while MIS provides a single
reproducible subset and label-free coverage redistribution; the choice depends on which of these matters for the
intended use of the reduced benchmark.

\appbarrier
\section{Unsupervised Structure of the Prompt Proximity Graph Feature Space}
\label{app:topology}

To understand what drives the topology of the proximity graphs, we perform an unsupervised clustering of all instances in the space of their graph-theoretic features and then characterize the resulting clusters \emph{post hoc} against the generative factors of each instance. The analysis shows that the feature geometry is governed almost entirely by two factors---the edge-retention threshold and the benchmark---while the embedding size, the chunking scheme, and the embedding model
leave essentially no detectable signature. The distance measure has only a second-order effect, driven by \texttt{seuclidean}; \texttt{cosine} and
\texttt{correlation} are effectively interchangeable.
 
\paragraph{Data and features.}
 
Each instance is a proximity graph summarized by nine graph-theoretic
features: the number of vertices and edges, density, average / maximum / minimum degree, the global clustering coefficient, the number of connected components, and the size of the largest connected component. Every instance is generated by a fully balanced factorial combination of six metadata factors, decoded from its identifier (Table~\ref{tab:app:factors}), giving $N=648$ instances. Two features are near-perfect proxies for design factors: the vertex count is fixed by the benchmark (GPQA${\approx}446$, IFEval${=}541$, MMLU-Pro and Omni-MATH${=}1000$), and it correlates
$0.99$ with the largest-component size; density is set directly by the retention threshold. This is important for interpreting the clusters below.
 
\begin{table}[!htbp]
\centering
\small
\caption{Generative factors of each instance (balanced factorial, $N=648$).
The embedding model is nested within the embedding size.}
\label{tab:app:factors}
\begin{tabular}{@{}lll@{}}
\toprule
Factor & Levels & Count/level \\
\midrule
Embedding size    & large, small & 432 / 216 \\
Chunking          & chunked, nochunk (large); fixed (small) & 216 each \\
Benchmark         & GPQA, IFEval, MMLU-Pro, Omni-MATH & 162 each \\
Embedding model   & nemotron, qwen4b, qwen8b (large); bge, e5, gist (small) & 144 / 72 \\
Distance measure  & correlation, cosine, seuclidean & 216 each \\
Threshold         & p10, p20, p50, p80, p90, p95 & 108 each \\
\bottomrule
\end{tabular}
\end{table}
 
\paragraph{Clustering procedure.}
 
\begin{figure}[!htbp]
\centering
\includegraphics[width=\linewidth]{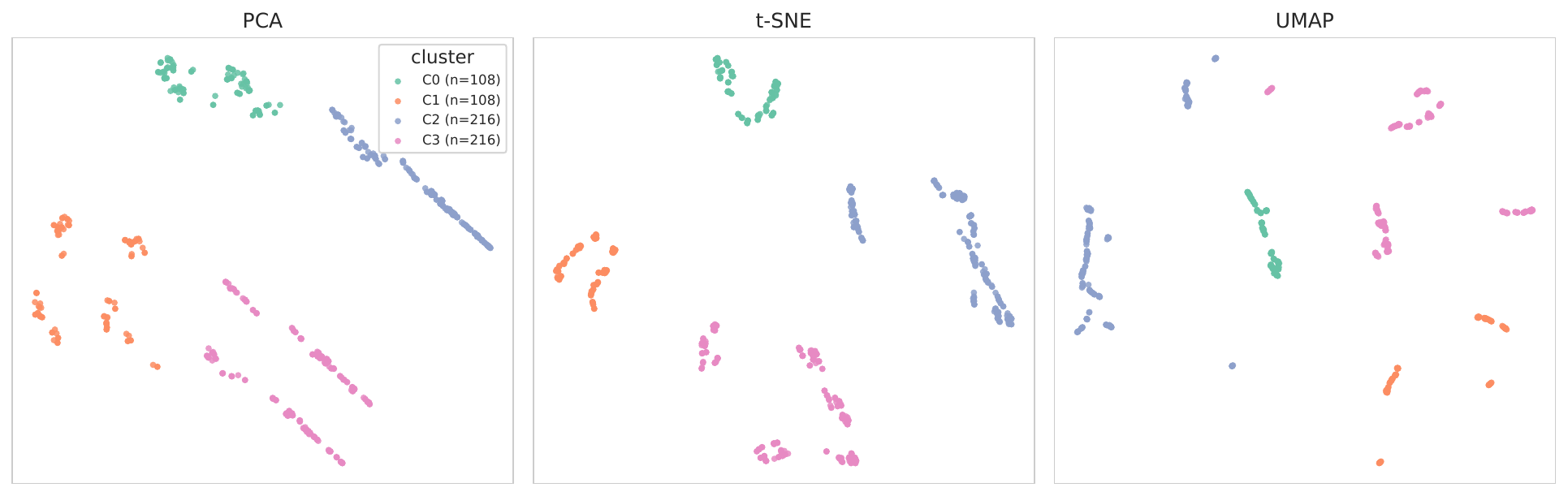}
\caption{The four clusters in two-dimensional projections of the standardized nine-dimensional feature space (PCA, $t$-SNE, UMAP). Colors denote the $k$-means cluster; the groups are well separated in all three projections.}
\label{fig:app:embeddings}
\end{figure}

The heavy-tailed features (edges, degrees, component count, and the size features) are $\log(1{+}x)$-transformed; all nine features are then standardized to zero mean and unit variance. We cluster with $k$-means (50 restarts) in the standardized feature space. The number of clusters is selected by jointly considering the silhouette coefficient, the Davies--Bouldin index, and the agreement between $k$-means and Ward agglomerative clustering measured by the adjusted Rand index (ARI); see Table~\ref{tab:app:kselect}. The silhouette is maximized at $k{=}4$ ($0.503$), the Davies--Bouldin index is near its minimum, and $k$-means and Ward clustering agree \emph{perfectly} at $k{=}4$ (ARI${=}1.000$). We therefore adopt $k{=}4$; the two- and three-dimensional projections in Figure~\ref{fig:app:embeddings} are used only for visualization. Clusters are indexed by increasing mean density.
 
\begin{table}[!htbp]
\centering
\small
\caption{Cluster-count selection. Silhouette (higher is better) peaks at $k{=}4$;
$k$-means and Ward clustering agree perfectly (ARI${=}1$) at $k{\le}4$.}
\label{tab:app:kselect}
\begin{tabular}{@{}rrrrr@{}}
\toprule
$k$ & Inertia & Silhouette $\uparrow$ & Davies--Bouldin $\downarrow$ & ARI ($k$-means vs.\ Ward) \\
\midrule
2 & 3142.3 & 0.408 & 1.047 & 0.693 \\
3 & 1968.7 & 0.460 & 0.873 & 1.000 \\
\textbf{4} & \textbf{1334.2} & \textbf{0.503} & \textbf{0.775} & \textbf{1.000} \\
5 & 1053.7 & 0.489 & 0.749 & 0.985 \\
6 &  863.1 & 0.470 & 0.822 & 0.936 \\
7 &  748.3 & 0.478 & 0.804 & 0.928 \\
8 &  648.3 & 0.480 & 0.806 & 0.902 \\
\bottomrule
\end{tabular}
\end{table}
 
\paragraph{Cluster structure.}
 
\begin{figure}[!htbp]
\centering
\includegraphics[width=0.86\linewidth]{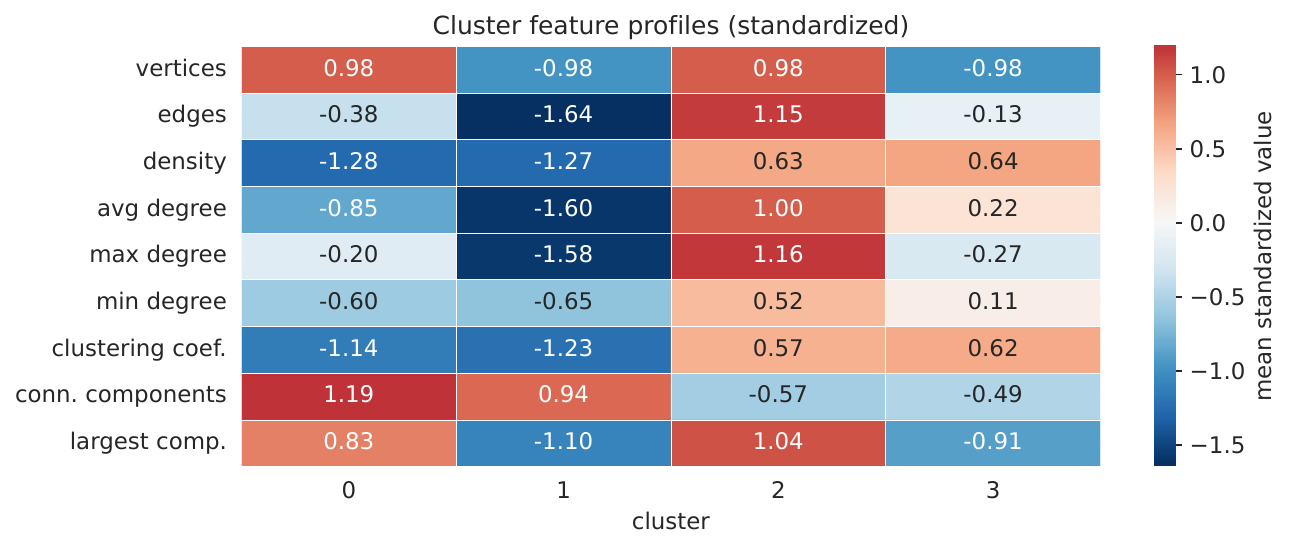}
\caption{Standardized feature profile of each cluster (mean $z$-score per feature).
Clusters 0/1 (sparse) have low density, low clustering, and many components; clusters
2/3 (dense) are the opposite. Within each density regime, clusters differ on the
size features (vertices, largest component), i.e.\ by benchmark.}
\label{fig:app:profiles}
\end{figure}

The four clusters form a clean $2\times2$ grid over exactly two factors: a
\emph{density regime} set by the threshold and a \emph{graph-size regime} set by the benchmark (Table~\ref{tab:app:clusters}, Figure~\ref{fig:app:profiles}). Sparse clusters (thresholds p10/p20) contain fragmented graphs with many connected components and low clustering; dense clusters (p50--p95) are near-fully connected.
Within each density regime, the small-benchmark cluster (GPQA/IFEval,
$\sim\!450$--$540$ vertices) separates from the large-benchmark cluster (MMLU-Pro/Omni-MATH, $1000$ vertices). The sparse clusters are half the size of the dense ones ($108$ vs.\ $216$) solely because the sparse regime spans two thresholds and the dense regime spans four. The projections in Figure~\ref{fig:app:embeddings} show the four groups as clearly separated in PCA, $t$-SNE, and UMAP, consistent with the high silhouette.
 
\begin{table}[!htbp]
\centering
\small
\caption{The four clusters as a $2\times2$ threshold\,$\times$\,benchmark grid.
Values are cluster means. Density and clustering are driven by the threshold;
vertex count is driven by the benchmark.}
\label{tab:app:clusters}
\begin{tabular}{@{}clcccll@{}}
\toprule
Cluster & Regime & $n$ & Density & Clust.\ coef. & Conn.\ comp. & Benchmarks / thresholds \\
\midrule
0 & sparse, large & 108 & 0.15 & 0.61 & 70 & MMLU-Pro, Omni-MATH / p10, p20 \\
1 & sparse, small & 108 & 0.15 & 0.60 & 35 & GPQA, IFEval / p10, p20 \\
2 & dense,\ \ large & 216 & 0.79 & 0.89 & \phantom{0}4 & MMLU-Pro, Omni-MATH / p50--p95 \\
3 & dense,\ \ small & 216 & 0.79 & 0.90 & \phantom{0}4 & GPQA, IFEval / p50--p95 \\
\bottomrule
\end{tabular}
\end{table}
 
\paragraph{Factor enrichment and invariance.}
 
\begin{figure}[!htbp]
\centering
\includegraphics[width=\linewidth]{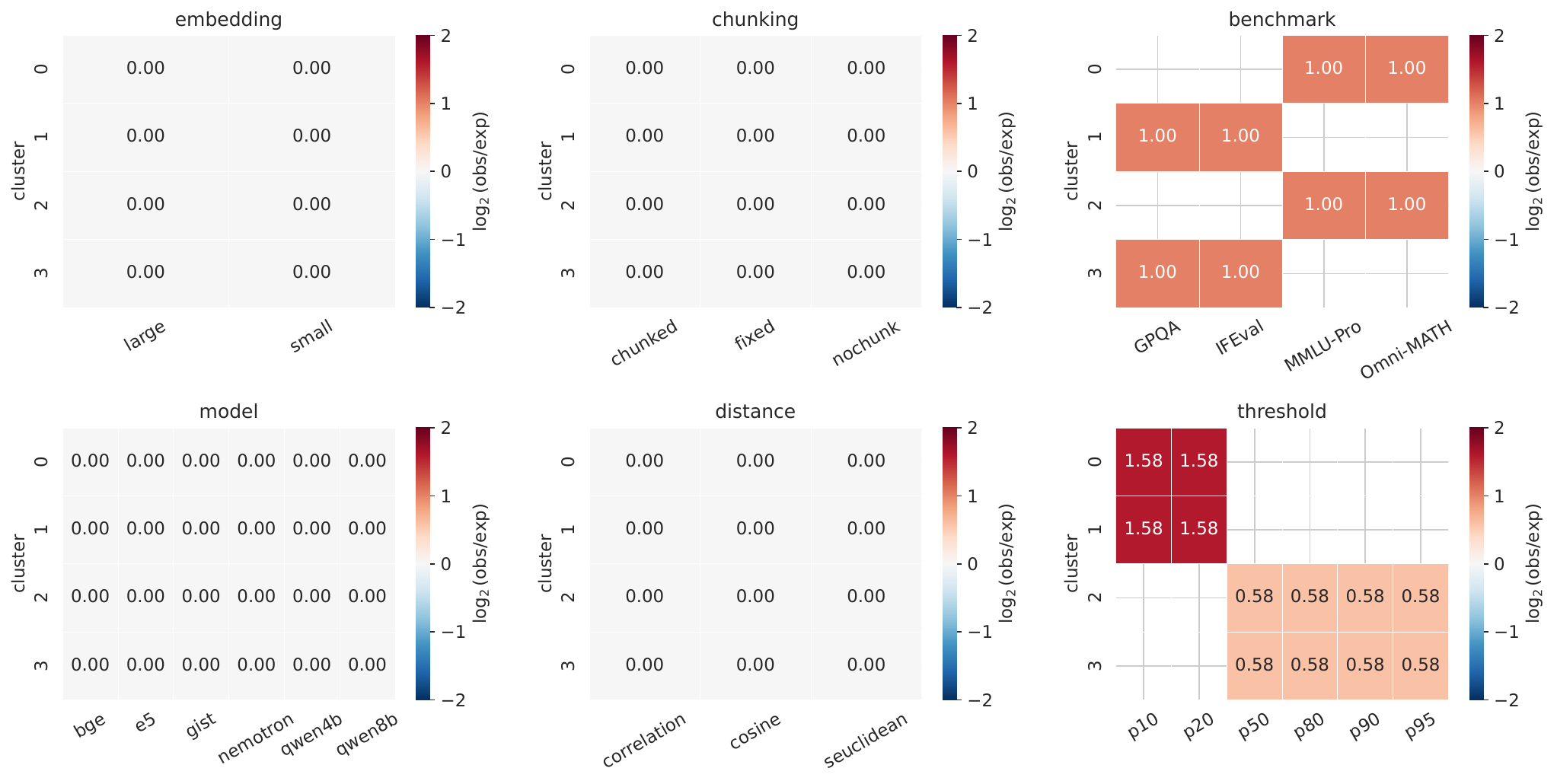}
\caption{Post-hoc factor enrichment per cluster, $\log_2(\text{obs}/\text{exp})$.
Red indicates over-representation, blue under-representation, and white indicates the
level is absent from that cluster. Only \emph{benchmark} and \emph{threshold}
discriminate the clusters; \emph{embedding}, \emph{chunking}, and \emph{model} are
flat at zero, indicating invariance.}
\label{fig:app:enrichment}
\end{figure}

\begin{figure}[!htbp]
\centering
\includegraphics[width=\linewidth]{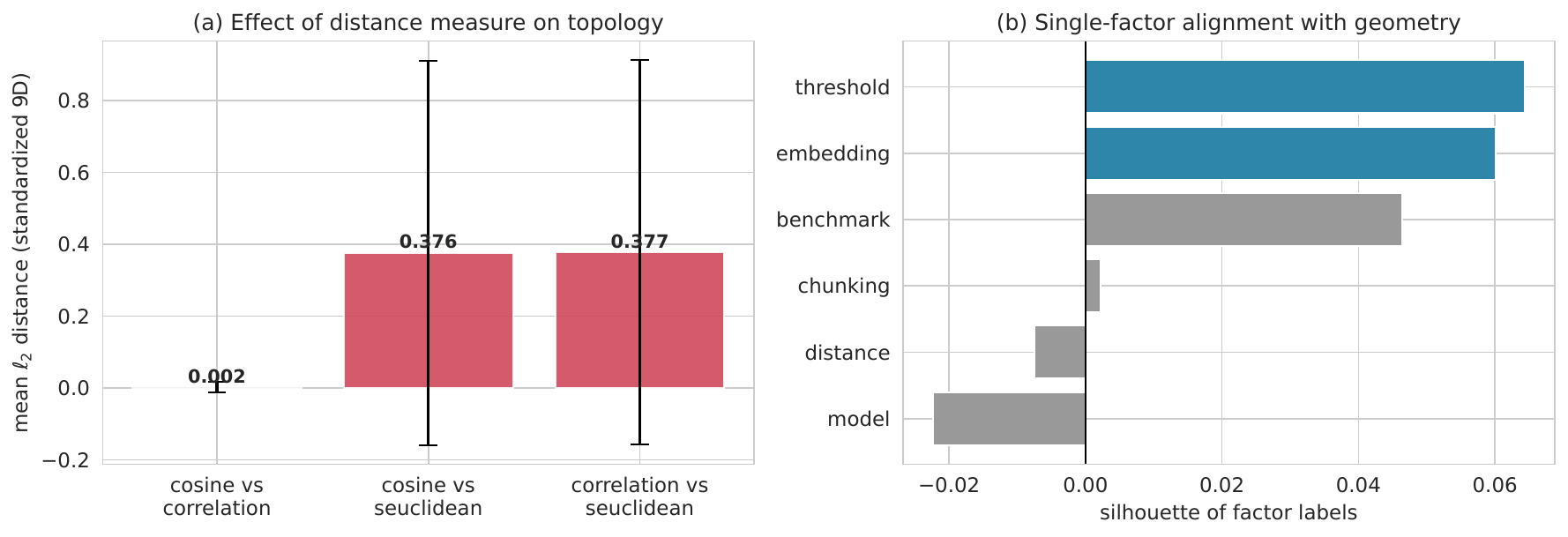}
\caption{(a) Mean $\ell_2$ separation between distance-measure variants of the same
cell in standardized feature space: \texttt{cosine} and \texttt{correlation} are
interchangeable, while \texttt{seuclidean} diverges. Error bars are standard
deviations over cells. (b) Silhouette of each factor's own labels in the feature
space; all values are small, indicating that no single factor explains the geometry
alone.}
\label{fig:app:factoreffects}
\end{figure}

We quantify the association between each factor and the clustering by the enrichment
$\log_2(\text{observed}/\text{expected})$ of every factor level in every cluster,
where the expected frequency is the level's overall frequency
(Figure~\ref{fig:app:enrichment}). Only \emph{benchmark} and \emph{threshold} deviate
from zero; the embedding size, chunking scheme, and embedding model are uniformly
distributed across all four clusters (enrichment $=0$ to two decimals everywhere).
Proximity-graph topology is therefore \emph{invariant} to these three factors at
the resolution of the clustering: changing the embedding size, chunking, or embedding
model does not move an instance across a cluster boundary. Consistent with this, no
single factor aligns strongly with the feature geometry on its own---the silhouette of
every factor's own labels is small (Figure~\ref{fig:app:factoreffects}b; threshold
$0.064$, embedding $0.060$, benchmark $0.046$, all others $\le 0.002$)---because the
observed structure is a \emph{joint} threshold\,$\times$\,benchmark-size partition
rather than a partition along any single axis.
 
\paragraph{Effect of the distance measure.}
 
The distance measure has a real but second-order effect that lives \emph{within} the
four coarse clusters rather than forming its own top-level split
(Figure~\ref{fig:app:factoreffects}a). Holding all other factors fixed and comparing
the three distance variants of each otherwise-identical cell in the standardized
feature space, \texttt{cosine} and \texttt{correlation} are almost indistinguishable
(mean $\ell_2$ distance $0.002$; they are assigned to the same cluster in every cell),
whereas \texttt{seuclidean} is a clear outlier (mean distance $0.376$--$0.377$ from the
other two). The distance measure thus perturbs the graph topology, but by an order of
magnitude less than the threshold or benchmark, which is why it does not appear in the
enrichment of the four clusters.
 
\paragraph{Interpretation and limitations.}
 
Because the vertex count is fixed by the benchmark and the density by the threshold,
the unsupervised clustering largely \emph{recovers two axes of the experimental
design} rather than uncovering latent structure orthogonal to it. The substantive
finding is therefore in what the clustering \emph{does not} separate: the embedding
size, chunking scheme, and embedding model produce proximity graphs that are
statistically indistinguishable at this resolution, and among the distance measures
only \texttt{seuclidean} behaves differently, with \texttt{cosine} and
\texttt{correlation} interchangeable. A scale-free analysis that normalizes out the
benchmark-driven size axis (e.g.\ clustering on density, clustering coefficient, and
size-normalized component statistics) could be used to test whether any finer
embedding- or distance-dependent structure emerges once the two dominant axes are
removed.

\appbarrier
\section{Distance Distribution Plots}
\label{app:data_driven}
\begin{figure}[!htbp]
    \centering
    \begin{subfigure}[b]{0.31\textwidth}\centering
        \includegraphics[width=\textwidth,height=95pt]{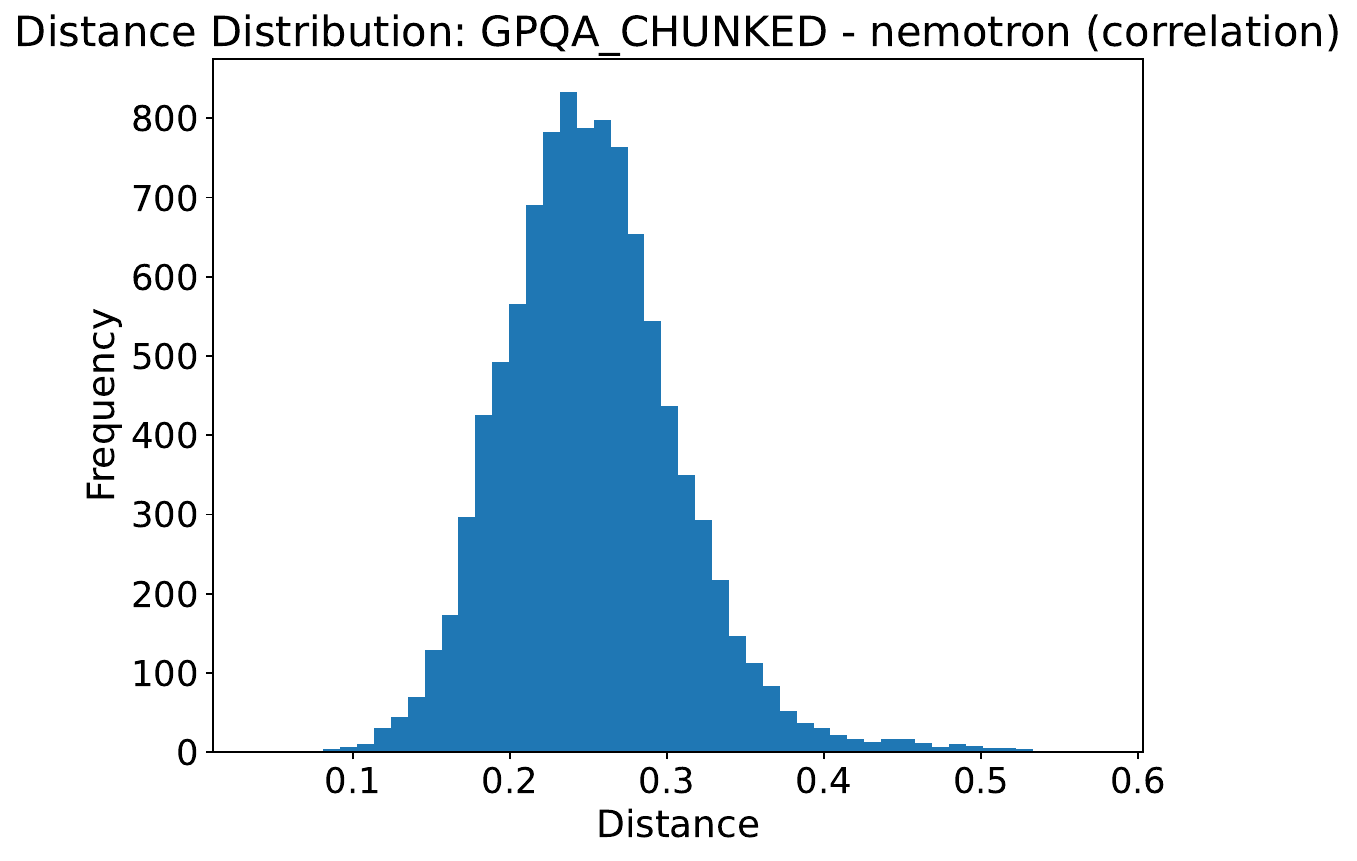}\caption{Nemotron}\end{subfigure}\hfill
    \begin{subfigure}[b]{0.31\textwidth}\centering
        \includegraphics[width=\textwidth,height=95pt]{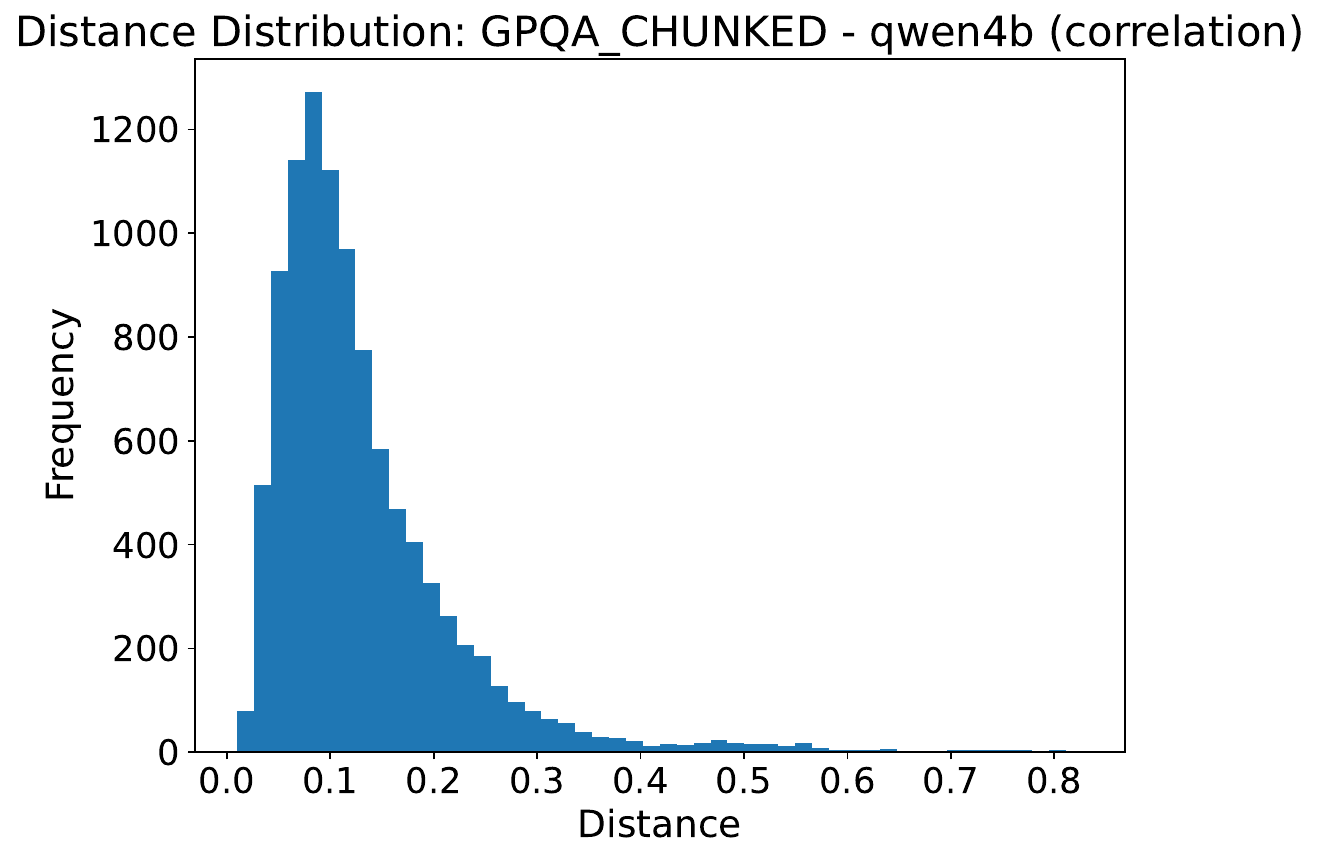}\caption{Qwen4B}\end{subfigure}\hfill
    \begin{subfigure}[b]{0.31\textwidth}\centering
        \includegraphics[width=\textwidth,height=95pt]{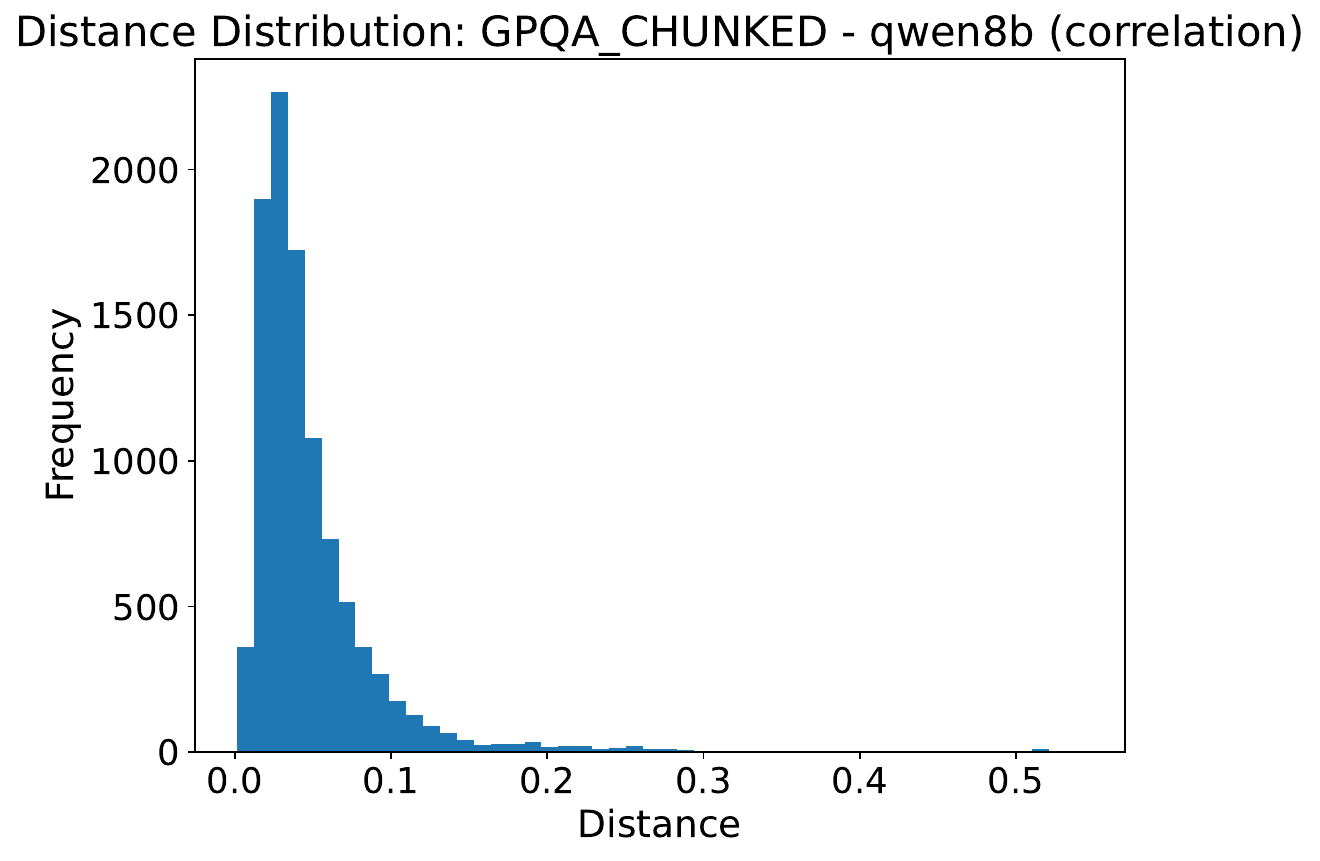}\caption{Qwen8B}\end{subfigure}
    \caption{Distance distributions for GPQA\_CHUNKED across models (correlation distance).}
    \label{fig:gpqa_distance_distributions}
\end{figure}

\begin{figure}[!htbp]
    \centering
    \begin{subfigure}[b]{0.31\textwidth}\centering
        \includegraphics[width=\textwidth,height=95pt]{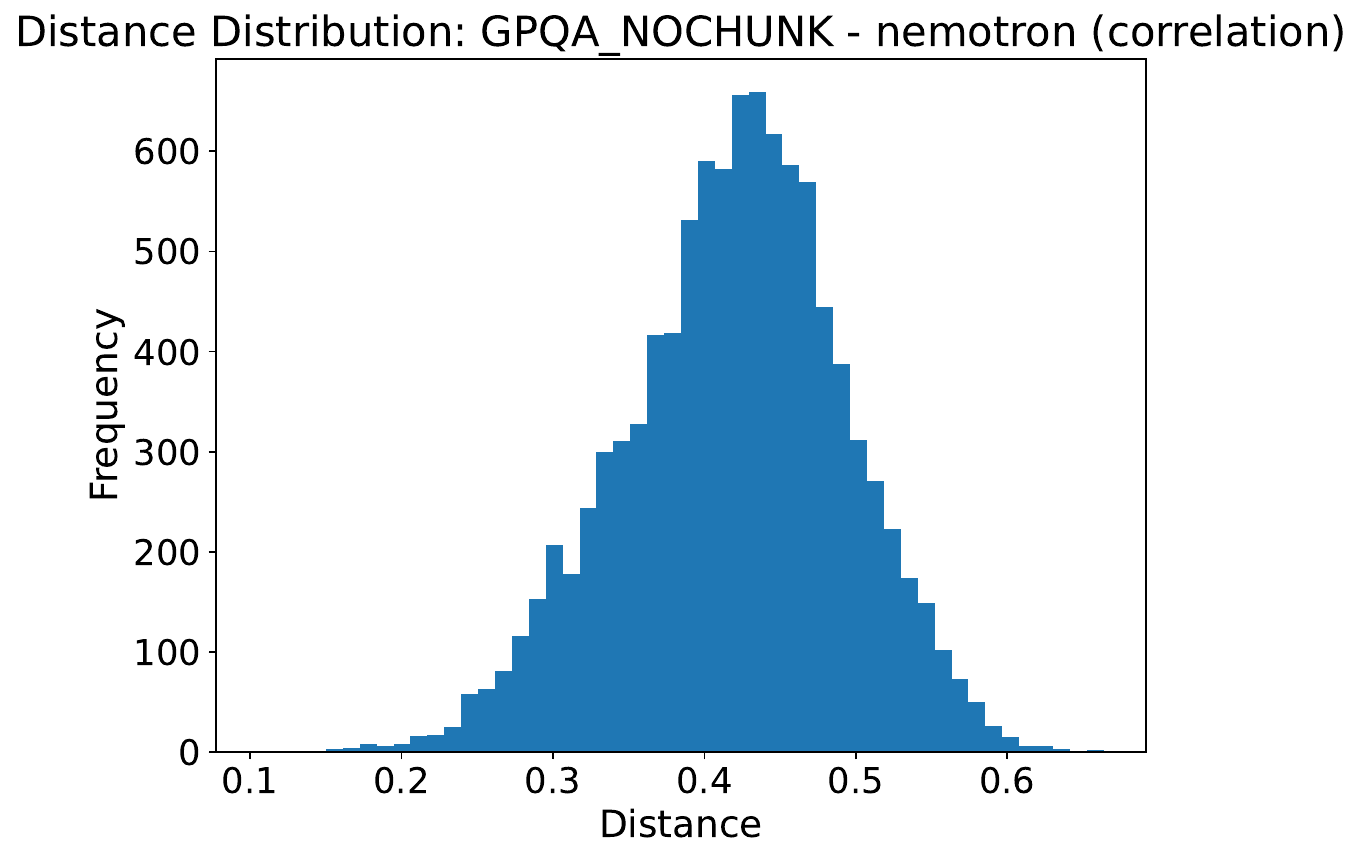}\caption{Nemotron}\end{subfigure}\hfill
    \begin{subfigure}[b]{0.31\textwidth}\centering
        \includegraphics[width=\textwidth,height=95pt]{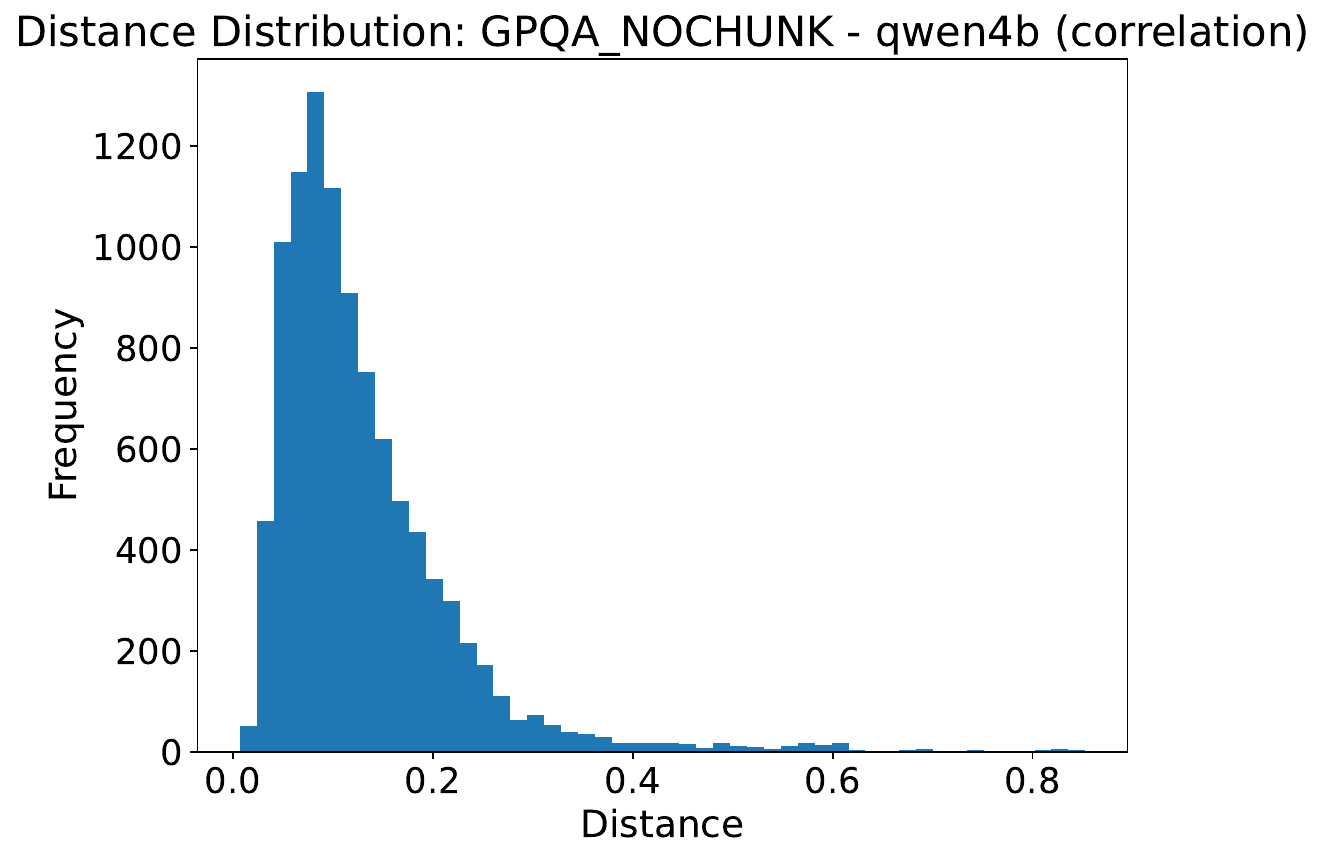}\caption{Qwen4B}\end{subfigure}\hfill
    \begin{subfigure}[b]{0.31\textwidth}\centering
        \includegraphics[width=\textwidth,height=95pt]{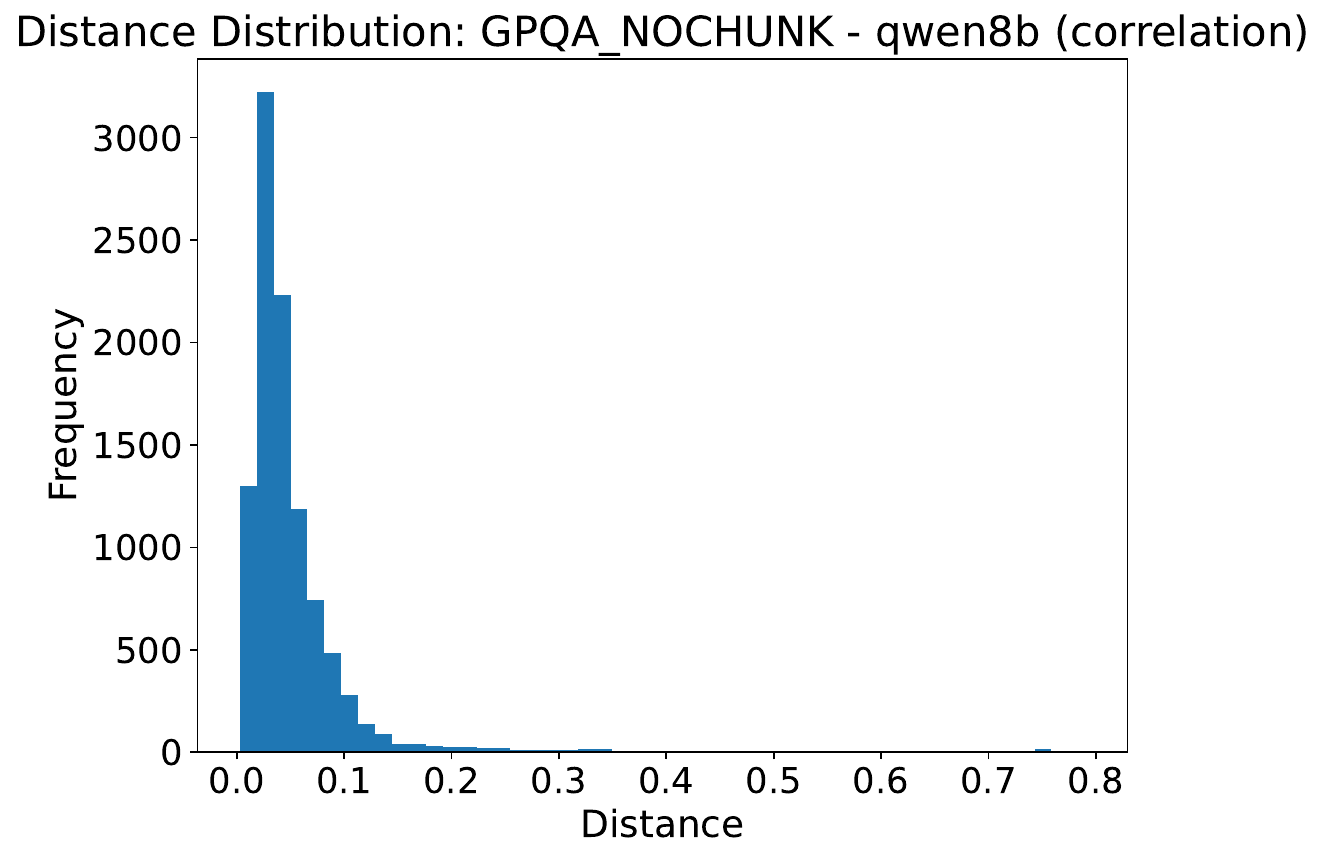}\caption{Qwen8B}\end{subfigure}
    \caption{Distance distributions for GPQA\_NOCHUNK across models (correlation distance).}
    \label{fig:nochunk_distance_distributions}
\end{figure}

\begin{figure}[!htbp]
    \centering
    \begin{subfigure}[b]{0.31\textwidth}\centering
        \includegraphics[width=\textwidth,height=95pt]{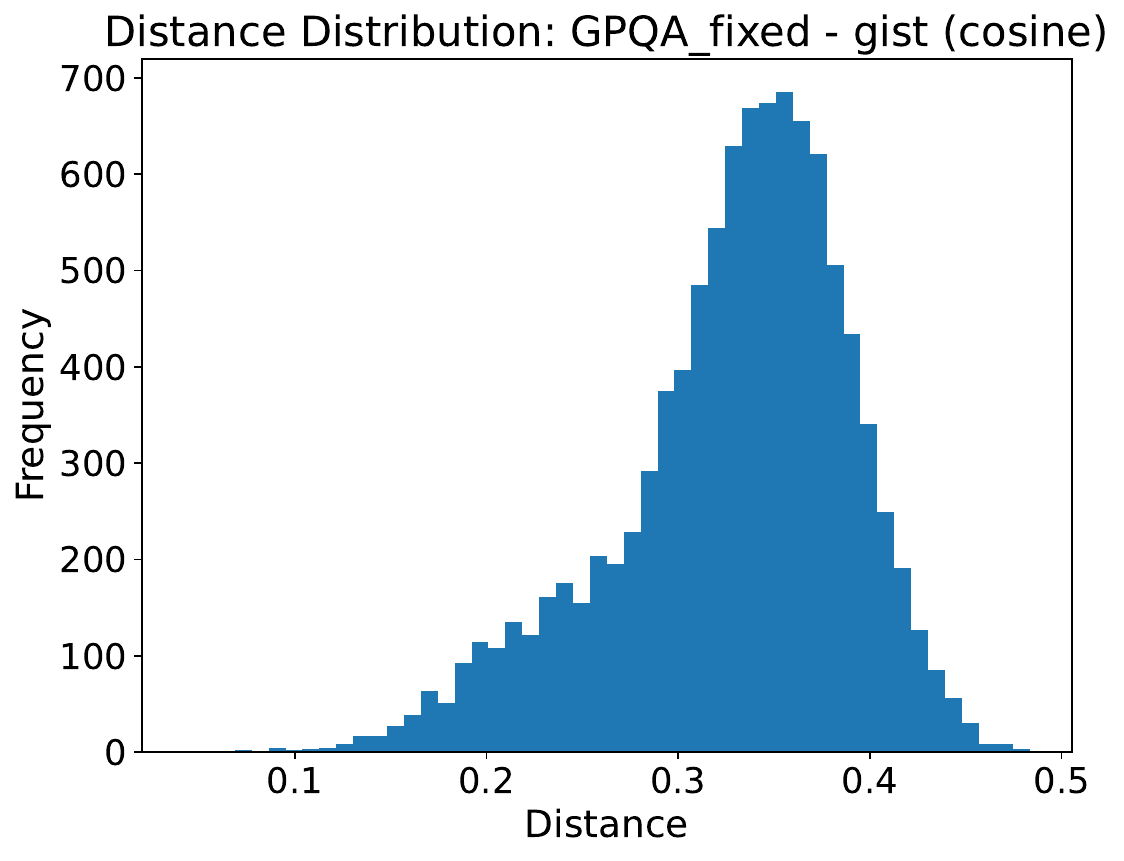}\caption{GIST}\end{subfigure}\hfill
    \begin{subfigure}[b]{0.31\textwidth}\centering
        \includegraphics[width=\textwidth,height=95pt]{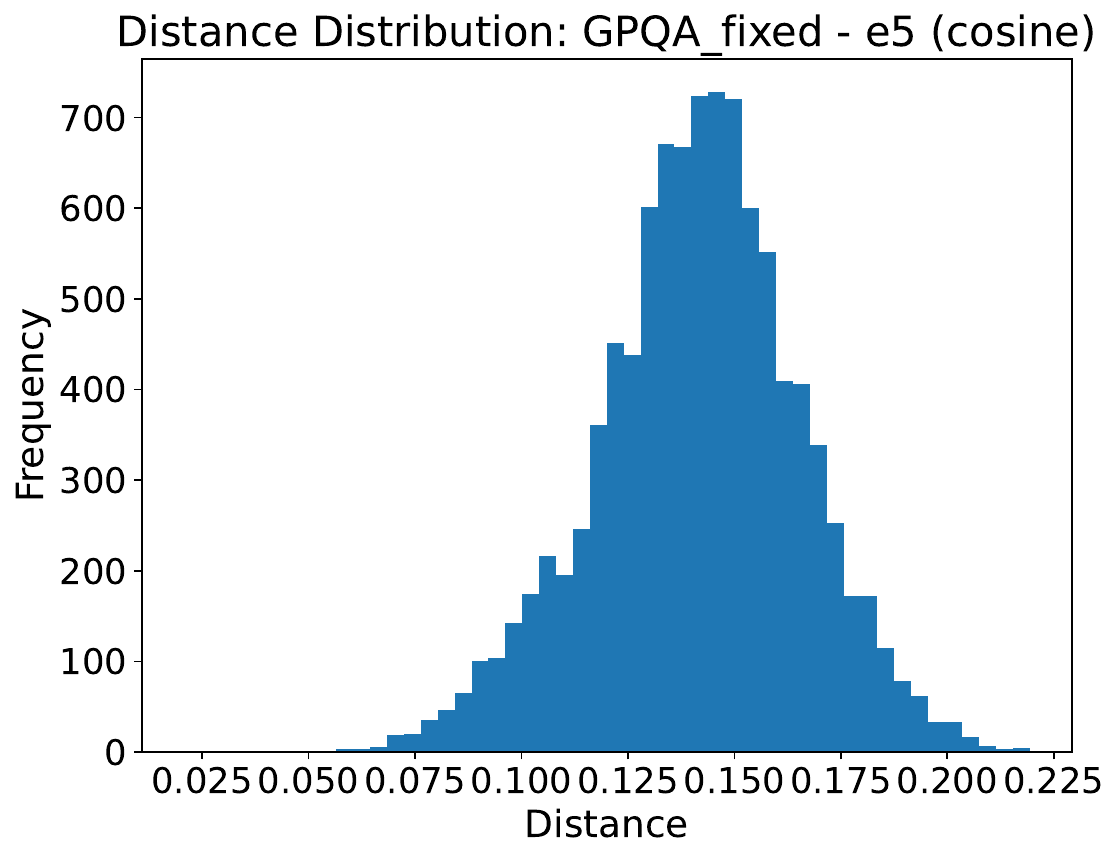}\caption{E5}\end{subfigure}\hfill
    \begin{subfigure}[b]{0.31\textwidth}\centering
        \includegraphics[width=\textwidth,height=95pt]{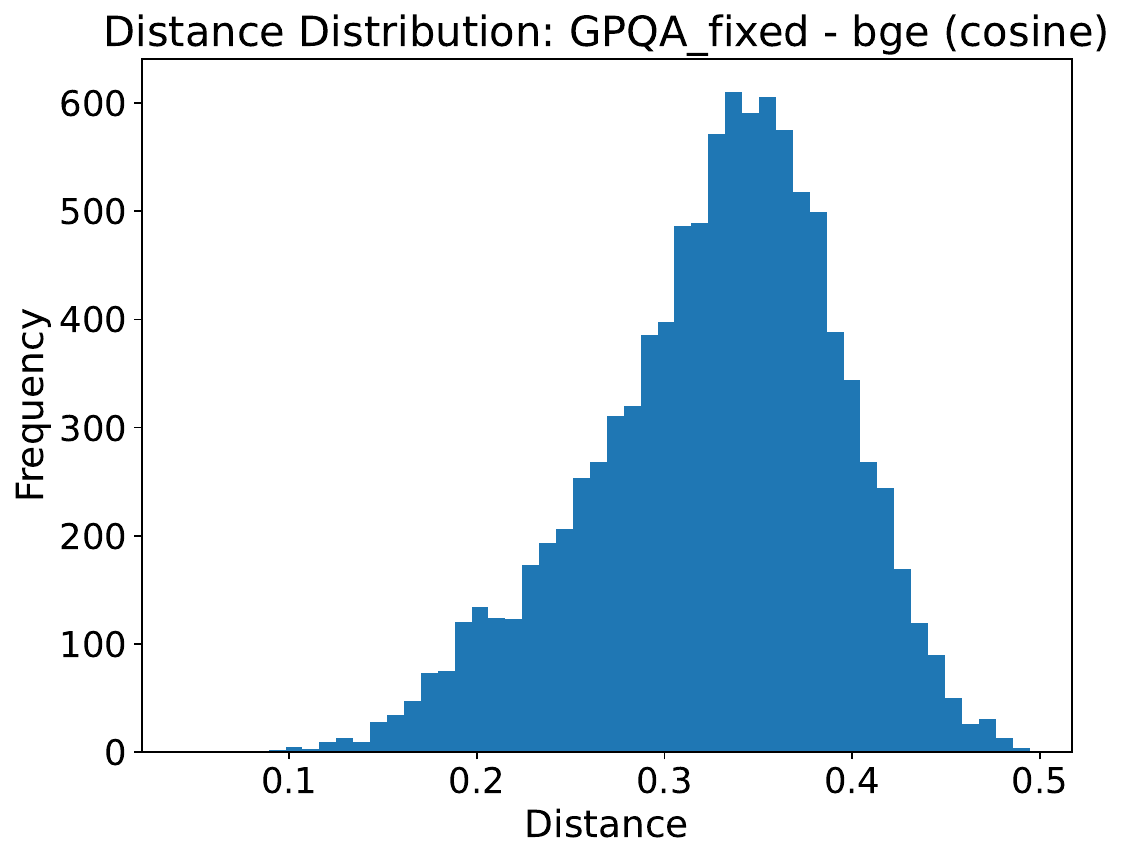}\caption{BGE}\end{subfigure}
    \caption{Distance distributions for GPQA across the small encoder models (cosine distance).}
    \label{fig:small_distance_distributions}
\end{figure}

Figures~\ref{fig:gpqa_distance_distributions}--\ref{fig:small_distance_distributions} show pairwise-distance distributions for GPQA under different embedding models and distance measures, motivating percentile-based thresholding over fixed absolute values.

\appbarrier
\section{Rank-Shift Heatmap}
\label{app:rank_shift}
\begin{figure}[!htbp]
  \centering
  \includegraphics[width=0.8\textwidth]{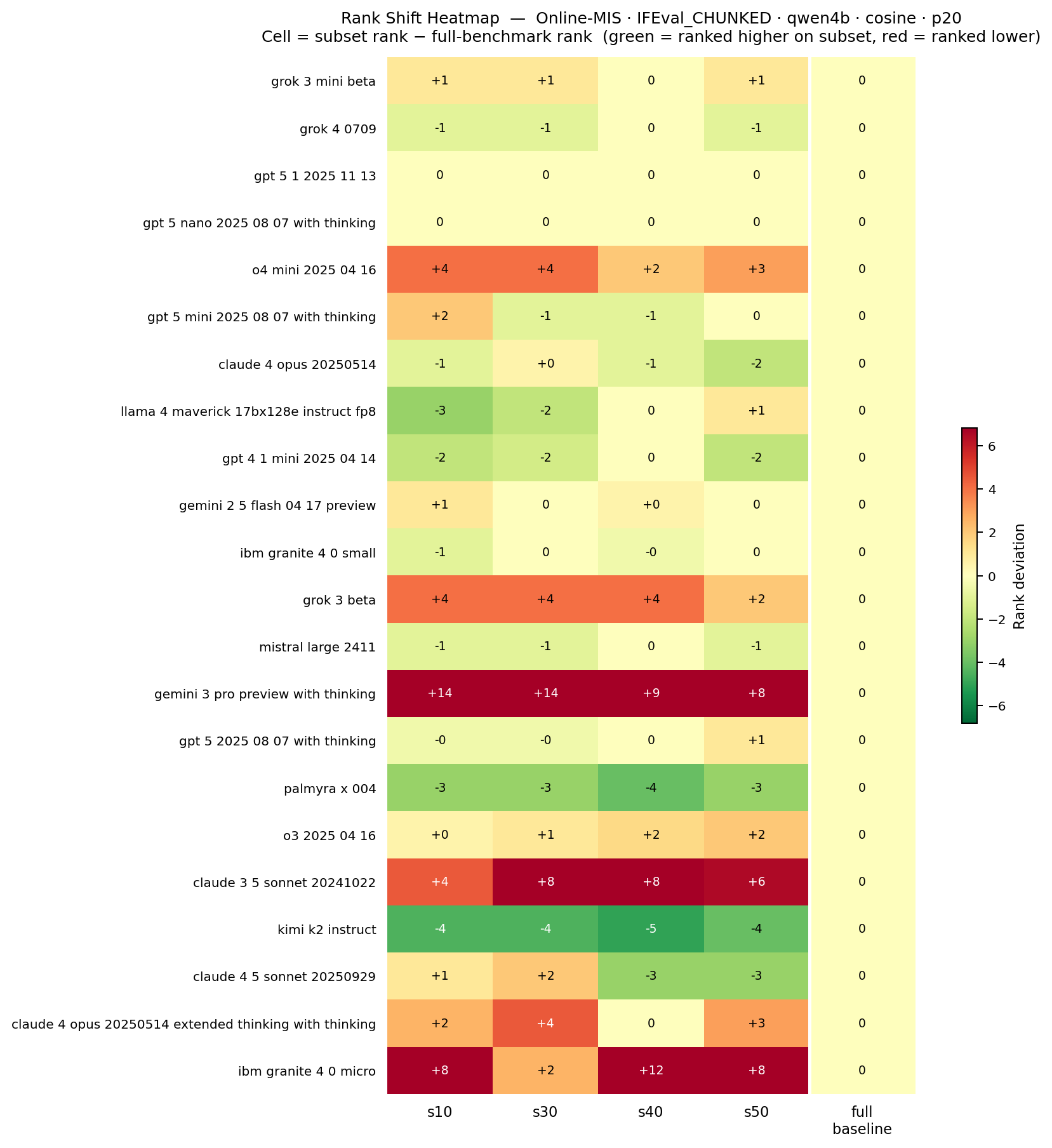}
  \caption{Rank-shift heatmap for Online-MIS on IFEval (\texttt{qwen4b}, cosine, $p_{20}$). Columns
  are the four completed seeds (the run with seed 20 did not complete); the last column is the full-benchmark baseline. Seeds agree closely
  ($W=0.994$) while the ranking differs from the baseline on a few models ($\rho=0.968$).}
  \label{fig:rankshift}
\end{figure}

Figure~\ref{fig:rankshift} presents the rank-shift heatmap for one example configuration (Online-MIS, IFEval, chunked, \texttt{qwen4b}, cosine, $p_{20}$). Each cell encodes the deviation of a model's subset rank from its full-benchmark rank (green = higher on the subset, red = lower); the rightmost column is the full baseline (zero by definition). The seed columns show near-identical patterns ($W=0.994$), while a few models shift relative to the baseline, giving $\rho=0.968$: the subset is highly stable across seeds yet re-weights a small number of models whose standing depends on the pruned region.

\appbarrier

\section{Embedding Proximity vs.\ Behavioral Similarity}
\label{app:redundancy}
\begin{figure}[!htbp]
  \centering
  \includegraphics[width=\textwidth]{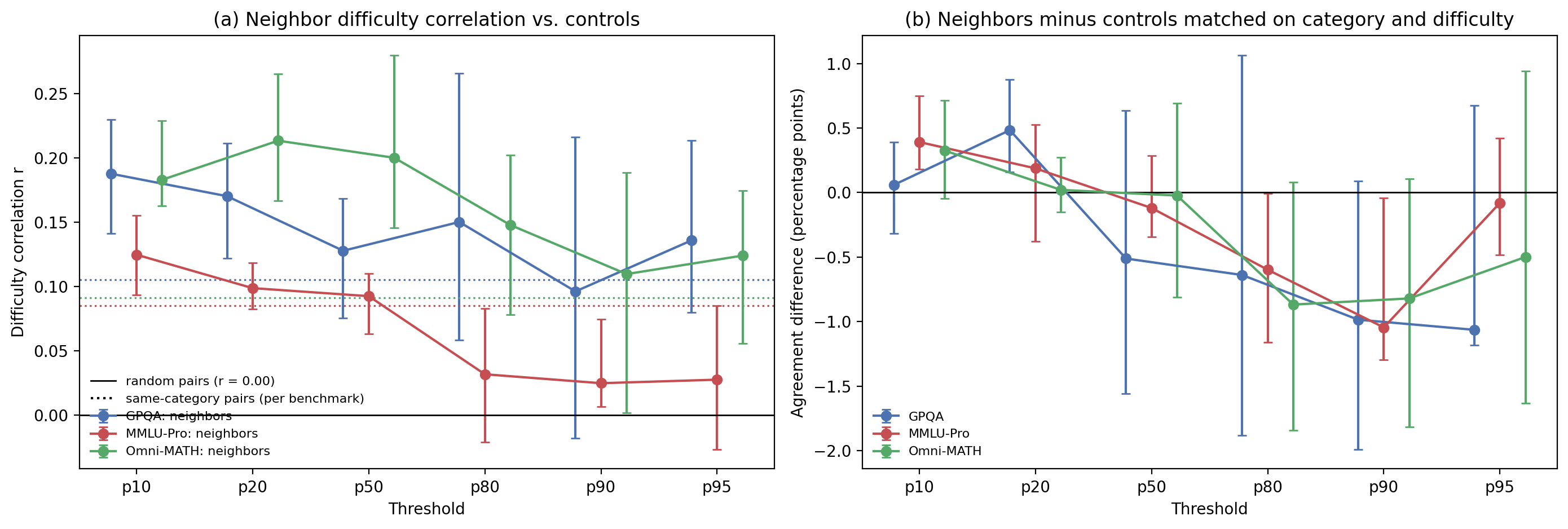}
  \caption{\textbf{Embedding neighbors vs.\ control pairs, all ReduMIS configurations.} (a)~Difficulty
  correlation between each discarded prompt and its nearest retained prompt by threshold (mean and
  interquartile range over configurations), compared with random pairs ($r=0$, solid) and random
  same-category pairs (dotted). (b)~Difference in agreement on model outcomes between neighbor pairs and
  control pairs matched on category and difficulty (percentage points). Neighbors are more similar in
  difficulty than random or same-category pairs, but not more similar in model outcomes once category
  and difficulty are matched.}
  \label{fig:redundancy_controls}
\end{figure}

This appendix extends the behavioral analysis of Section~\ref{sec:results} from a single configuration to all
$631$ ReduMIS configurations (six embeddings, three distances, six thresholds, four benchmarks). For every
configuration, each discarded prompt is paired with its nearest retained prompt under the same embedding and
distance as the selection. A prompt's \emph{difficulty} is the share of the $66$ LLMs that solve it (graded
outcomes binarized at $0.5$). We compare these neighbor pairs with three kinds of control pairs: random prompt
pairs, random pairs from the same category, and \emph{matched} pairs that keep the discarded prompt but replace
its neighbor by a random prompt from the same category with the neighbor's difficulty.
Figure~\ref{fig:redundancy_controls} summarizes the results; the tables below give all values, and all
per-configuration results are released with the code.

\paragraph{Difficulty of neighbors, by embedding and distance.} The difficulty correlation between neighbors is
positive in $89\%$ of all configurations and positive on average for every embedding and distance
(Table~\ref{tab:rev_red_emb}), whereas random prompt pairs show no correlation ($|r|\le0.01$ on every benchmark).
The correlations vary between configurations but not systematically: no embedding is consistently strongest
across benchmarks, large decoders and small encoders reach similar values at $p_{20}$ ($0.10$--$0.23$ vs.\
$0.10$--$0.21$ across the four benchmarks), and standardized Euclidean gives the highest or tied-highest
correlation on three of the four benchmarks. Without using labels, neighbors are as similar in difficulty as
random same-category pairs or more similar ($0.17$ vs.\ $0.11$ on GPQA, $0.10$ vs.\ $0.09$ on MMLU-Pro and
$0.21$ vs.\ $0.09$ on Omni-MATH at $p_{20}$). The alignment is strongest at mild thresholds and weakens at
aggressive thresholds, where neighbors are further apart (Table~\ref{tab:rev_red_thr},
Figure~\ref{fig:redundancy_controls}a).

\paragraph{Difficulty annotations.} Because the difficulty above is derived from the same $66$ LLMs that are
ranked, we repeated the analysis with the difficulty ratings that ship with the benchmarks and involve no model
evaluations (Table~\ref{tab:rev_red_ann}). On Omni-MATH, the official $1$--$10$ difficulty ratings of neighbors
correlate at $r=0.44$ at $p_{20}$, positively in every configuration and for every embedding. On GPQA, the
question writers' four-level ratings correlate at $r=0.22$ (positively in all $p_{10}$ and $p_{20}$
configurations), whereas the coarser three-level validator rating shows only a weak correlation ($r=0.05$).
Embedding proximity thus captures difficulty structure that benchmark authors annotated independently of any
model. MMLU-Pro and IFEval have no difficulty annotations.

\paragraph{Proximity and behavior across all prompt pairs.} Within the narrow distance band of retained--discarded
neighbors, the exact distance is only weakly related to behavioral similarity, because the threshold truncates
the range of distances. Across \emph{all} prompt pairs, closer prompts elicit more similar model behavior beyond
what their difficulty explains (excess agreement: the share of LLMs with the same outcome on both prompts minus
the agreement expected from their difficulties) for $19$ of $24$ embedding--benchmark combinations
(Table~\ref{tab:rev_red_allpairs}). The effect is clearest on MMLU-Pro ($r=0.07$--$0.23$) and small on the other
benchmarks.

\paragraph{Matched controls.} Neighbors share a category far more often than random prompt pairs ($95\%$ vs.\
$37\%$ on GPQA, $59\%$ vs.\ $8\%$ on MMLU-Pro and $60\%$ vs.\ $23\%$ on Omni-MATH at $p_{20}$), so part of their
similarity may simply reflect shared category and difficulty. To isolate the effect of proximity, we compare each
neighbor pair with a control pair that keeps the same discarded prompt and replaces the neighbor by a random
prompt from the same category with the neighbor's difficulty (mean absolute difficulty difference $0.004$). Once
category and difficulty are matched, neighbors agree on model outcomes as often as the controls
(Table~\ref{tab:rev_red_matched}; at $p_{20}$: $58.2\%$ vs.\ $57.7\%$ on GPQA, $64.2\%$ vs.\ $64.0\%$ on MMLU-Pro
and $62.5\%$ vs.\ $62.5\%$ on Omni-MATH), with differences within $\pm1.3$ percentage points for every embedding
(Table~\ref{tab:rev_red_matched_emb}). The small positive difference at mild thresholds turns slightly negative at
aggressive thresholds (Figure~\ref{fig:redundancy_controls}b).

\paragraph{Interpretation.} Embedding proximity captures category and difficulty structure without labels or
model evaluations, and this structure accounts for the behavioral similarity of neighbors, including the
similarity beyond difficulty observed across all pairs. It does not, however, identify prompts that are
interchangeable for evaluation beyond what category and difficulty explain. We therefore describe MIS as removing
\emph{semantic} redundancy in the chosen embedding space; its effect on the subset is the coverage redistribution
reported in Section~\ref{sec:results}.

\paragraph{Figure~\ref{fig:redundancy_difficulty}.} The single configuration shown in the main text
(\texttt{qwen8b}, standardized Euclidean, $p_{20}$, no chunking) gives $r=0.18$, $0.18$ and $0.29$ on GPQA,
MMLU-Pro and Omni-MATH; the chunked variant of the same configuration gives $0.13$, $0.13$ and $0.21$. Both lie
within the range of the full analysis above.

\begin{table}[!htbp]
\centering
\caption{Difficulty correlation between each discarded prompt and its nearest retained prompt, by embedding and distance ($p_{20}$, ReduMIS).}
\label{tab:rev_red_emb}
\small
\begin{tabular}{llrrrr}
\toprule
Embedding & Distance & GPQA & IFEval & MMLU-Pro & Omni-MATH \\
\midrule
bge (S) & cosine & 0.20 & 0.21 & 0.10 & 0.06 \\
 & correlation & 0.20 & 0.20 & 0.09 & 0.05 \\
 & std. Euclidean & 0.22 & 0.21 & 0.09 & 0.16 \\
\midrule
e5 (S) & cosine & 0.32 & 0.06 & 0.11 & 0.25 \\
 & correlation & 0.32 & 0.11 & 0.09 & 0.25 \\
 & std. Euclidean & 0.22 & 0.12 & 0.08 & 0.21 \\
\midrule
gist (S) & cosine & 0.12 & -0.05 & 0.10 & 0.23 \\
 & correlation & 0.12 & -0.05 & 0.15 & 0.22 \\
 & std. Euclidean & 0.17 & 0.04 & 0.08 & 0.18 \\
\midrule
nemotron (L) & cosine & 0.21 & 0.14 & 0.14 & 0.18 \\
 & correlation & 0.23 & 0.16 & 0.13 & 0.17 \\
 & std. Euclidean & 0.23 & 0.11 & 0.16 & 0.18 \\
\midrule
qwen4b (L) & cosine & 0.15 & 0.02 & 0.01 & 0.28 \\
 & correlation & 0.13 & 0.04 & 0.05 & 0.29 \\
 & std. Euclidean & 0.08 & 0.15 & 0.06 & 0.33 \\
\midrule
qwen8b (L) & cosine & 0.05 & 0.08 & 0.10 & 0.15 \\
 & correlation & 0.05 & 0.08 & 0.07 & 0.24 \\
 & std. Euclidean & 0.15 & 0.14 & 0.15 & 0.25 \\
\midrule
Control & random prompt pairs & 0.00 & -0.01 & 0.00 & 0.00 \\
 & random same-category pairs & 0.11 & -- & 0.09 & 0.09 \\
\bottomrule
\end{tabular}
\\[3pt]{\footnotesize Difficulty = share of the 66 LLMs solving the prompt. Large-embedding values averaged over chunked and non-chunked variants. S = small encoder, L = large decoder. Omni-MATH: eight top-level categories. IFEval has no category labels.}
\end{table}

\begin{table}[!htbp]
\centering
\caption{Same correlation using the benchmarks’ own difficulty annotations (no model evaluations), averaged over all configurations.}
\label{tab:rev_red_ann}
\small
\resizebox{\textwidth}{!}{%
\begin{tabular}{lrrrrrr}
\toprule
Threshold & GPQA writer rating: r & configs r $>$ 0 & GPQA validator rating: r & configs r $>$ 0 & Omni-MATH rating: r & configs r $>$ 0 \\
\midrule
$p_{10}$ & 0.27 & 100\% & 0.05 & 73\% & 0.46 & 100\% \\
$p_{20}$ & 0.22 & 100\% & 0.05 & 96\% & 0.44 & 100\% \\
$p_{50}$ & 0.11 & 96\% & 0.06 & 96\% & 0.34 & 100\% \\
$p_{80}$ & 0.10 & 85\% & 0.06 & 81\% & 0.28 & 100\% \\
$p_{90}$ & 0.07 & 88\% & 0.02 & 65\% & 0.26 & 100\% \\
$p_{95}$ & 0.07 & 85\% & 0.05 & 81\% & 0.26 & 100\% \\
Random pairs & 0.00 &  & -0.01 &  & 0.00 &  \\
\bottomrule
\end{tabular}}
\\[3pt]{\footnotesize At $p_{20}$, the per-embedding range is 0.13--0.25 for the GPQA writer rating and 0.31--0.51 for Omni-MATH. MMLU-Pro and IFEval have no difficulty annotations.}
\end{table}

\begin{table}[!htbp]
\centering
\caption{Difficulty correlation by threshold (mean over embeddings, distances and chunking variants).}
\label{tab:rev_red_thr}
\small
\begin{tabular}{lrrrrr}
\toprule
Threshold & GPQA & IFEval & MMLU-Pro & Omni-MATH & Configurations with r $>$ 0 \\
\midrule
$p_{10}$ & 0.19 & 0.12 & 0.12 & 0.18 & 99\% \\
$p_{20}$ & 0.17 & 0.10 & 0.10 & 0.21 & 97\% \\
$p_{50}$ & 0.13 & 0.11 & 0.09 & 0.20 & 97\% \\
$p_{80}$ & 0.15 & 0.09 & 0.03 & 0.15 & 83\% \\
$p_{90}$ & 0.10 & 0.09 & 0.02 & 0.11 & 78\% \\
$p_{95}$ & 0.14 & 0.06 & 0.03 & 0.12 & 81\% \\
\bottomrule
\end{tabular}
\end{table}

\begin{table}[!htbp]
\centering
\caption{Correlation between embedding proximity and behavioral similarity (agreement beyond what difficulty predicts) across all prompt pairs.}
\label{tab:rev_red_allpairs}
\small
\begin{tabular}{llrrrr}
\toprule
Embedding & Distance & GPQA & IFEval & MMLU-Pro & Omni-MATH \\
\midrule
bge (S) & cosine & 0.00 & 0.04 & 0.17 & 0.02 \\
 & correlation & 0.00 & 0.04 & 0.17 & 0.02 \\
 & std. Euclidean & -0.01 & 0.05 & 0.17 & 0.02 \\
\midrule
e5 (S) & cosine & -0.01 & 0.06 & 0.18 & 0.10 \\
 & correlation & -0.01 & 0.06 & 0.18 & 0.10 \\
 & std. Euclidean & -0.02 & 0.06 & 0.18 & 0.10 \\
\midrule
gist (S) & cosine & -0.01 & 0.05 & 0.09 & 0.02 \\
 & correlation & -0.01 & 0.05 & 0.09 & 0.02 \\
 & std. Euclidean & -0.02 & 0.04 & 0.07 & 0.02 \\
\midrule
nemotron (L) & cosine & 0.05 & 0.04 & 0.22 & 0.11 \\
 & correlation & 0.05 & 0.04 & 0.22 & 0.11 \\
 & std. Euclidean & 0.05 & 0.04 & 0.23 & 0.11 \\
\midrule
qwen4b (L) & cosine & 0.12 & 0.02 & 0.10 & 0.00 \\
 & correlation & 0.12 & 0.02 & 0.10 & 0.00 \\
 & std. Euclidean & 0.14 & 0.05 & 0.13 & 0.00 \\
\midrule
qwen8b (L) & cosine & 0.11 & 0.02 & 0.08 & -0.01 \\
 & correlation & 0.11 & 0.02 & 0.08 & -0.01 \\
 & std. Euclidean & 0.15 & 0.05 & 0.09 & -0.04 \\
\bottomrule
\end{tabular}
\\[3pt]{\footnotesize Excess agreement = share of the 66 LLMs with the same outcome on both prompts minus the agreement expected from the two prompts’ difficulties.}
\end{table}

\begin{table}[!htbp]
\centering
\caption{Agreement on model outcomes: embedding neighbors vs. control pairs matched on category and difficulty.}
\label{tab:rev_red_matched}
\small
\resizebox{\textwidth}{!}{%
\begin{tabular}{llrrrr}
\toprule
Benchmark & Thr. & Neighbors in same category & Agreement: neighbors & Agreement: matched controls & Difference (pts) \\
\midrule
GPQA & $p_{10}$ & 97\% & 58.5\% & 58.4\% & +0.1 \\
 & $p_{20}$ & 95\% & 58.2\% & 57.7\% & +0.5 \\
 & $p_{50}$ & 90\% & 56.7\% & 57.2\% & -0.5 \\
 & $p_{80}$ & 81\% & 56.6\% & 57.3\% & -0.6 \\
 & $p_{90}$ & 76\% & 56.4\% & 57.4\% & -1.0 \\
 & $p_{95}$ & 70\% & 56.1\% & 57.2\% & -1.1 \\
\midrule
MMLU-Pro & $p_{10}$ & 63\% & 65.6\% & 65.3\% & +0.4 \\
 & $p_{20}$ & 59\% & 64.2\% & 64.0\% & +0.2 \\
 & $p_{50}$ & 49\% & 65.3\% & 65.4\% & -0.1 \\
 & $p_{80}$ & 36\% & 63.2\% & 63.8\% & -0.6 \\
 & $p_{90}$ & 27\% & 63.2\% & 64.3\% & -1.0 \\
 & $p_{95}$ & 25\% & 63.8\% & 63.9\% & -0.1 \\
\midrule
Omni-MATH & $p_{10}$ & 61\% & 63.4\% & 63.1\% & +0.3 \\
 & $p_{20}$ & 60\% & 62.5\% & 62.5\% & 0.0 \\
 & $p_{50}$ & 54\% & 62.8\% & 62.8\% & 0.0 \\
 & $p_{80}$ & 47\% & 59.8\% & 60.7\% & -0.9 \\
 & $p_{90}$ & 46\% & 59.4\% & 60.2\% & -0.8 \\
 & $p_{95}$ & 41\% & 60.8\% & 61.3\% & -0.5 \\
\bottomrule
\end{tabular}}
\\[3pt]{\footnotesize Control: same discarded prompt paired with a random same-category prompt whose difficulty matches the retained neighbor (mean absolute difficulty difference 0.004). Same-category rate for random prompt pairs: 37\% (GPQA), 8\% (MMLU-Pro), 23\% (Omni-MATH).}
\end{table}

\begin{table}[!htbp]
\centering
\caption{Difference in agreement (neighbors minus matched controls, percentage points) by embedding at $p_{20}$.}
\label{tab:rev_red_matched_emb}
\small
\begin{tabular}{lrrr}
\toprule
Embedding & GPQA & MMLU-Pro & Omni-MATH \\
\midrule
bge (S) & +0.4 & +0.4 & +0.2 \\
e5 (S) & +0.8 & +0.9 & -0.2 \\
gist (S) & +0.4 & +0.1 & 0.0 \\
nemotron (L) & +0.6 & +0.5 & -0.4 \\
qwen4b (L) & +1.3 & -0.4 & +0.1 \\
qwen8b (L) & -0.4 & 0.0 & +0.3 \\
\bottomrule
\end{tabular}
\end{table}

\appbarrier
\section{Computational Details}
\label{app:compute}
\textbf{Large embedding generation.} Dense representations were generated with Qwen3-8B, Qwen3-4B, and Nemotron-8B in float16 via HuggingFace Transformers. Each prompt was embedded under two conditions. \emph{No-chunk:} hard truncation at $4{,}096$ tokens, single forward pass, mean pooling over non-padding positions, L2 normalization. \emph{Chunked:} overlapping windows of $510$ tokens (stride $256$, $50\%$ overlap); each chunk embedded independently and combined as a length-weighted average, then L2-normalized. For Nemotron-8B each prompt was prefixed with ``\texttt{Instruct: Represent the following text for clustering: }''; no prefix was used for Qwen models. The full pipeline over all four datasets, both conditions, and all three models completed in 52\,min\,42\,s of wall-clock time on a node with $2\times$ NVIDIA L40S GPUs, a 64-core AMD CPU, and $\sim$503\,GB RAM (SLING HPC).

\textbf{Small embedding generation.} We additionally used three dedicated sentence encoders: BGE-large-en-v1.5, E5-large, and GIST-small-Embedding-v0, run locally (Intel Core Ultra 7 258V, 32\,GB RAM, Intel Arc 140V, 16\,GB VRAM). Task-specific prefixes were used per model (BGE: ``\texttt{Represent this sentence for searching relevant passages: }''; E5: ``\texttt{query: }''; GIST: none). Non-overlapping $510$-token chunks were used given the $512$-token context limit, with length-weighted aggregation and L2 normalization.

\textbf{Optimization experiments.} MIS experiments ran on the VEGA EuroHPC system (Atos BullSequana XH2000; dual-socket AMD EPYC 7H12 @ 2.6\,GHz; 256\,GB--1\,TB RAM; HDR100 InfiniBand; SLURM). All MIS runs were single-threaded for fair comparison, with $32$\,GB RAM and a $600$\,s time limit per job; stochastic solvers used five seeds. Table~\ref{tab:rev_solvers} compares the four solvers on the recommended configuration.

\begin{table}[!htbp]
\centering
\caption{Recommended configuration (qwen8b / std. Euclidean / $p_{20}$ / chunked) with each solver.}
\label{tab:rev_solvers}
\small
\begin{tabular}{llrrr}
\toprule
Benchmark & Solver & Subset size & $\rho$ (mean over seeds) & $\rho$ SD across seeds \\
\midrule
GPQA & CPLEX & 116 & 0.975 & -- \\
 & ReduMIS & 116 & 0.975 & 0.0000 \\
 & Online-MIS & 116 & 0.975 & 0.0032 \\
 & GREEDY & 115 & 0.976 & -- \\
\midrule
IFEval & CPLEX & 200 & 0.977 & -- \\
 & ReduMIS & 200 & 0.976 & 0.0000 \\
 & Online-MIS & 200 & 0.975 & 0.0016 \\
 & GREEDY & 198 & 0.974 & -- \\
\midrule
MMLU-Pro & CPLEX & 284 & 0.986 & -- \\
 & ReduMIS & 284 & 0.985 & 0.0000 \\
 & Online-MIS & 284 & 0.985 & 0.0005 \\
 & GREEDY & 284 & 0.982 & -- \\
\midrule
Omni-MATH & CPLEX & 332 & 0.997 & -- \\
 & ReduMIS & 332 & 0.996 & 0.0000 \\
 & Online-MIS & 332 & 0.997 & 0.0004 \\
 & GREEDY & 328 & 0.997 & -- \\
\bottomrule
\end{tabular}
\\[3pt]{\footnotesize CPLEX and GREEDY are deterministic (one run).}
\end{table}

\appbarrier

\section{Comparison with Clustering and Response-based Baselines}
\label{app:rev_selectors}

\noindent\textbf{Comparison to clustering baseline.} We evaluated a clustering-based sampling strategy that uses the structure
of the embedding space to determine which instances to retain. For each
combination of benchmark, embedding model, and chunking setting, we
normalized the embeddings to unit length, reduced them to 50 dimensions
using principal component analysis (PCA), and applied $k$-means clustering.
We selected $k \in \{2,\ldots,20\}$ by maximizing the silhouette score.
We retained all instances in the smallest cluster and sampled the same
number of instances uniformly without replacement from each remaining
cluster. We repeated the sampling procedure five times for each
configuration. The clustering baseline achieved a mean Spearman's $\rho$ of $0.993$
relative to the full-benchmark ranking of the 66 LLMs and a mean
Kendall's $W$ of $0.993$ across the five draws. The resulting subsets
retained between 29\% and 100\% of the original instances, with 15 of the
36 configurations retaining at least 85\%. Silhouette-based selection
often yielded two clusters of similar size. In these cases, using the
smallest cluster as the per-cluster quota retained nearly the entire
benchmark. Both the retained fraction and the agreement with the
full-benchmark ranking may depend on the criterion used to select $k$
and on the rule used to allocate and sample instances across clusters. In contrast, MIS determines the subset size as part of the optimization. Even the
largest MIS subsets were typically 0.2 to 0.9 times as large as the
corresponding clustering subsets. MIS achieved Spearman's $\rho \geq 0.95$
with as few as 52 instances on GPQA, 47 on MMLU-Pro, and 25 on Omni-MATH.
The median clustering subset sizes on these benchmarks were 224, 992,
and 734 instances, respectively. IFEval required about 110 instances to
reach $\rho \geq 0.95$, and the median clustering subset size was 422
instances. 

\noindent\textbf{Comparison to MIS-informed clustering baselines.} Because the above-mentioned clustering baseline does not control the subset size, we additionally implemented four MIS-informed clustering baselines and ran them at exactly the subset sizes of the recommended configuration, using the same embeddings and distance (qwen8b, standardized Euclidean): $k$-center (greedy farthest-point selection), $k$-medoids, $k$-means (the prompt nearest to each centroid) and SemDeDup~\citep{abbas2023semdedup} ($k$-means clusters with within-cluster cosine deduplication, with the deduplication threshold tuned to reach the target size), each with five seeds. All selectors return exactly the same number of prompts $n$ as the recommended MIS subset at each benchmark and threshold (e.g., 116, 200, 284 and 332 prompts at \texttt{p20} on GPQA, IFEval, MMLU-Pro and Omni-MATH). $k$-center does not form clusters: starting from a random prompt, it repeatedly adds the prompt farthest from all prompts already selected until $n$ prompts are chosen. $k$-medoids and $k$-means use $k = n$ clusters and select exactly one prompt per cluster: the medoid ($k$-medoids) or the prompt nearest to the cluster centroid ($k$-means). SemDeDup uses a fixed number of clusters, $k = N/20$ (22, 27, 50 and 50 clusters for GPQA, IFEval, MMLU-Pro and Omni-MATH), and the number of prompts kept per cluster is determined by the redundancy within the cluster: a prompt is kept only if its cosine similarity to every prompt already kept in its cluster is below $1 - \varepsilon$, with $\varepsilon$ chosen by bisection so that the total reaches $n$ (if slightly more than $n$ remain, the most redundant prompts are removed). Dense clusters of near-duplicates therefore contribute few prompts and diverse clusters many. The baselines fall into two groups. Diversity-oriented methods (MIS, $k$-center and SemDeDup) behave very similarly: at \texttt{p20}, $\rho = 0.983$, $0.982$ and $0.983$ (mean over the four benchmarks), and they remain within $0.02$ of each other up to \texttt{p90} ($0.04$ at \texttt{p95}; Tables~\ref{tab:rev_sel_full}, \ref{tab:rev_sel_mae} and~\ref{tab:rev_sel_bench}). Representativeness-oriented methods ($k$-means and $k$-medoids), which select prompts in proportion to density, match random sampling from \texttt{p10} to \texttt{p80} (at \texttt{p20}, $\rho = 0.987$ for $k$-medoids and $0.984$ for $k$-means vs.\ $0.986$ for random sampling) and have a lower MAE. The small fidelity difference between MIS and random sampling is therefore shared by diversity-oriented selection in general, which deliberately down-weights dense regions of the benchmark, rather than being specific to the MIS formulation.


\noindent\textbf{Response-based selection.} Anchor points~\citep{vivek2024anchor}, fitted on the responses of $33$ randomly chosen LLMs and evaluated on the remaining $33$ ($20$ random splits; all methods evaluated on the same held-out models), do not outperform label-free selection at mild compression (Table~\ref{tab:rev_sel_heldout}); only at $p_{95}$ are they slightly more faithful, with large variation between splits, driven mainly by GPQA. 
Response-based methods use information that MIS deliberately does not require: evaluations of a pool of models on the full benchmark.

We evaluated compression using a two-parameter logistic item response
theory (2PL IRT) model~\citep{polo2024tinybenchmarks} with 11-fold
cross-validation across the 66 LLMs. In each fold, the procedure
clusters prompts by their item parameters fitted on the training models
and uses the selected prompts to estimate full-benchmark scores for
the held-out models. To achieve a mean absolute error (MAE) of at most
3 percentage points, the procedure retained 75 prompts on MMLU-Pro,
75 on GPQA, 50 on Omni-MATH, and 30 on IFEval, for a total of 230 of
2,987 prompts (7.7\%). At a stricter MAE target of 2 percentage points,
it retained 625 prompts (20.9\%), with Kendall's $\tau$ of at least
$0.82$ between rankings based on the estimated and full-benchmark
scores. The IRT analysis quantifies how many prompts are needed to
meet a given score-error target.

\begin{table}[!htbp]
\centering
\caption{Spearman $\rho$ with the full ranking of all 66 LLMs for embedding-only selectors at the subset sizes of the recommended MIS configuration (mean over the four benchmarks).}
\label{tab:rev_sel_full}
\small
\resizebox{\textwidth}{!}{%
\begin{tabular}{lrrrrrrr}
\toprule
Type & Method & $p_{10}$ & $p_{20}$ & $p_{50}$ & $p_{80}$ & $p_{90}$ & $p_{95}$ \\
\midrule
Diversity & MIS (recommended) & 0.986 & 0.983 & 0.970 & 0.901 & 0.860 & 0.829 \\
Diversity & k-center & 0.986 & 0.982 & 0.969 & 0.909 & 0.863 & 0.836 \\
Diversity & SemDeDup & 0.987 & 0.983 & 0.963 & 0.900 & 0.846 & 0.794 \\
Representative & k-means (nearest to centroid) & 0.990 & 0.984 & 0.969 & 0.932 & 0.877 & 0.827 \\
Representative & k-medoids & 0.991 & 0.987 & 0.971 & 0.931 & 0.875 & 0.828 \\
Baseline & Random & 0.990 & 0.986 & 0.971 & 0.933 & 0.890 & 0.850 \\
\bottomrule
\end{tabular}}
\\[3pt]{\footnotesize Embedding-based selectors use qwen8b with standardized Euclidean distance (five seeds); random: 1,000 draws.}
\end{table}

\begin{table}[!htbp]
\centering
\caption{Spearman $\rho$ on held-out models: anchor points fitted on 33 LLMs, all methods evaluated on the other 33 (mean over 20 random splits and four benchmarks).}
\label{tab:rev_sel_heldout}
\small
\resizebox{\textwidth}{!}{%
\begin{tabular}{lrrrrrrr}
\toprule
Type & Method & $p_{10}$ & $p_{20}$ & $p_{50}$ & $p_{80}$ & $p_{90}$ & $p_{95}$ \\
\midrule
Diversity & MIS (recommended) & 0.983 & 0.979 & 0.966 & 0.899 & 0.860 & 0.833 \\
Diversity & k-center & 0.983 & 0.978 & 0.965 & 0.906 & 0.863 & 0.837 \\
Diversity & SemDeDup & 0.984 & 0.980 & 0.959 & 0.897 & 0.845 & 0.793 \\
Representative & k-means (nearest to centroid) & 0.987 & 0.981 & 0.965 & 0.928 & 0.876 & 0.829 \\
Representative & k-medoids & 0.988 & 0.983 & 0.967 & 0.927 & 0.873 & 0.830 \\
Baseline & Random & 0.988 & 0.983 & 0.967 & 0.928 & 0.884 & 0.845 \\
Response-based & Anchor points (response-based) & 0.979 & 0.971 & 0.945 & 0.910 & 0.879 & 0.856 \\
\bottomrule
\end{tabular}}
\end{table}

\begin{table}[!htbp]
\centering
\caption{Mean absolute error of model accuracy (accuracy points, all 66 LLMs) at the same subset sizes.}
\label{tab:rev_sel_mae}
\small
\resizebox{\textwidth}{!}{%
\begin{tabular}{lrrrrrrr}
\toprule
Type & Method & $p_{10}$ & $p_{20}$ & $p_{50}$ & $p_{80}$ & $p_{90}$ & $p_{95}$ \\
\midrule
Diversity & MIS (recommended) & 2.53 & 3.97 & 4.61 & 8.75 & 10.33 & 9.68 \\
Diversity & k-center & 2.37 & 3.32 & 4.57 & 8.24 & 8.90 & 8.49 \\
Diversity & SemDeDup & 2.85 & 3.14 & 4.52 & 5.92 & 7.31 & 9.19 \\
Representative & k-means (nearest to centroid) & 1.66 & 2.10 & 3.02 & 5.10 & 7.01 & 8.71 \\
Representative & k-medoids & 1.63 & 2.12 & 3.08 & 4.68 & 6.10 & 8.68 \\
Baseline & Random & 1.54 & 1.94 & 3.05 & 4.93 & 6.57 & 8.00 \\
\bottomrule
\end{tabular}}
\end{table}

\begin{table}[!htbp]
\centering
\caption{Spearman $\rho$ at $p_{20}$ per benchmark (all 66 LLMs).}
\label{tab:rev_sel_bench}
\small
\begin{tabular}{lrrrr}
\toprule
Method & GPQA & IFEval & MMLU-Pro & Omni-MATH \\
\midrule
MIS (recommended) & 0.975 & 0.976 & 0.985 & 0.996 \\
k-center & 0.974 & 0.977 & 0.981 & 0.997 \\
SemDeDup & 0.977 & 0.979 & 0.980 & 0.995 \\
k-means (nearest to centroid) & 0.981 & 0.971 & 0.989 & 0.996 \\
k-medoids & 0.983 & 0.977 & 0.990 & 0.997 \\
Random & 0.980 & 0.976 & 0.992 & 0.996 \\
\bottomrule
\end{tabular}
\end{table}

\appbarrier
\end{document}

%% file: math_commands.tex
\usepackage{amsmath,amsfonts,bm}

\def\eqref#1{equation~\ref{#1}}

\def\1{\bm{1}}

\DeclareMathAlphabet{\mathsfit}{\encodingdefault}{\sfdefault}{m}{sl}
\SetMathAlphabet{\mathsfit}{bold}{\encodingdefault}{\sfdefault}{bx}{n}

\DeclareMathOperator*{\argmax}{arg\,max}
\DeclareMathOperator*{\argmin}{arg\,min}